\documentclass[10pt,twocolumn,letterpaper]{article}

\usepackage{cvpr}

\usepackage{graphicx}
\usepackage{subcaption}
\usepackage[font=small]{caption}
\usepackage{float}
\usepackage{lscape}
\usepackage[lined,ruled,linesnumbered]{algorithm2e}
\usepackage{booktabs}
\usepackage{multirow}
\usepackage{paralist}
\usepackage{enumitem}
\usepackage[numbers, sort]{natbib}
\usepackage{bm}
\usepackage{epsfig}
\usepackage{mathtools}
\usepackage{amsmath}
\usepackage{amsfonts}
\usepackage{color,colortbl}
\usepackage{siunitx} 
\usepackage{comment}
\usepackage{url}  
\usepackage[pagebackref=true,breaklinks=true,colorlinks,urlcolor=blue,citecolor=blue,linkcolor=blue,bookmarks=false]{hyperref}
\usepackage{listings}
\usepackage{xspace}
\usepackage{setspace}
\usepackage{nicefrac}
\usepackage{microtype}
\usepackage[utf8]{inputenc} 
\usepackage[T1]{fontenc}    

\newlength\paramargin
\definecolor{LavenderBlue}{rgb}{0.8,0.8,1.0}
\definecolor{Lightapricot}{rgb}{0.99,0.84,0.69}

\newcommand{\mpage}[2]
{
\begin{minipage}{#1\linewidth}\centering
#2
\end{minipage}
}

\newcommand{\algoName}{NSM\xspace}
\newcommand{\algoNameFull}{Neural Shape Mating\xspace}

\begin{document}

\title{
Neural Shape Mating: \\
Self-Supervised Object Assembly with Adversarial Shape Priors
}

\author{
Yun-Chun Chen$^{1,2}$ \hspace{15pt} 
Haoda Li$^{1,2}$ \hspace{15pt} 
Dylan Turpin$^{1,2}$ \hspace{15pt} 
Alec Jacobson$^{1,4}$ \hspace{15pt} 
Animesh Garg$^{1,2,3}$ \\
\vspace{5mm} 
$^{1}$University of Toronto \hspace{20pt} 
$^{2}$Vector Institute \hspace{20pt} 
$^{3}$NVIDIA \hspace{20pt} 
$^{4}$Adobe Research, Toronto
}

\twocolumn[{
\renewcommand\twocolumn[1][]{#1}
\maketitle
\begin{center}
  \centering
  \includegraphics[width=0.95\linewidth]{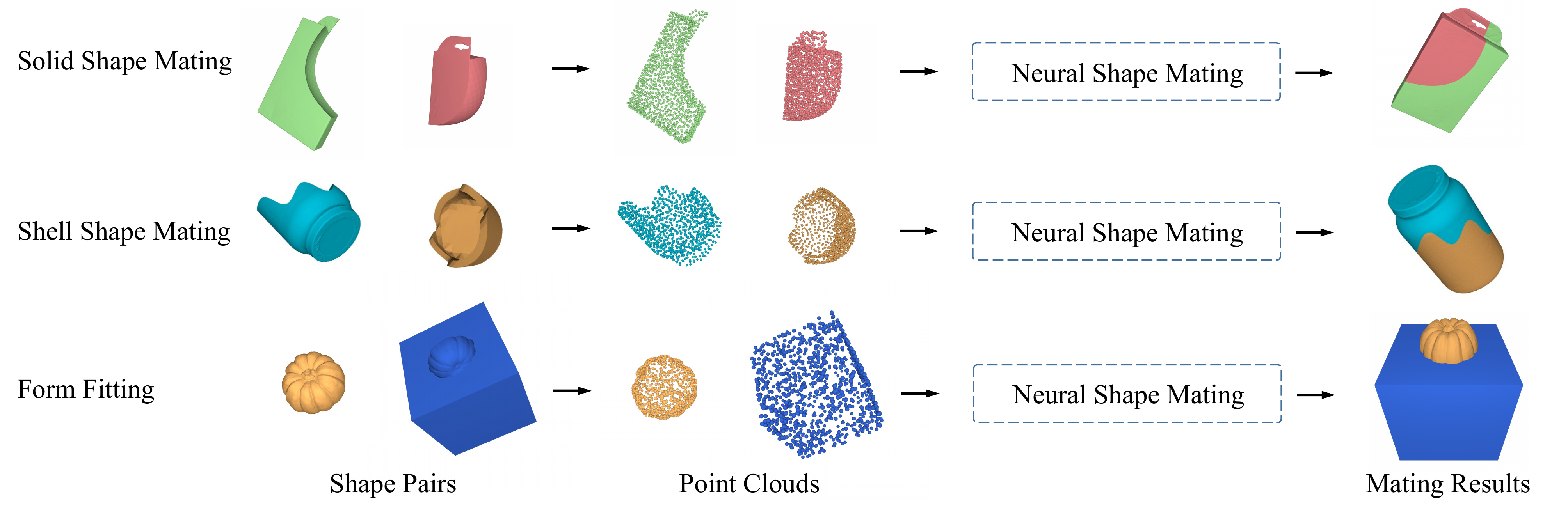}
  \vspace{-3.5mm}
  \captionof{figure}{ 
  \textbf{Pairwise 3D geometric shape mating.} 
  \algoNameFull (\algoName) takes as input the point clouds of a pair of shapes and predicts the mating configuration as output.
  \algoName learns to mate shapes together with self-supervision and does not require semantic information or target shapes as guidance at test time.
  Our method can be applied to various shape mating settings, including solid shape mating (\emph{top row}), shell shape mating (\emph{middle row}), and form fitting (\emph{bottom row}).
  }
  \label{fig:teaser}
\end{center}
}]

\begin{abstract}
\vspace{-3.0mm}
Learning to autonomously assemble shapes is a crucial skill for many robotic applications.
While the majority of existing part assembly methods focus on correctly posing semantic parts to recreate a whole object, we interpret assembly more literally: as mating geometric parts together to achieve a snug fit.
By focusing on shape alignment rather than semantic cues, we can achieve across-category generalization.
In this paper, we introduce a novel task, pairwise 3D geometric shape mating, and propose \algoNameFull (\algoName) to tackle this problem.
Given the point clouds of two object parts of an unknown category, \algoName learns to reason about the fit of the two parts and predict a pair of 3D poses that tightly mate them together.
We couple the training of \algoName with an implicit shape reconstruction task to make \algoName more robust to imperfect point cloud observations.
To train \algoName, we present a self-supervised data collection pipeline that generates pairwise shape mating data with ground truth by randomly cutting an object mesh into two parts, resulting in a dataset that consists of 200K shape mating pairs from numerous object meshes with diverse cut types.
We train \algoName on the collected dataset and compare it with several point cloud registration methods and one part assembly baseline.
Extensive experimental results and ablation studies under various settings demonstrate the effectiveness of the proposed algorithm.
Additional material is available at: \href{https://neural-shape-mating.github.io/}{https://neural-shape-mating.github.io/}.
\end{abstract}

\vspace{-6.0mm}
\section{Introduction}
\vspace{-1.0mm}

The human-built world is filled with objects that are shaped to fit, snap, connect, or mate together.  
Reassembling a broken object and inserting a plug into a socket are both instances of shape mating. 
This kind of geometric mating has many practical applications and appears in domains ranging from computer graphics~\cite{li2012stackabilization}, 3D design~\cite{chen2015dapper,jacobson2017generalized}, robotics~\cite{wang2019stable,devgon2021kit,zakka2020form2fit,zeng2020transporter}, and biology~\cite{sverrisson2021fast}. 

There have been many attempts that learn shape-to-shape  matching algorithms in application-specific domains: furniture assembly~\cite{lee2019ikea,3DPartAssembly,huang2020generative}, object assembly~\cite{agrawala2003designing,litany2016non}, and object packing~\cite{wang2019stable}.
Most of these assembly algorithms operate under the assumption that each shape corresponds to a recognizable semantic object part~\cite{lee2019ikea,3DPartAssembly,huang2020generative}.
While these results are promising, they rely heavily on semantic information (e.g., part segmentation), target shapes~\cite{3DPartAssembly} as guidance, and ground-truth part pose annotations~\cite{3DPartAssembly,huang2020generative}. This reliance makes these methods application-specific, hard to scale, and difficult to generalize.

In this paper, we consider shape mating from a geometric perspective, without relying on semantic information or prespecified target shapes as guidance. 
Specifically, we study the \emph{pairwise} 3D geometric shape mating task, where shape mating is done based solely on geometric cues.
To achieve this, we propose \algoNameFull (\algoName).
As shown in Figure~\ref{fig:teaser}, given a pair of shapes in the form of point clouds with random initial poses, \algoName predicts a plausible mating configuration for them by reasoning about geometric fits.
The proposed task is challenging yet practical with applications in robotics such as object kitting~\cite{devgon2021kit} and form fitting~\cite{zakka2020form2fit} and in biology such as protein binding~\cite{sverrisson2021fast} (where the binding between proteins requires reasoning about the geometric fit between two proteins), and can also be seen as an integral subroutine in the broader problem of multi-part geometric assembly including applications in robotics~\cite{lee2019ikea} and AR/VR~\cite{syberfeldt2015visual}.

We formulate the proposed task as a pose prediction problem and develop a Transfomer-based network~\cite{Transformer} that takes as input the point clouds of the two shapes, reasons about the fit by attending to asymmetric correlations between local geometric cues, and predicts respective poses that bring them together.
In addition, we adopt an adversarial learning scheme that learns shape priors for evaluating the plausibility of the predicted shape mating configurations.
To account for imperfect point cloud observations (e.g., noisy point clouds), we couple the training of \algoName with an implicit shape reconstruction task~\cite{park2019deepsdf,sitzmann2020metasdf}.

To train \algoName, we present a self-supervised data collection pipeline that generates pairwise shape mating data with ground truth by randomly cutting an object mesh into two parts with different types of cuts.
We collect object meshes from the Thingi10K~\cite{Thingi10K}, Google Scanned Objects~\cite{GoogleScannedObjects}, and ShapeNet~\cite{ShapeNet} datasets and apply our data generation algorithm to each object mesh.
The resulting geometric shape mating dataset covers a diverse set of cut types applied to numerous object instances of 11 categories, combining a total of 200K shape pairs suitable for studying the proposed task.
We train \algoName on the collected dataset in a self-supervised fashion and compare our method with several point cloud registration algorithms and one part assembly baseline approach.
Extensive experimental results and ablation studies under various settings demonstrate the effectiveness of the proposed algorithm.

\vspace{\paramargin}
\paragraph{Summary of contributions:}
\begin{enumerate}[
    itemsep=-0.5ex,
    topsep=0pt,
    leftmargin=*]
  \item We introduce a novel task of pairwise geometric shape mating and propose \algoNameFull that predicts mating configurations based on geometric cues.
  \item We collect a large-scale geometric shape mating dataset for studying the proposed task.
  \item We compare \algoName with several point cloud registration methods and one part assembly baseline approach.
  \item Experimental results and analysis support our design choices and demonstrate the robustness of \algoName when presented with realistically noisy observations.
\end{enumerate}

\section{Related Work}\label{sec:rel}

\paragraph{3D shape assembly.} 
A distinct, but related, line of work investigates generative models that learn to represent objects as concatenations of simple 3D shapes.
\cite{abstractionTulsiani17} trains per-class models that generate objects by assembling volumetric primitives (cuboids).
\cite{khan2019unsupervised} trains a single model that can generate cuboid primitives across all classes.
\cite{jones2020shapeassembly} models objects with ShapeAssembly programs, learned by a variational autoencoder (VAE)~\cite{kingma2013auto}, which can be executed to generate 3D shapes as concatenations of cuboids.
These methods provide powerful abstractions and reveal correspondences between objects by abstracting away details of local geometry.
In contrast, we consider the problem of discovering plausible fits between shapes with complex geometry that do not correspond to any semantic part or natural object decomposition.
The validity of a fit relies on the alignment of detailed local geometric features, which provide cues for shape mating.

The task that comes closest to our own is part assembly~\cite{yin2020coalesce,3DPartAssembly,huang2020generative,harish2022rgl,jones2021automate,willis2021joinable}, which aims at making a complete object from a set of parts. 
\cite{3DPartAssembly} learns to predict translations and rotations for part point clouds to assemble a target object specified by an image.
\cite{huang2020generative,harish2022rgl} frame the part assembly task as graph learning and learn to assemble parts into complete objects by iterative message passing.
These methods use the PartNet~\cite{Mo_2019_CVPR} dataset, and thus the parts to assemble are always a reasonable semantic decomposition of the target object.
Shape is important in part assembly, but cues can also be taken from part semantics directly, bypassing the geometric cues.
In contrast, we consider the problem of learning to fit together pieces with no particular semantics and without a provided target. 

\vspace{\paramargin}
\paragraph{Pose estimation.}
Existing pose estimation methods predict poses for known objects by aligning a provided model with an observation~\cite{ICP,zeng2017multi}.
Learning-based approaches predict poses for novel objects as  bounding box corners~\cite{law2019cornernet} or semantic keypoints~\cite{you2020keypointnet,wang20206} or mappings to a normalized coordinate space~\cite{wang2019normalized}.
Rather than independently estimating the current pose of a single observed object, our method leverages cross-shape information to predict a new pose for each observed shape that assembles them into a whole object.

\vspace{\paramargin}
\paragraph{Learning shape priors.} 
Our model includes an adversarial prior implemented by a discriminator that learns to distinguish between ground-truth assembled shape pairs and the predicted ones.
Conditional generative adversarial networks~\cite{goodfellow2014generative,mirza2014conditional} have achieved impressive results on image generation tasks even when the paired ground truth is unavailable, as in unpaired image-to-image translation~\cite{zhu2017unpaired}, or when the ground truth is available but multiple plausible outputs exist, as in MUNIT~\cite{MUNIT}.
Even when the ground truth is available and a unimodal correct output exists, adversarial losses lead to enhanced detail and more realistic outputs, e.g., for super-resolution \cite{lucas2019generative}.
In our problem, we learn shape priors with an adversarial loss that encourages our model to generate plausible shape mating configurations.

\vspace{\paramargin}
\paragraph{Implicit shape reconstruction.} 
A core problem in computer vision is reconstructing 3D shapes from 2D observations.
Rather than representing the reconstructed shapes as finite sets of points, voxels, or meshes, a recent line of work aims to represent them as implicit functions parameterized by neural networks.  
These encode shapes by their signed distance functions (SDFs)~\cite{park2019deepsdf,sitzmann2020metasdf} or the indicator functions~\cite{OccupancyNetworks}, which are continuous, differentiable, and can be queried at arbitrary resolution.
DeepSDF~\cite{park2019deepsdf} learns SDFs for many shape classes with a feedforward neural network.
Further work~\cite{genova2019learning,Genova_2020_CVPR} adds an additional structure to improve reconstruction accuracy and memory efficiency.
We follow a similar approach to \cite{karunratanakul2020Grasping,jiang2021synergies,zhu2022correspondence}, which take inspiration from implicit reconstruction to improve performance on a pose prediction task.
Specifically, as in \cite{jiang2021synergies,zhu2022correspondence}, we include implicit shape reconstruction as an auxiliary task and show, through ablation, that this improves performance on our main shape mating task, suggesting significant synergies between shape mating and shape reconstruction.

\vspace{\paramargin}
\paragraph{Point cloud registration.} 
If we had access to additional information, our problem would reduce to point cloud registration~\cite{zhou2016fast,ICP,SparseICP}.
Specifically, if we had a segmentation of the interface of each piece (the subset of its surface that contacts the other piece in the assembled pose),
computing the assembled pose would reduce to aligning the paired interfaces.
If we were given correspondences between these interfaces, alignment would become a well-characterized optimization problem solvable with Procrustes.
Without correspondences, we would be left with a registration problem.
Feature-free methods such as Iterative Closest Point (ICP)~\cite{ICP} approximate correspondences simply as pairs of closest points.
Sparse ICP~\cite{SparseICP} improves robustness to noise by distinguishing between inliers and outliers.
Learning-based methods seek to approximate correspondences in a learned feature space~\cite{deng2018ppfnet,gojcic2019perfect,DCP}.
Unlike registration methods which aim to align two point clouds with (partial) overlap, our method is designed to predict paired poses that bring two \emph{disjoint} shapes together to form a single whole object.

\section{Pairwise 3D Geometric Shape Mating}

We formulate the pairwise 3D geometric shape mating task as a pose prediction problem.
In this task, we assume we are given the point clouds $P_A$ and $P_B$ of the two shapes $S_A$ and $S_B$, where $P_A = \{p_i^A\}_{i=1}^N$, $p_i^A \in \mathbb{R}^3$ is a point in $P_A$, $P_B = \{p_j^B\}_{j=1}^M$, $p_j^B \in \mathbb{R}^3$ is a point in $P_B$, and $N$ and $M$ denote the number of points in point clouds $P_A$ and $P_B$, respectively.
Shape $S_A$ and shape $S_B$ are the two parts of a whole object $S$.
We aim to learn a model that takes as input the two point clouds $P_A$ and $P_B$ and predicts a canonical SE(3) pose $\{(R_k, T_k)\ |\ R_k \in \mathbb{R}^{3 \times 3}, T_k \in \mathbb{R}^3\}$ for each point cloud $P_k$, where $R_k$ denotes the rotation matrix, $T_k$ is the translation vector, and $k \in \{A, B\}$.\footnote{We follow recent methods~\cite{3DPartAssembly,huang2020generative} and perform canonical pose estimation to ensure that the model is symmetric.}
The predicted SE(3) poses will then be applied to transform the pose of each respective input point cloud.
The union of the two pose-transformed point clouds $\bigcup_{k \in \{A, B\}}R_kP_k+T_k$ forms the shape mating result.

\section{Method: \algoNameFull}

We describe \algoNameFull and the loss functions used to train our model in this section.

\begin{figure*}[t]
  \centering
  \includegraphics[width=0.9\linewidth]{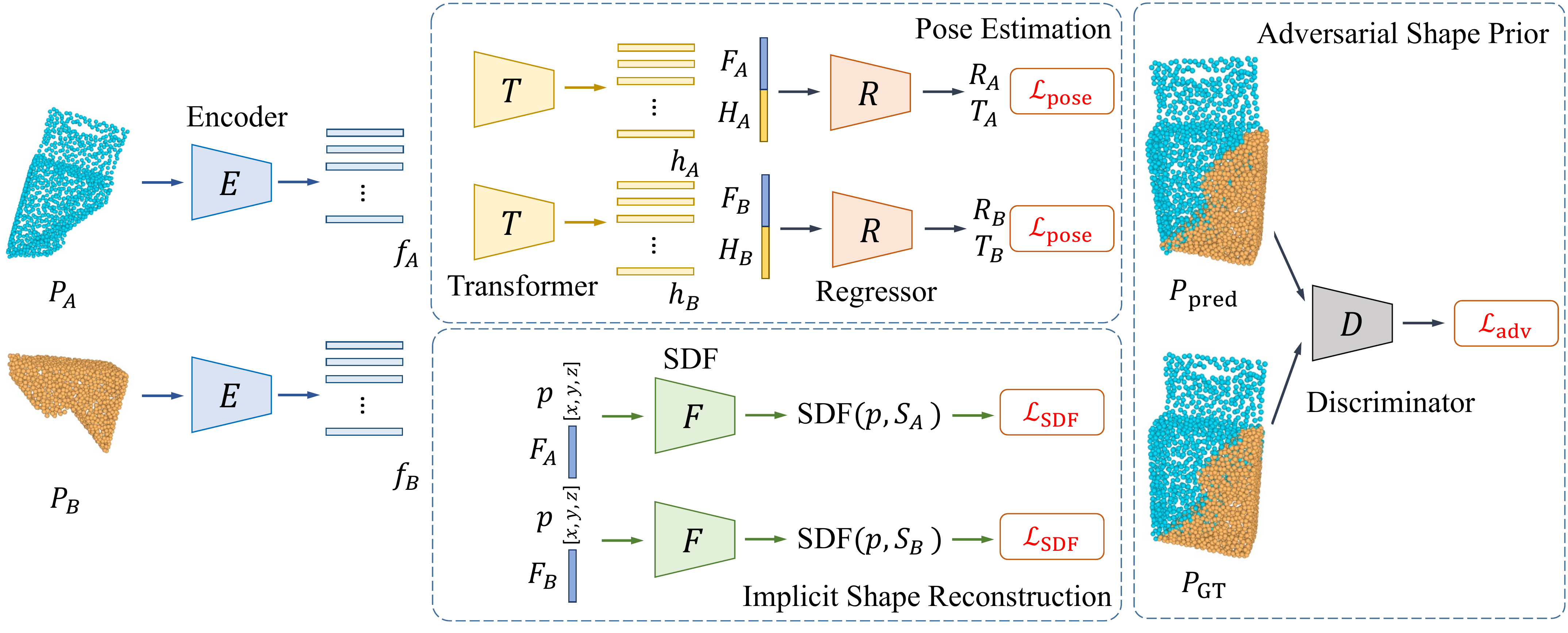}
  \vspace{-3.0mm}
  \caption{
  \textbf{Overview of \algoNameFull.} 
  \algoNameFull is composed of four main components: a point cloud encoder $E$, a pose estimation module that consists of a Transformer network $T$ and a regressor network $R$, an implicit shape reconstruction module that learns signed distance functions (SDFs), and a discriminator $D$ for learning shape priors.
  }
  \label{fig:model}
  \vspace{-5.0mm}
\end{figure*}

\subsection{Algorithmic overview}

Given two point clouds $P_A$ and $P_B$, our goal is to learn a model that predicts an SE(3) pose for each input point cloud.
We propose \algoNameFull, which comprises four components: 1) a point cloud encoder, 2) a pose estimation network, 3) an adversarial shape prior module, and 4) an implicit shape reconstruction network. 

As shown in Figure~\ref{fig:model}, we first apply the point cloud encoder $E$ to point clouds $P_A$ and $P_B$ to extract point features $f_A = E(P_A) = \{f_i^A\}_{i=1}^N$ and $f_B = E(P_B) = \{f_j^B\}_{j=1}^M$, respectively, where $f_i^A \in \mathbb{R}^d$ and $f_j^B \in \mathbb{R}^d$.
Next, the point features $f_A$ and $f_B$ are passed to the pose estimation network to reason about the fit between the two point clouds $P_A$ and $P_B$ and predict SE(3) poses $\{R_k, T_k\}$ for $k \in \{A, B\}$ for them.
The point features $f_A$ and $f_B$ are also passed to the SDF network $F$ for learning implicit shape reconstruction.
The predicted SE(3) poses are then applied to transform the pose of the respective input point clouds.
The union of the two pose-transformed point clouds forms the shape mating result $P_\text{pred}$.
To learn plausible shape mating configurations, we have a discriminator that takes as input the predicted mating configuration $P_\text{pred}$ and the ground truth $P_\text{GT}$ and distinguishes whether the mating configurations look visually realistic or not.

\vspace{\paramargin}
\paragraph{Point cloud encoder.}
There are several point cloud models such as PointNet~\cite{PointNet}, PointNet++~\cite{PointNet++}, and DGCNN~\cite{DGCNN} that are applicable for learning point features.
In this work, we follow DCP~\cite{DCP} and adopt DGCNN as our point cloud encoder $E$. 
The dimension $d$ of the point features $f_i^A$ and $f_j^B$ is $1,024$ (i.e., $f_i^A \in \mathbb{R}^{1024}$ and $f_j^B \in \mathbb{R}^{1024}$).
We refer the reader to \cite{DGCNN} for more details of DGCNN.

\vspace{\paramargin}
\paragraph{Rotation representation.}
We follow prior work~\cite{3DPartAssembly} and use quaternion to represent rotations.

\subsection{Pose estimation for shape assembly}

To achieve shape mating, we predict an SE(3) pose for each input point cloud.
Unlike existing object pose estimation methods~\cite{you2020keypointnet,wang20206} that independently predict a pose for each object, our task requires reasoning about the fit between the two input point clouds for pose prediction.
To achieve this, we have a feature correlation module $T$ that captures cross-shape information for providing geometric cues and a regressor $R$ for predicting poses.

We adopt a Transformer network~\cite{Transformer} as our feature correlation module $T$, as it allows the model to learn asymmetric cross-shape information.
Given the point features $f_A = \{f_i^A\}_{i=1}^N$ and $f_B = \{f_j^B\}_{j=1}^M$ as input, the feature correlation module $T$ computes pairwise feature correlation between each point feature $f_i^A \in f_A$ and each point feature $f_j^B \in f_B$ to obtain feature $h_A = \{h_i^A\}_{i=1}^N$ for point cloud $P_A$ and feature $h_B = \{h_j^B\}_{j=1}^M$ for point cloud $P_B$, where $h_i^A \in \mathbb{R}^d$ and $h_j^B \in \mathbb{R}^d$.
The details of the Transformer network are provided in Appendix~\ref{app:transformer}.

To predict SE(3) poses, we aggregate all features $h_i^A$ in $h_A$ to obtain the feature $H_A \in \mathbb{R}^d$ and all features $h_j^B$ in $h_B$ to obtain the feature $H_B \in \mathbb{R}^d$.
Similarly, we aggregate all point features $f_i^A$ in $f_A$ to obtain the feature $F_A \in \mathbb{R}^d$ and all point features $f_j^B$ in $f_B$ to obtain the feature $F_B \in \mathbb{R}^d$.
We use max pooling for feature aggregation as in PointNet~\cite{PointNet}.
The features $F_A$ and $H_A$ are concatenated (resulting in a feature of dimension $2d$) and passed to the regressor $R$ to predict a unit quaternion $q_A$ (which can be converted to a rotation matrix $R_A$) and a translation vector $T_A$.
The prediction of $q_B$ (or $R_B$) and $T_B$ can be similarly derived.\footnote{We normalize the predicted quaternion $q_k$ by its length so that $\|q_k\|_2~=~1$ for $k \in \{A, B\}$.}

To guide the learning of the pose estimation network, we have a pose prediction loss $\mathcal{L}_\text{pose}$, which is defined as
\begin{equation}
  \mathcal{L}_\text{pose} = \sum_{k \in \{A, B\}}^{}\|R_k^\top R_k^\text{GT} - I\| + \|T_k - T_k^\text{GT}\|, 
  \label{eq:pose}
\end{equation}
where $R_k^\text{GT}$ and $T_k^\text{GT}$ denote the ground-truth rotation and translation, respectively, and $I$ is the identity matrix.

\begin{figure*}[t]
  \begin{center}
  \includegraphics[width=0.90\linewidth]{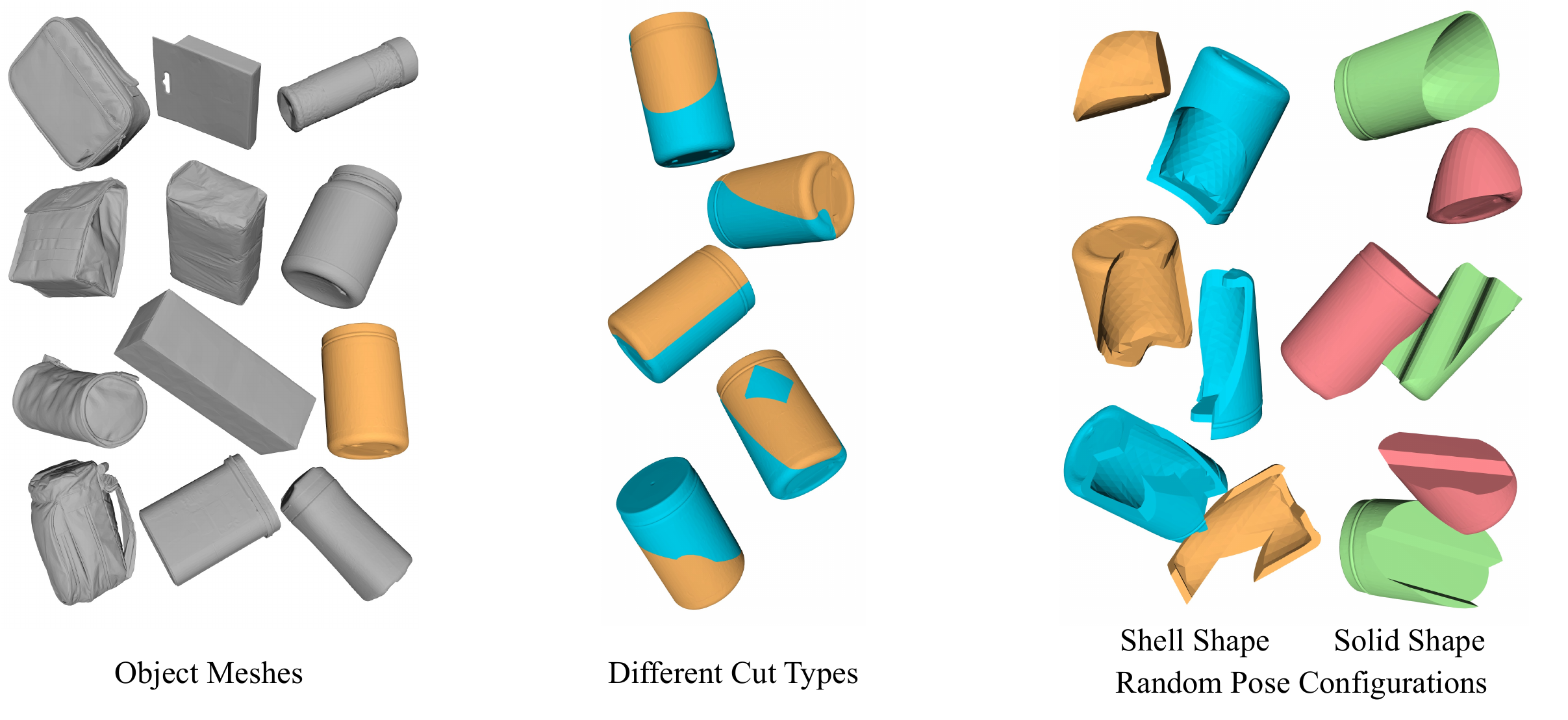}
  \vspace{-4.0mm}
  \caption{
  \textbf{Dataset overview.} 
  (\emph{Left}) Our dataset is composed of object meshes from 11 categories.
  (\emph{Middle}) We define five different types of cut functions. Each object mesh can then be cut with many different ways using varying parametric cut functions.
  (\emph{Right}) Each pair of parts can be randomized with an initial SE(3) pose. In our dataset, we also generate solid and shell variations of each shape when cutting a mesh to create different mating interfaces for the same problem instance.
  }
  \label{fig:dataset}
  \end{center}
  \vspace{-8.0mm}
\end{figure*}

\subsection{Adversarial learning of shape priors}

To encourage \algoName to predict plausible shape mating results, we propose to learn global shape priors to further constrain the prediction space.
We exploit the idea that when the two point clouds are mated together using the predicted poses, the mating configuration should look visually realistic like an object.
We cast this as an adversarial learning problem and introduce a discriminator $D$ that takes as input the predicted mating result $P_\text{pred} = \bigcup_{k \in \{A, B\}}R_kP_k+T_k$ and the ground-truth mating configuration $P_\text{GT} = \bigcup_{k \in \{A, B\}}R_k^\text{GT}P_k+T_k^\text{GT}$ and distinguishes whether the input mating configurations look visually realistic like an object or not.

To achieve this, we have a loss $\mathcal{L}_\text{G}$ for training the generator (i.e., the point cloud encoder and the pose prediction network), which is defined as
\begin{equation}
  \mathcal{L}_\text{G} = \mathbb{E}_{}\big[\|D(P_\text{pred}) - 1\|\big],
  \label{eq:G}
\end{equation}
and an adversarial loss $\mathcal{L}_\text{adv}$ for training the discriminator $D$, which is defined as
\begin{equation}
  \mathcal{L}_\text{adv} = \mathbb{E}_{}\big[\|D(P_\text{pred})\|\big] + \mathbb{E}_{}\big[\|D(P_\text{GT}) - 1\|\big].
  \label{eq:adv}
\end{equation}

\noindent Having this adversarial training scheme allows \algoName to predict poses that result in plausible mating results.
The details of adversarial learning are provided in Appendix~\ref{app:adv-training}.

\subsection{Implicit shape reconstruction}

Since the same object can be described by different point clouds, we couple the training of \algoName with an implicit shape reconstruction task to account for the noise in point cloud sampling.
This is motivated by recent advances in implicit shape modeling~\cite{DeepSDF,OccupancyNetworks}, where learning SDFs allows the model to learn more robust shape representations.
Specifically, we have an SDF network $F$ that takes as input the aggregated features $F_A$ and $F_B$, respectively, and a point $p \in \mathbb{R}^3$, and predicts the signed distances between point $p$ and shape $S_A$ and between point $p$ and shape $S_B$.

To train the SDF network, we have an SDF regression loss $\mathcal{L}_\text{SDF}$, which is defined as
\begin{equation}
  \mathcal{L}_\text{SDF} = \sum_{k \in \{A, B\}}^{}\|\text{SDF}(p, S_k) - \text{SDF}_\text{GT}(p, S_k)\|, 
  \label{eq:sdf}
\end{equation}
where $\text{SDF}(p, S_k)$ and $\text{SDF}_\text{GT}(p, S_k)$ denote the predicted and the ground-truth signed distances between the point $p$ and the shape $S_k$, respectively.

\begin{table*}[!t]
  \begin{center}
  \scriptsize
  \caption{
  \textbf{Experimental results of geometric shape mating.}
  $R$ and $T$ denote rotation and translation, respectively. 
  Lower is better on all metrics. 
  It is worth noting that many methods can get reasonably close in position, but be completely off in orientation as demonstrated by the RMSE error in rotation. 
  \algoName outperforms the best baseline in predicting the correct orientation by up to $4\times$ in MAE. 
  }
  \label{exp:main-results}
  \vspace{-2.5mm}
  {
  \begin{tabular}{lS[table-format=5.2]|S[table-format=5.2]S[table-format=5.2]S[table-format=5.2]|S[table-format=5.2]S[table-format=5.2]S[table-format=5.2]}
  \toprule
  \rowcolor{LavenderBlue}
  Method & \text{Shape Matching Type} & \text{MSE ($R$)} & \text{RMSE ($R$)} & \text{MAE ($R$)} & \text{MSE ($T$)} & \text{RMSE ($T$)} & \text{MAE ($T$)}  \\ 
  \midrule
  \rowcolor{Lightapricot}
  \multicolumn{2}{l}{Solid Shape Mating} & \multicolumn{3}{c}{} & \text{($\times10^{-3}$)} & \text{($\times10^{-3}$)} & \text{($\times10^{-3}$)} \\ 
  \midrule
  ICP (point-to-point)~\cite{ICP} & \text{\multirow{4}{*}{Local}} & 9108.79 & 95.44 & 83.31 & 211.76 & 460.18 & 381.03 \\
  ICP (point-to-plane)~\cite{ICP} &  & 6748.62 & 82.15 & 74.31 & 82.15 & 286.61 & 203.17 \\
  Sparse ICP (point-to-point)~\cite{SparseICP} &  & 4751.34 & 68.93 & 65.44 & 33.89 & 184.09 & 152.63 \\
  Sparse ICP (point-to-plane)~\cite{SparseICP} &  & 3267.27 & 57.16 & 53.27 & 39.30 & 198.23 & 178.35 \\
  \midrule
  DCP~\cite{DCP} & \text{\multirow{3}{*}{Global}} & 3400.06 & 58.31 & 56.17 & 55.26 & 235.08 & 227.41 \\
  GNN Assembly~\cite{3DPartAssembly} &  & 1087.68 & 32.98 & 30.42 & 19.23 & 138.67 & 125.08 \\
  \textbf{\algoNameFull (\algoName)} & & 94.67 & 9.73 & 7.61 & 15.48 & 124.40 & 110.44 \\ 
  \midrule
  \rowcolor{Lightapricot}
  \multicolumn{8}{l}{Shell Shape Mating} \\ 
  \midrule
  ICP (point-to-point)~\cite{ICP} & \text{\multirow{4}{*}{Local}} & 8725.43 & 93.41 & 89.07 & 608.81 & 780.26 & 747.32 \\
  ICP (point-to-plane)~\cite{ICP} & & 6696.15 & 81.83 & 79.63 & 463.88 & 681.09 & 658.13 \\
  Sparse ICP (point-to-point)~\cite{SparseICP} & & 5099.39 & 71.41 & 69.93 & 395.74 & 629.08 & 601.44 \\
  Sparse ICP (point-to-plane)~\cite{SparseICP} & & 3517.68 & 59.31 & 56.62 & 327.91 & 572.63 & 556.74 \\
  \midrule
  DCP~\cite{DCP} & \text{\multirow{3}{*}{Global}} & 3861.38 & 62.14 & 59.03 & 336.53 & 580.11 & 653.02 \\
  GNN Assembly~\cite{3DPartAssembly} & & 1662.19 & 40.77 & 38.49 & 113.69 & 337.18 & 320.01 \\
  \textbf{\algoNameFull (\algoName)} & & 290.02 & 17.03 & 14.52 & 107.60 & 328.03 & 301.19 \\ 
  \bottomrule
  \end{tabular}
  }
  \end{center}
  \vspace{-4.0mm}
\end{table*}

\begin{figure*}[t]
  \begin{center}
  \mpage{0.123}{\includegraphics[width=\linewidth]{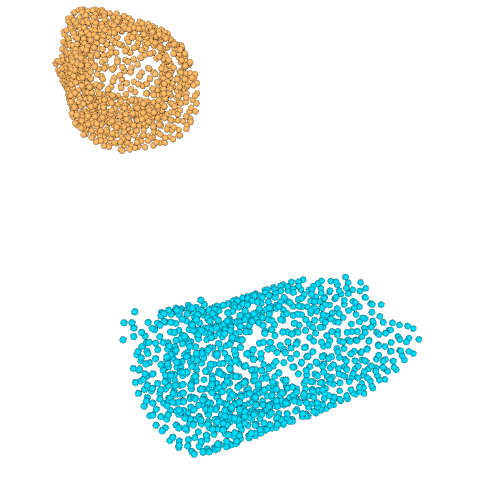}} \hfill
  \mpage{0.123}{\includegraphics[width=\linewidth]{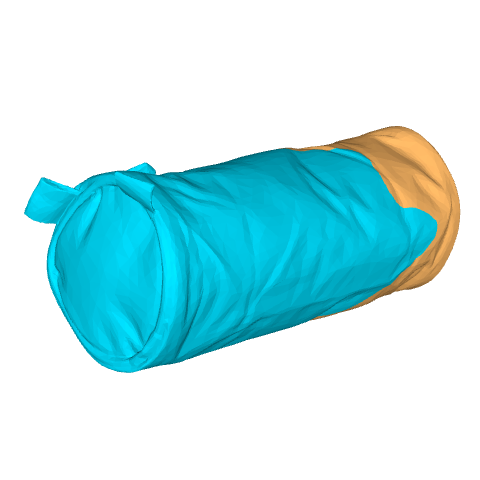}} \hfill
  \mpage{0.123}{\includegraphics[width=\linewidth]{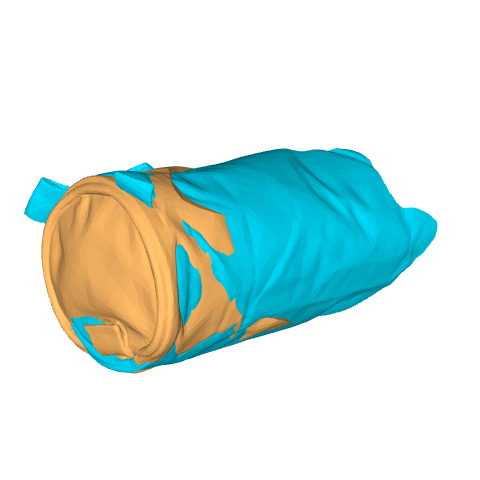}} \hfill
  \mpage{0.123}{\includegraphics[width=\linewidth]{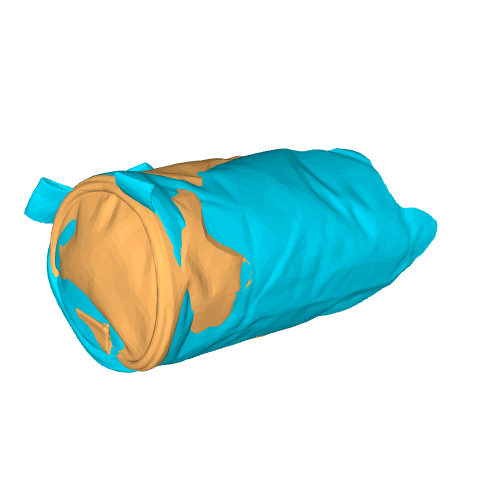}} \hfill
  \mpage{0.123}{\includegraphics[width=\linewidth]{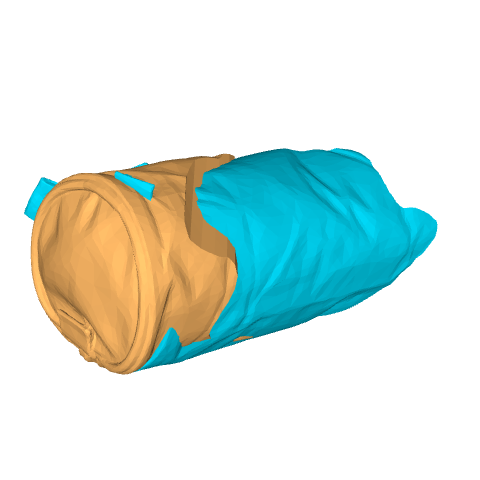}} \hfill
  \mpage{0.123}{\includegraphics[width=\linewidth]{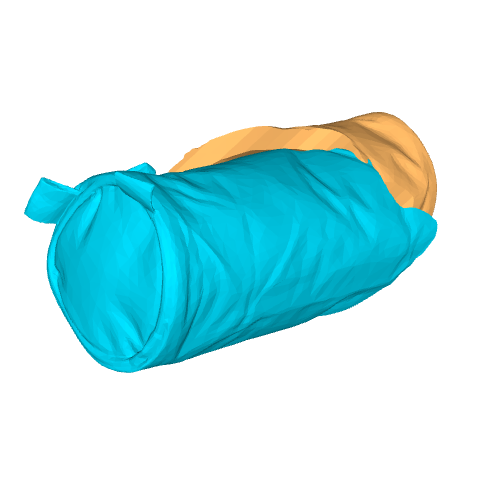}} \hfill
  \mpage{0.123}{\includegraphics[width=\linewidth]{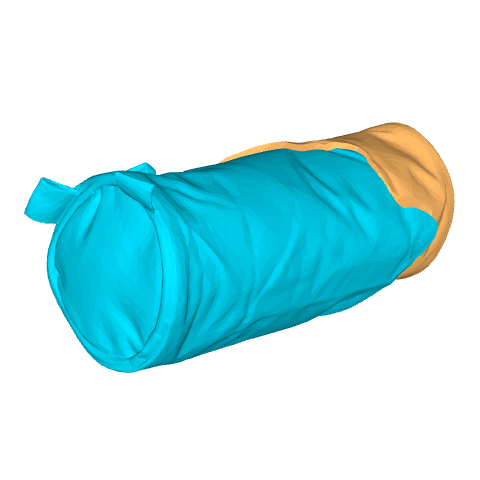}} \\
  \vspace{-0.5mm}
  \mpage{0.123}{\small Input} \hfill
  \mpage{0.123}{\small Ground Truth} \hfill
  \mpage{0.123}{\small ICP} \hfill
  \mpage{0.123}{\small Sparse ICP} \hfill
  \mpage{0.123}{\small DCP} \hfill
  \mpage{0.123}{\small GNN Assembly} \hfill
  \mpage{0.123}{\small \algoName} \\
  \vspace{-2.0mm}
  \caption{
  \textbf{Visual results of pairwise geometric shape mating.}
  \algoName predicts poses to accurately mate the two shapes together to make a bag.
  ICP, Sparse~ICP and DCP methods estimate a pose for the yellow point cloud that aligns with the blue one.
  Both GNN Assembly and NSM (our method) predict poses for both the yellow and the blue point clouds.
  }
  \label{fig:geometric-shape-assembly-results}
  \end{center}
  \vspace{-8.0mm}
\end{figure*}

\section{The Geometric Shape Mating Dataset}

To train \algoName, we present a self-supervised method that generates pairwise geometric shape mating data with ground truth by randomly cutting an object mesh into two parts.

\vspace{\paramargin}
\paragraph{Mesh cutting.}
We normalize each object mesh by the longest bounding box length such that the object mesh has a maximum bounding box length of 1 and the aspect ratio remains the same.
To cut the object, we use the mesh boolean functions provided by \texttt{libigl}~\cite{libigl}.
We construct a heightfield that will be used to intersect the object mesh for mesh cutting.
The heightfield can be parameterized by different functions.
In our work, we define five different types of functions, including a planar function, a sine function, a parabolic function, a square function, and a pulse function (see Appendix~\ref{app:dataset-stats} for more details).
Each of these functions will result in a type of cut.
We generate two types of shapes when performing cutting: the solid shape and the shell shape.
To generate solid shape data, we use the heightfield to intersect with each object mesh.
To generate shell shape data, we first construct an offset surface at the $-0.05$ level set of an object.
We then compute the difference between the original object mesh and the generated offset surface and use the heightfield to intersect with it.
We set the cut-off ratio to be no less than 25\% (each object part mesh has a volume of at least 25\% of the uncut mesh).
Figure~\ref{fig:dataset} presents example data generated by applying different types of cuts.
More visual examples are provided in Appendix~\ref{app:dataset-visual-results}.

\vspace{\paramargin}
\paragraph{Point cloud sampling.} 
We uniformly sample 1,024 points on each object part mesh (i.e., $N = 1,024$ and $M = 1,024$).

\vspace{\paramargin}
\paragraph{Signed distance ground truth.} 
We use the Fast Winding Numbers method~\cite{barill2018fast} for computing ground-truth signed distances.
For each part mesh, we sample 40,000 points that are close to the mesh surface.

\vspace{\paramargin}
\paragraph{Pose transformation.} 
Each point cloud is zero-centered.
During training, we sample two rotation matrices on the fly and apply them to transform the pose of the two input point clouds, respectively.

\vspace{\paramargin}
\paragraph{Statistics.}
We use 11 shape categories: \texttt{bag}, \texttt{bowl}, \texttt{box}, \texttt{hat}, \texttt{jar}, \texttt{mug}, \texttt{plate}, \texttt{shoe}, \texttt{sofa}, \texttt{table}, and \texttt{toy} in initial dataset version due to computational reasons.
We note that the proposed procedural data generation can be extended naively to other shape categories.
The object meshes are collected from the Thingi10K~\cite{Thingi10K}, Google Scanned Objects~\cite{GoogleScannedObjects}, and ShapeNet~\cite{ShapeNet} datasets.
The dataset statistics are provided in Appendix~\ref{app:dataset-stats}.

\section{Experiments}

We perform evaluations and analysis of NSM to answer the following questions:
\begin{enumerate}[
    itemsep=-0.5ex,
    topsep=0.0pt,
    leftmargin=*]
  \item How well does \algoName perform when compared to point cloud registration methods and graph network-based assembly baseline approaches?
  \item Can \algoName generalize to unseen object categories and unseen cut types?
  \item How does \algoName perform when presented with more realistic, noisy point clouds?
  \item How much do the adversarial, reconstruction and pose losses contribute to final performance?
\end{enumerate}

\begin{table*}[!t]
  \begin{center}
  \scriptsize
  \caption{
  \textbf{Experimental results of model generalization.}
  (a) Results of unseen category geometric shape mating.
  The test set contains shape pairs from the box and bag category.
  The training set contains shape pairs from the remaining 9 categories.
  (b) Results of unseen cut type geometric shape mating.
  The training set contains the planar, sine, square and pulse cut types.
  The test set contains the parabolic cut type.
  Results reported are in the solid shape mating setting. 
  }
  \vspace{-2.0mm}
  \subcaption{Unseen category geometric shape mating.}
  \label{exp:unseen-categories-solid-objects}
  \vspace{-0.5mm}
  {
  \begin{tabular}{l|S[table-format=5.2]S[table-format=5.2]S[table-format=5.2]|S[table-format=5.2]S[table-format=5.2]S[table-format=5.2]}
  \toprule
  \rowcolor{LavenderBlue}
  Method & \text{MSE ($R$)} & \text{RMSE ($R$)} & \text{MAE ($R$)} & \text{MSE ($T$)} & \text{RMSE ($T$)} & \text{MAE ($T$)}  \\ 
  \midrule
  \rowcolor{Lightapricot}
  \multicolumn{1}{l}{Solid Shape Mating} & \multicolumn{3}{c}{} & \text{($\times10^{-3}$)} & \text{($\times10^{-3}$)} & \text{($\times10^{-3}$)} \\ 
  \midrule
  DCP~\cite{DCP}                     & 6567.48 & 81.04 & 78.92 & 108.67 & 329.65 & 305.44 \\
  GNN Assembly~\cite{3DPartAssembly} & 2413.76 & 49.13 & 46.52 & 55.30 & 235.16 & 214.96 \\
  \textbf{\algoNameFull (\algoName)} & 266.34 & 16.32 & 13.74 & 55.22 & 234.98 & 212.57 \\
  \bottomrule
  \end{tabular}
  }
  \vspace{2.0mm}
  \subcaption{Unseen cut type geometric shape mating.}
  \label{exp:unseen-cuts-solid-objects}
  \vspace{-0.5mm}
  {
  \begin{tabular}{l|S[table-format=5.2]S[table-format=5.2]S[table-format=5.2]|S[table-format=5.2]S[table-format=5.2]S[table-format=5.2]}
  \toprule
  \rowcolor{LavenderBlue}
  Method & \text{MSE ($R$)} & \text{RMSE ($R$)} & \text{MAE ($R$)} & \text{MSE ($T$)} & \text{RMSE ($T$)} & \text{MAE ($T$)}  \\ 
  \midrule
  \rowcolor{Lightapricot}
  \multicolumn{1}{l}{Solid Shape Mating} & \multicolumn{3}{c}{} & \text{($\times10^{-3}$)} & \text{($\times10^{-3}$)} & \text{($\times10^{-3}$)} \\ 
  \midrule
  DCP~\cite{DCP}                      & 5905.92 & 76.85 & 74.02 & 88.90 & 298.16 & 274.32 \\
  GNN Assembly~\cite{3DPartAssembly}  & 2143.69 & 46.30 & 43.91 & 58.33 & 241.51 & 220.49 \\
  \textbf{\algoNameFull (\algoName)}  & 251.54 & 15.86 & 13.46 & 53.34 & 230.96 & 207.58 \\
  \bottomrule
  \end{tabular}
  }
  \end{center}
  \vspace{-4.0mm}
\end{table*}

\begin{figure*}[t]
  \begin{center}
  \mpage{0.125}{\includegraphics[width=\linewidth]{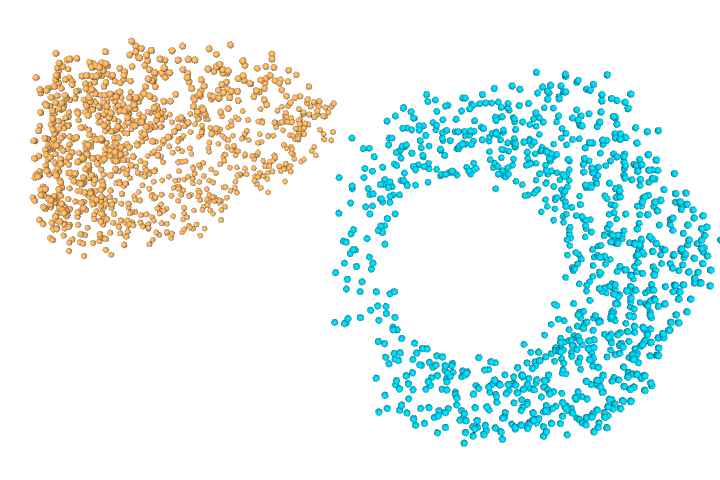}} \hfill
  \mpage{0.120}{\includegraphics[width=\linewidth]{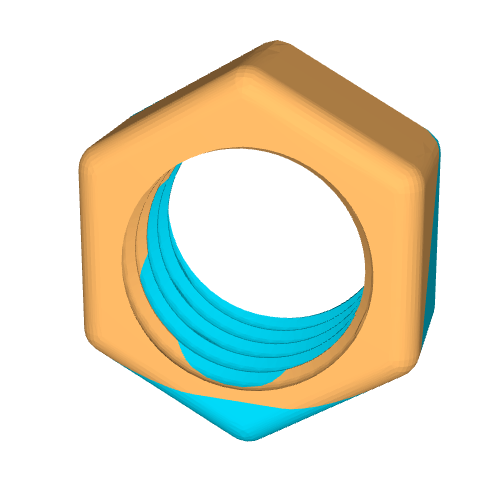}} \hfill
  \mpage{0.120}{\includegraphics[width=\linewidth]{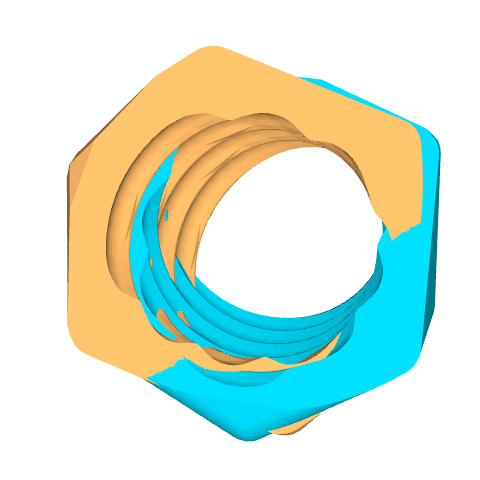}} \hfill
  \mpage{0.120}{\includegraphics[width=\linewidth]{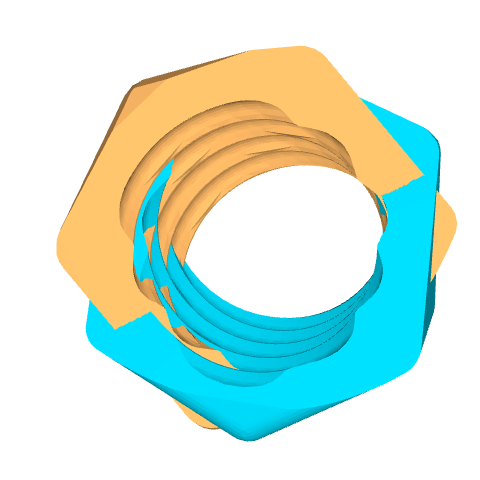}} \hfill
  \mpage{0.120}{\includegraphics[width=\linewidth]{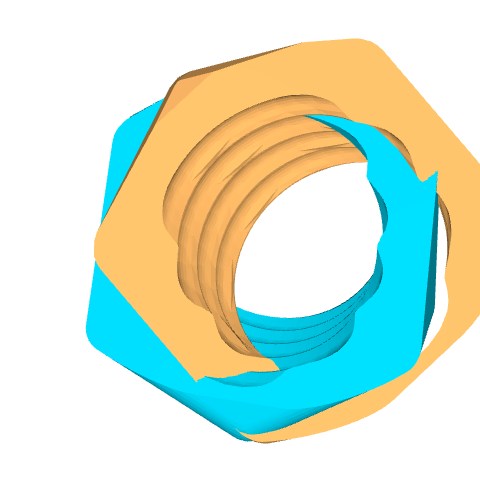}} \hfill
  \mpage{0.120}{\includegraphics[width=\linewidth]{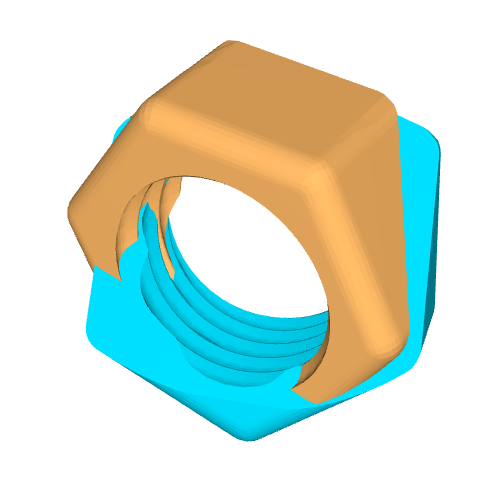}} \hfill
  \mpage{0.120}{\includegraphics[width=\linewidth]{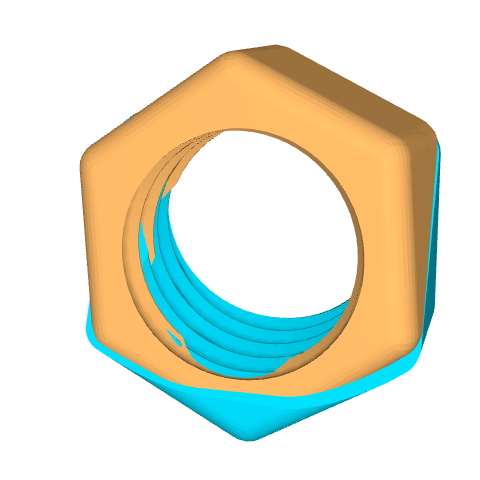}} \\
  \vspace{1.0mm}
  \mpage{0.125}{\small Input} \hfill
  \mpage{0.120}{\small Ground Truth} \hfill
  \mpage{0.120}{\small ICP} \hfill
  \mpage{0.120}{\small Sparse ICP} \hfill
  \mpage{0.120}{\small DCP} \hfill
  \mpage{0.120}{\small GNN Assembly} \hfill
  \mpage{0.120}{\small \algoName} \\
  \vspace{-1.0mm}
  \caption{
  \textbf{Visual results of noisy point cloud pairwise geometric shape mating.} 
  The two part meshes are obtained by applying a planar cut to a nut mesh.
  Gaussian noise with mean $0.0$ and standard deviation $0.05$ is added to each point.
  When noisy point clouds are presented, our method is still able to predict plausible mating configurations.
  }
  \label{fig:noisy-shape-assembly}
  \end{center}
  \vspace{-8.0mm}
\end{figure*}

\subsection{Experimental setup} \label{sec:expsetup}

\paragraph{Evaluation metrics.}
We follow the evaluation scheme from DCP~\cite{DCP}.
We compute the mean squared error (MSE), root mean squared error (RMSE), and mean absolute error (MAE) between the predicted rotation and translation values and the ground truth values.
The unit of rotation is degree.

\vspace{\paramargin}
\paragraph{Baselines.}
We compare our model with several point cloud registration methods: ICP~\citep{ICP}, 
Sparse ICP~\citep{SparseICP} and DCP~\citep{DCP} as well as a graph-based part assembly approach adapted from~\cite{3DPartAssembly}, denoted as GNN Assembly.
The three registration methods are all correspondence-based. 
That is, they approximate
correspondences between point clouds and then find poses that minimize an energy based on those correspondences.
ICP estimates correspondences as closest points and proceeds to iterate between updating
poses (from the latest correspondences) and updating correspondences (from the latest poses).
Since ICP weighs all correspondences equally, it can be thrown off by a few bad points.
Sparse ICP improves robustness to noise by downweighting outliers.
We include two variants of ICP and Sparse ICP, one computing distances point-to-point and the other point-to-plane (using ground-truth normals).
DCP is a learning-based method, which learns to compute correspondences from which a final pose is generated with SVD.
GNN Assembly is another learning-based method, but predicts rotations and translations with a message passing algorithm without correspondences (see Section~\ref{sec:rel} for more details).
In each experiment, DCP, GNN Assembly, and \algoName (our method) are all trained on the same training data.

\vspace{\paramargin}
\paragraph{Implementation details.}
We implement \algoName using PyTorch~\cite{PyTorch}.
We use the ADAM~\cite{ADAM} optimizer for training.
The learning rate is set to $1 \times 10^{-3}$ with a learning rate decay of $1 \times 10^{-6}$.
We train \algoName for 5,000 epochs using four NVIDIA P100 GPUs with 12GB memory each.
The network details are provided in Appendix~\ref{app:nsm-arch}.
We use the Open3D implementation for ICP.
The implementations of Sparse ICP\footnote{Sparse ICP: \href{https://github.com/OpenGP/sparseicp}{https://github.com/OpenGP/sparseicp}.} and DCP\footnote{DCP: \href{https://github.com/WangYueFt/dcp}{https://github.com/WangYueFt/dcp}.} are from their official GitHub repositories.
We use the codebase from \cite{3DPartAssembly} for GNN Assembly and remove the part segmentation network branch.\footnote{GNN Assembly: \href{https://github.com/AntheaLi/3DPartAssembly}{https://github.com/AntheaLi/3DPartAssembly}.}

\begin{table*}[!t]
  \begin{center}
  \scriptsize
  \caption{
  \textbf{Ablation study on \algoNameFull model design choices.}
  Testing performance with each loss removed.
  The training and test sets remain the same as in the main experiment (as presented in Table~\ref{exp:main-results}).
  }
  \vspace{-2.5mm}
  \label{exp:loss-ablation-complete}
  {
  \begin{tabular}{l|S[table-format=5.2]S[table-format=5.2]S[table-format=5.2]|S[table-format=5.2]S[table-format=5.2]S[table-format=5.2]}
  \toprule
  \rowcolor{LavenderBlue}
  Method & \text{MSE ($R$)} & \text{RMSE ($R$)} & \text{MAE ($R$)} & \text{MSE ($T$)} & \text{RMSE ($T$)} & \text{MAE ($T$)}  \\ 
  \midrule
  \rowcolor{Lightapricot}
  \multicolumn{4}{l}{\scriptsize Solid Shape Mating} & \multicolumn{3}{|c}{($\times10^{-3}$)} \\ 
  \midrule
  \algoName                                 & 94.67     & 9.73   & 7.61      & 15.48   & 124.40    & 110.44 \\ 
  \algoName w/o $\mathcal{L}_\text{adv}$ and $\mathcal{L}_\text{G}$ & 324.36    & 18.01  & 15.04     & 22.76   & 150.87    & 135.41     \\
  \algoName w/o $\mathcal{L}_\text{SDF}$  & 185.78    & 13.63  & 11.42     & 20.18   & 142.06    & 128.90    \\
  \algoName w/o $\mathcal{L}_\text{pose}$ & 7826.94   & 88.47  & 85.09     & 710.75  & 843.06    & 776.43  \\ 
  \midrule 
  \rowcolor{Lightapricot}
  \multicolumn{7}{l}{\scriptsize Shell Shape Mating} \\ 
  \midrule 
  \algoName                                 & 290.02    & 17.03  & 14.52      & 107.60 & 328.03 & 301.19 \\
  \algoName w/o $\mathcal{L}_\text{adv}$ and $\mathcal{L}_\text{G}$ & 446.90    & 21.14  & 18.63      & 133.83 & 365.83 & 341.07 \\
  \algoName w/o $\mathcal{L}_\text{SDF}$  & 418.61    & 20.46  & 17.44      & 111.74 & 334.27 & 311.07 \\
  \algoName w/o $\mathcal{L}_\text{pose}$ & 9374.11   & 96.82  & 93.02      & 593.90 & 770.65 & 689.44 \\
  \bottomrule
  \end{tabular}
  }
  \end{center}
  \vspace{-8.0mm}
\end{table*}

\subsection{Performance evaluation and comparisons}

\subsubsection{Comparison to existing approaches}

We compare the performance of our method with existing approaches on pairwise 3D geometric shape mating.
In this evaluation, we use 80\% of the shape pairs for training, 10\% for validation and the remaining 10\% for testing (metrics are reported on this holdout set).
Table~\ref{exp:main-results} presents results for both solid and shell shape mating settings.
Figure~\ref{fig:geometric-shape-assembly-results} presents a visual comparison between methods.
Figure~\ref{fig:app-main} presents more visual comparisons.

Quantiatively, results in both settings follow a similar pattern.
NSM achieves the best rotation MSE by an order of magnitude.
For translation prediction, NSM and GNN Assembly both achieve strong results.

\vspace{\paramargin}
\paragraph{Point cloud registration methods.}
NSM outperforms registration methods by a large margin in rotation prediction.
This may be surprising as shape mating and point cloud registration are similar problems.
In fact, shape mating reduces to point cloud alignment given an interface segmentation.
Despite this, these results suggest that existing point cloud registration methods are insufficient for the shape mating task.
In our qualitative results, we can see registration methods often attempt to overlay pieces rather than mating them together
and this matches our hypothesis that the inferior performance of registration methods is due to their correspondence assumptions.
In point cloud registration, it is assumed that the inputs correspond usually to a rigid transformation and some observation noise.
Even with outlier handling, they are unable to leave the non-interface portion of the surface out of correspondence
in order to precisely align the interface portions.
More surprisingly, this may be true even for learning-based methods like DCP, where the interpolation of correspondences
may force consideration of non-interface points.
These results highlight that shape mating is a distinct problem from registration, requiring more specialized method design.

\vspace{\paramargin}
\paragraph{Part assembly.}
NSM outperforms GNN Assembly on rotation prediction and performs similarly on translation prediction.
The GNN Assembly architecture is designed for the part assembly task where semantic cues are available
and fine-grained geometric details are not as important for alignment.
We hypothesize that our adversarial loss for learning shape priors and the Transformer architecture for capturing cross-shape information
are better suited to the \emph{geometric} shape mating task which relies on these details.

These results support our conviction that the proposed task is distinct from point cloud registration and part assembly, and that progress will require further investigation into the \emph{geometric} shape mating problem specifically.

\subsubsection{Generalization to unseen categories and cut types}

\paragraph{Unseen category evaluations.}
To test the generalization across categories, we test on the box and bag categories and train on the remaining 9 categories.
Table~\ref{exp:unseen-categories-solid-objects} presents the results of the solid shape mating setting.
Notably, \algoName is category agnostic and relies mainly on aligning surface geometry details rather than class-specific semantic cues.
We expect strong generalization. 
Compared to in-category testing, while the performance degrades slightly, \algoName still performs favorably against existing methods.

\vspace{\paramargin}
\paragraph{Unseen cut type evaluations.}
To test the generalization across different cut types, we test on the parabolic cuts and train on the remaining 4 cut types.
Table~\ref{exp:unseen-cuts-solid-objects} presents the results of the solid shape mating setting.
As with unseen cut types, the performance degrades for all methods, while \algoName still achieves the best results.

\subsubsection{Evaluation on noisy point clouds}

Real-world point cloud data, e.g., captured by depth cameras, contains measurement error that the point clouds in our training set do not.
For our framework to be applicable to real-world problems, it must be robust to noise in the point cloud observations.
To test robustness to noise, we train and test each model on a noise-augmented version of our dataset.
Gaussian noise with mean $0.0$ and standard deviation $0.05$ is added to each point.
As can be seen in Figure~\ref{fig:noisy-shape-assembly}, while the performance of all methods, including ours, does decline, NSM is still able to predict reasonable mating poses.
The performance of correspondence-based methods (ICP, Sparse ICP, and even learning-based DCP) all show large drops in performance.

\subsubsection{Ablation study: Contribution of loss functions}

To evaluate our design choices, we conduct an ablation study by removing one loss function at a time.
Table~\ref{exp:loss-ablation-complete} presents the results of both solid and shell shape mating settings.
The training and test sets remain the same as in the main experiment (as presented in Table \ref{exp:main-results}).
Performance declines significantly without adversarial learning (i.e., without $\mathcal{L}_\text{adv}$ and $\mathcal{L}_\text{G}$), confirming
our hypothesis that adversarial learning can serve as a pose refinement
or regularizer and improve predictions even when ground truth is available.
Performance also declines without learning implicit shape reconstruction (i.e., without $\mathcal{L}_\text{SDF}$), suggesting that there are useful synergies between shape mating and geometry reconstruction.
Without the pose loss $\mathcal{L}_\text{pose}$, the model does not learn shape mating at all, which suggests
adversarial training with implicit shape reconstruction alone is not sufficient.

\section{Conclusions}

This paper introduces a new problem with broad applications, an insightful procedural data generation method, and an algorithmic architecture designed for the proposed task.
The self-supervised data generation pipeline allows NSM to learn shape mating without ground truth.
Since NSM learns to align geometric features rather than semantic ones, it is able to generalize across categories and across surface cut types.
Experimental results show that NSM predicts plausible mating configurations and outperforms all competing methods.
An ablation study suggests that the novel adversarial training scheme significantly improves performance (even though ground truth is available) and the performance benefits of an auxiliary implicit shape reconstruction task suggest synergies between shape reconstruction and shape mating.
We hope that this paper can convincingly establish geometric shape mating as a meaningful task, distinct from semantic part assembly.
Pairwise geometric shape mating is a core task to solve multi-step reasoning required for assembling parts to form an object.
Natural extensions of NSM would go beyond pairwise shape mating to consider the problem of mating multiple parts.

{\small
\vspace{\paramargin}
\paragraph{Acknowledgments.}
We thank Ziyi Wu, Yuliang Zou, Shih-Yang Su and Tsun-Yi Yang for providing feedback to early draft.
Alec Jacobson is supported by Canada Research Chairs Program and gifts by Adobe Systems.
Animesh Garg is supported by CIFAR AI Chair, NSERC Discovery Award, University of Toronto XSeed award, and gifts from LG. 
}

\appendix

\section*{Appendix}

\section{Neural Shape Mating Model Details} \label{app:nsm-arch}

\paragraph{Point cloud encoder.} 
We follow DCP~\cite{DCP} and adopt DGCNN~\cite{DGCNN} as our point cloud encoder $E$.
The point cloud encoder $E$ consists of one $k$ nearest neighbor layer and five convolution layers.
In our work, $k$ is set to $20$ as in DCP~\cite{DCP}.
The numbers of channels for each convolution layer are $64$, $64$, $128$, $256$, and $1,024$.
Each convolution layer is followed by a batch normalization layer and a LeakyReLU activation function with a negative solpe of $0.2$.

\vspace{\paramargin}
\paragraph{Transformer network.} 
The Transformer network consists of an encoder and a decoder.
Both the encoder and the decoder consist of one attention module, respectively.
Each attention module is composed of three fully connected layers $Q$, $K$ and $V$ for encoding the input feature to the query $q$, the key $k$, and the value $v$, respectively.
Each fully connected layer has an output size of $1,024$ and is followed by a ReLU activation and a layer normalization layer~\cite{ba2016layer}.

\vspace{\paramargin}
\paragraph{Regressor.} 
Our regressor consists of one fully connected layer shared between the quaternion (rotation) prediction head and the translation prediction head, and two fully connected layers, one for predicting quaternion and the other for predicting translation.
The shared fully connected layer has an output size of $256$ and is followed by a batch normalization layer and a LeakyReLU activation function with a negative slope of $0.2$.
The fully connected layer in the quaternion prediction head has an output size of $3$.
We apply an L2 normalization to the output of the quaternion prediction head.
The fully connected layer in the translation prediction head has an output size of $3$.

\vspace{\paramargin}
\paragraph{Discriminator.} 
Our discriminator $D$ contains a DGCNN~\cite{DGCNN} network (the same as that in the point cloud encoder $E$) and a fully connected layer.
The fully connected layer has an output size of $1$ and is followed by a sigmoid activation function.
We first pass the predicted mating configuration $P_\text{pred}$ and the ground-truth mating configuration $P_\text{GT}$ to the DGCNN network to encode point features, respectively.
We then apply a max pooling layer to aggregate the point features of $P_\text{pred}$ to derive the shape feature $F_\text{pred}$.
The shape feature $F_\text{GT}$ for $P_\text{GT}$ can be similarly derived.
Next, the fully connected layer takes as input the shape features and predicts whether the input shape features look visually realistic like an object or not.

\vspace{\paramargin}
\paragraph{SDF network.} 
We follow DeepSDF~\cite{DeepSDF} and use eight fully connected layers for our SDF network.
The first seven fully connected layers have an output size of $256$ and are all followed by a batch normalization layer and a ReLU activation function.
The last fully connected layer has an output size of $1$.
Same as DeepSDF~\cite{DeepSDF}, we also use a skip connection that bypasses the input to the fifth layer.
The bypassed feature and the output of the fourth layer are concatenated and then become the input to the fifth layer.

\begin{figure*}[t]
  \centering
  \includegraphics[width=0.7\linewidth]{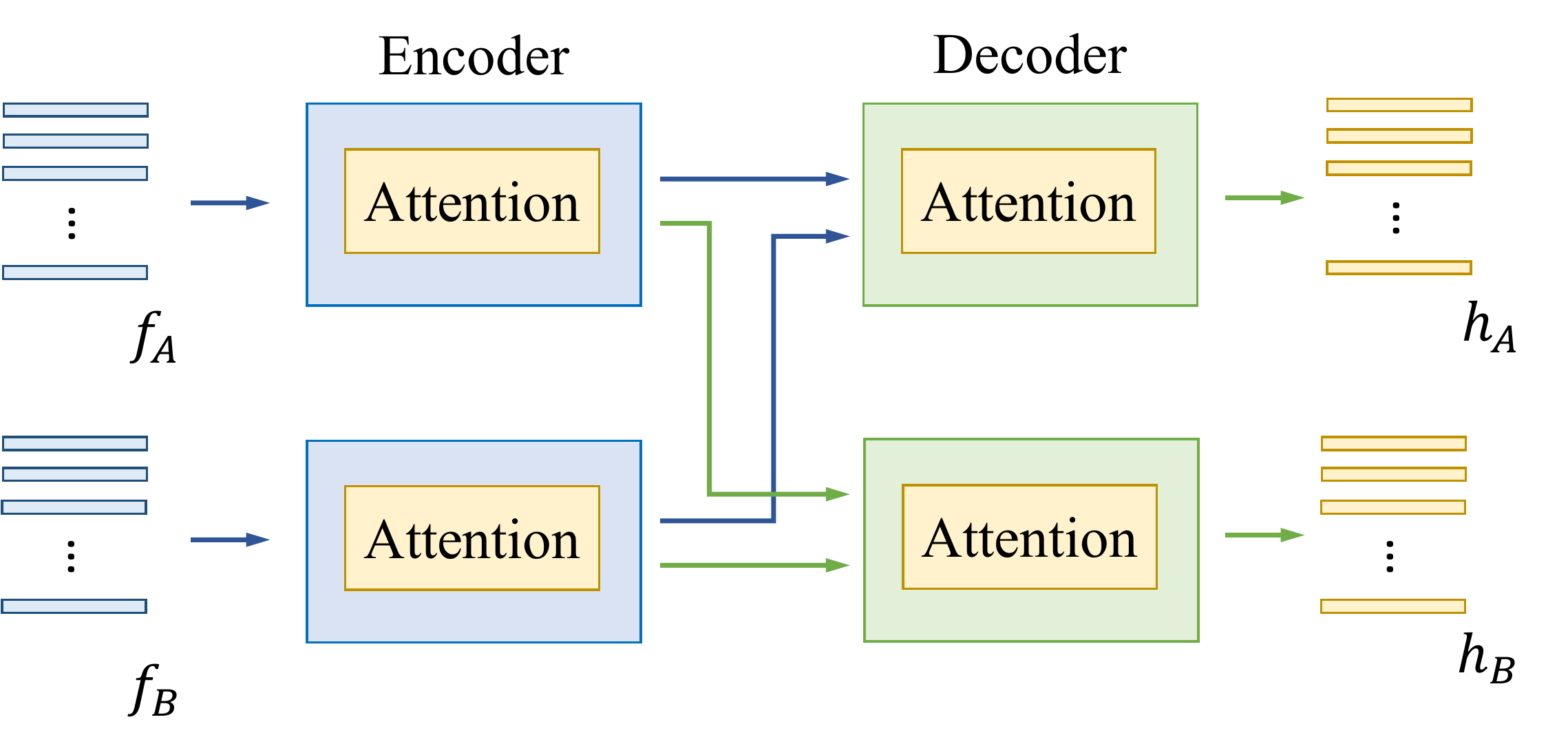}
  \vspace{-3.0mm}
  \caption{
  \textbf{Overview of Transformer.} 
  The Transformer network consists of an encoder and a decoder.
  }
  \label{fig:transformer}
\end{figure*}

\begin{figure*}[t]
  \centering
  \includegraphics[width=0.6\linewidth]{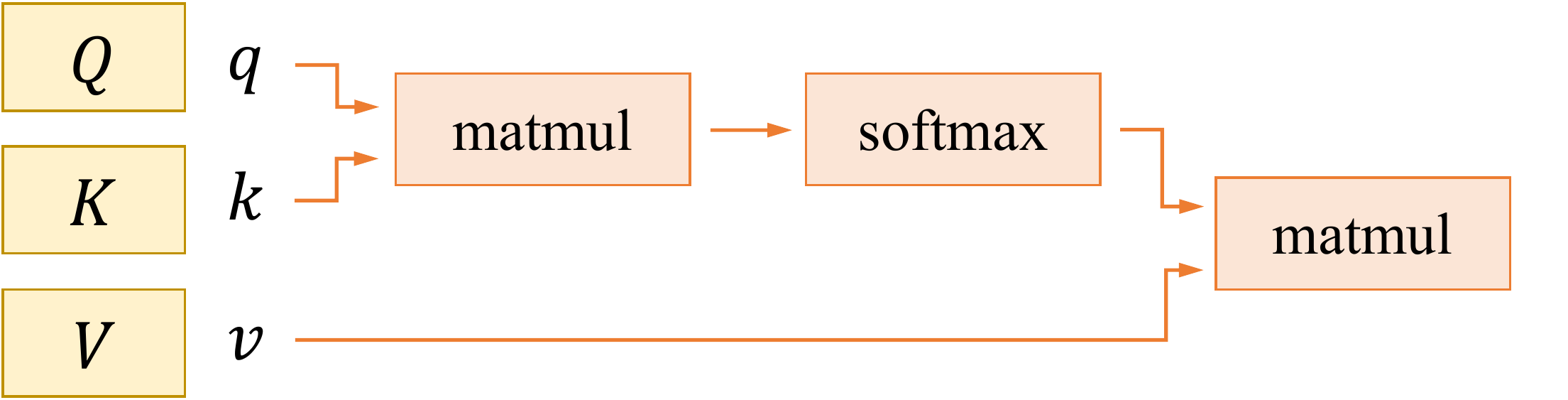}
  \caption{
  \textbf{Overview of the attention module.} 
  The attention module consists of three fully connected layers $Q$, $K$ and $V$ for encoding the input to the quert $q$, the key $k$, and the value $v$, respectively.
  }
  \label{fig:attention}
  \vspace{-4.0mm}
\end{figure*}

\section{Details of Transformer} \label{app:transformer}

Since our task requires reasoning about the fit between the two input point clouds for pose estimation, we adopt a Transformer network~\cite{Transformer} as our feature correlation module, as it allows our model to learn asymmetric cross-shape information.
Figure~\ref{fig:transformer} and Figure~\ref{fig:attention} present the overview of the Transformer network and the attention module, respectively.

Given the point feature $f_A = \{f_i^A\}_{i=1}^N$ of point cloud $P_A$ and the point feature $f_B = \{f_j^B\}_{j=1}^M$ of point cloud $P_B$ as input, we first encode intra-shape information for each point cloud by passing the point features $f_A$ and $f_B$ to the encoder in the Transformer network, respectively.
Specifically, we first compute the query $q_A$, the key $k_A$, and the value $v_A$ for point cloud $P_A$ by
\begin{equation}
    q_A = Q_E(f_A),
\end{equation}
\begin{equation}
    k_A = K_E(f_A), \text{ and}
\end{equation}
\begin{equation}
    v_A = V_E(f_A),
\end{equation}
where $Q_E$, $K_E$ and $V_E$ denote the fully connected layers in the attention module of the encoder in the Transformer network for computing the query, the key, and the value, respectively.
The query $q_B$, the key $k_B$, and the value $v_B$ for point cloud $P_B$ can be similarly derived.

Then, to encode intra-shape information, we compute the feature $s_A$ for point cloud $P_A$ by
\begin{equation}
  s_A = \text{Attention}(q_A, k_A, v_A) = \text{softmax}(\frac{q_Ak_A^\top}{\sqrt{d}})v_A,
\end{equation}
where $d$ is the dimension of the point features $f_i^A$ and $f_j^B$ (which is $1,024$ in this work).
Similarly, the feature $s_B$ for point cloud $P_B$ can be computed by $s_B = \text{Attention}(q_B, k_B, v_B)$.

The attention module in the encoder of the Transformer network allows the model to capture intra-shape information, which is encoded in the feature $s_A$ for point cloud $P_A$ and in the feature $s_B$ for point cloud $P_B$.

Next, we have a decoder in the Transformer network that takes as input features $s_A$ and $s_B$ and outputs features $h_A$ and $h_B$ for point clouds $P_A$ and $P_B$, which encode cross-shape information for pose estimation, respectively.
Specifically, we first compute the query $q_{s_A}$, the key $k_{s_A}$, and the value $v_{s_A}$ for point cloud $P_A$ by
\begin{equation}
    q_{s_A} = Q_D(s_A),
\end{equation}
\begin{equation}
    k_{s_A} = K_D(s_A), \text{ and}
\end{equation}
\begin{equation}
    v_{s_A} = V_D(s_A),
\end{equation}
where $Q_D$, $K_D$ and $V_D$ denote the fully connected layers in the attention module of the decoder in the Transformer network for computing the query, the key, and the value, respectively.
The query $q_{s_B}$, the key $k_{s_B}$, and the value $v_{s_B}$ for point cloud $P_B$ can be similarly derived.

To encode cross-shape information, we compute the feature $h_A$ for point cloud $P_A$ by
\begin{equation}
  h_A = \text{Attention}(q_{s_A}, k_{s_B}, v_{s_B}) = \text{softmax}(\frac{q_{s_A}k_{s_B}^\top}{\sqrt{d}})v_{s_B},
\end{equation}
where $d$ is the dimension of the point features $f_i^A$.
Similarly, the feature $h_B$ for point cloud $P_B$ can be derived by $h_B = \text{Attention}(q_{s_B}, k_{s_A}, v_{s_A})$.

The attention module in the decoder jointly considers the feature $s_A$ from point cloud $P_A$ and the feature $s_B$ from point cloud $P_B$ and outputs features $h_A$ and $h_B$ that encode cross-shape information for point clouds $P_A$ and $P_B$, respectively.

The features $h_A$ and $h_B$ are then passed to the regressor $R$ for predicting poses for point clouds $P_A$ and $P_B$, respectively.

\begin{table*}[t]
  \begin{center}
  \scriptsize
  \caption{
  \textbf{Dataset statistics.}
  We summarize the number of shape pairs of the Geometric Shape Mating dataset.
  Our dataset contains a large number shape pairs, covering a diverse combination of different shape types, object categories, and cut types.
  }
  \vspace{-2.0mm}
  \label{dataset-stats}
  \centering
  {
  \begin{tabular}{lc|rrrrrr|rrrrrr}
  \toprule
  \multirow{2}{*}{Category} & \multicolumn{1}{c|}{Number of} & \multicolumn{6}{c|}{Solid shape pairs} & \multicolumn{6}{c}{Shell shape pairs} \\
  \cmidrule{3-14}
  & \multicolumn{1}{c|}{objects} & Plane & Parabola & Sine & Square & Pulse & All & Plane & Parabola & Sine & Square & Pulse & All \\
  \midrule
  Bag & 28 & 500 & 500 & 500 & 500 & 500 & 2,500 & 480 & 480 & 480 & 480 & 480 & 2,400 \\
  Bowl & 157 & 2,800 & 2,800 & 2,400 & 2,400 & 2,400 & 12,800 & 2,100 & 2,100 & 2,100 & 2,100 & 2,100 & 10,500 \\
  Box & 191 & 3,600 & 3,750 & 2,750 & 2,800 & 3,150 & 16,050 & 5,200 & 4,600 & 4,800 & 4,200 & 4,200 & 23,000 \\
  Hat & 16 & 320 & 320 & 320 & 320 & 320 & 1,600 & 280 & 320 & 320 & 280 & 280 & 1,480 \\
  Jar & 106 & 2,800 & 3,200 & 3,000 & 2,800 & 2,800 & 14,600 & 2,400 & 2,600 & 2,600 & 1,800 & 1,800 & 11,200 \\
  Mug & 71 & 1,800 & 2,200 & 1,900 & 1,900 & 1,600 & 9,400 & 2,100 & 2,200 & 1,800 & 1,600 & 1,500 & 9,200 \\
  Plate & 35 & 980 & 980 & 960 & 960 & 920 & 4,800 & 840 & 860 & 860 & 780 & 750 & 4,090 \\
  Shoe & 168 & 3,100 & 3,200 & 2,800 & 2,800 & 3,200 & 15,100 & 2,400 & 2,800 & 2,400 & 2,100 & 1,800 & 11,500 \\
  Sofa & 200 & 2,100 & 1,800 & 2,150 & 1,620 & 1,780 & 9,450 & 1,200 & 1,600 & 2,100 & 1,750 & 1,900 & 8,550 \\
  Table & 196 & 2,500 & 2,200 & 2,100 & 1,600 & 1,800 & 10,200 & 2,200 & 1,600 & 1,550 & 1,350 & 1,750 & 8,450 \\
  Toy & 78 & 1,650 & 1,650 & 1,650 & 1,450 & 1,450 & 7,850 & 1,520 & 1,350 & 1,480 & 1,260 & 1,260 & 6,870 \\
  All & 1,246 & 22,150 & 22,600 & 20,530 & 19,150 & 19,920 & 104,350 & 20,720 & 20,510 & 20,490 & 17,700 & 17,820 & 97,240 \\
  \bottomrule
  \end{tabular}
  }
  \end{center}
  \vspace{-6.0mm}
\end{table*}

\section{Details of Adversarial Training} \label{app:adv-training}

We adopt an adversarial training scheme to train our model.
Specifically, the model training process consists of two alternating phases: 
1) training the generator (i.e., the point cloud encoder $E$, the pose estimation network and the SDF network $F$) with the parameters of the discriminator being fixed and
2) training the discriminator $D$ with the parameters of the generator being fixed.

\vspace{\paramargin}
\paragraph{Training the generator.} 
In this phase, we train the generator with the parameters of the discriminator being fixed.
Specifically, we use the pose prediction loss $\mathcal{L}_\text{pose}$ (Equation~\eqref{eq:pose}), the loss $\mathcal{L}_\text{G}$ (Equation~\eqref{eq:G}), and the SDF regression loss $\mathcal{L}_\text{SDF}$ (Equation~\eqref{eq:sdf}) to train the generator.

\vspace{\paramargin}
\paragraph{Training the discriminator.} 
In this phase, we train the discriminator $D$ with the parameters of the generator being fixed.
Specifically, we use the adversarial loss $\mathcal{L}_\text{adv}$ (Equation~\eqref{eq:adv}) to train the discriminator.

\section{Dataset Details} \label{app:dataset-stats}

\paragraph{Planar function.}
\begin{equation*}
    z = ax + by + c.
\end{equation*}

For each shape pair, we randomly sample a set of numbers for coefficients $a$, $b$, and $c$, subject to 
\begin{equation*}
    10 \geq a \geq -10,
\end{equation*}
\begin{equation*}
    10 \geq b \geq -10, \text{ and}
\end{equation*}
\begin{equation*}
    1 \geq c \geq -1.
\end{equation*}

\vspace{-6.0mm}
\paragraph{Sine function.} 
\begin{equation*}
    z = h \sin(ax + by + c) + k.
\end{equation*}

For each shape pair, we randomly sample a set of numbers for coefficients $a$, $b$, $c$, $h$, and $k$, subject to 
\begin{equation*}
    100 \geq a \geq -100,
\end{equation*}
\begin{equation*}
    100 \geq b \geq -100,
\end{equation*}
\begin{equation*}
    1 \geq c \geq -1,
\end{equation*}
\begin{equation*}
    1 \geq h \geq -1, \text{ and}
\end{equation*}
\begin{equation*}
   1 \geq k \geq -1.
\end{equation*}

\vspace{-6.0mm}
\paragraph{Parabolic function.} 
\begin{equation*}
    z = ax^2 + by^2 + c.
\end{equation*}

For each shape pair, we randomly sample a set of numbers for coefficients $a$, $b$, and $c$, subject to
\begin{equation*}
    10 \geq a \geq -10,
\end{equation*}
\begin{equation*}
    10 \geq b \geq -10, \text{ and}
\end{equation*}
\begin{equation*}
    1 \geq c \geq -1.
\end{equation*}

\vspace{-6.0mm}
\paragraph{Square function.} 
\begin{equation*}
  z = 
  \begin{cases}
    h, \ \text{if } t \geq x \geq -t, \\
    0, \ \text{otherwise}.
  \end{cases}
\end{equation*}

For each shape pair, we randomly sample a set of numbers for coefficients $t$ and $h$, subject to
\begin{equation*}
    1 \geq t > 0 \text{ and}
\end{equation*}
\begin{equation*}
    1 \geq h > 0.
\end{equation*}

\vspace{-6.0mm}
\paragraph{Pulse function.} 
\begin{equation*}
  z = 
  \begin{cases}
    h, \ \text{if } t \geq x \geq -t \text{ and } t \geq y \geq -t, \\
    0, \ \text{otherwise}.
  \end{cases}
\end{equation*}

For each shape pair, we randomly sample a set of numbers for coefficients $t$ and $h$, subject to
\begin{equation*}
    1 \geq t > 0 \text{ and}
\end{equation*}
\begin{equation*}
    1 \geq h > 0.
\end{equation*}

\vspace{\paramargin}
\paragraph{Mesh cutting time.}
The average time for generating a solid shape pair is around 18 seconds.
The average time for generating a shell/hollow shape pair is around 24 seconds.
We note that while our data generation scheme can be applied at each training iteration to generate new shape pairs, for the sake of efficiency, we choose to collect a dataset and train the model on the generated dataset with random poses applied on the fly at each training iteration.

\vspace{\paramargin}
\paragraph{Dataset statistics.}
We summarize the statistics of the Geometric Shape Mating dataset in Table~\ref{dataset-stats}.

\section{Dataset Visual Examples} \label{app:dataset-visual-results}

\begin{figure*}[t]
  \centering
  \includegraphics[width=\linewidth]{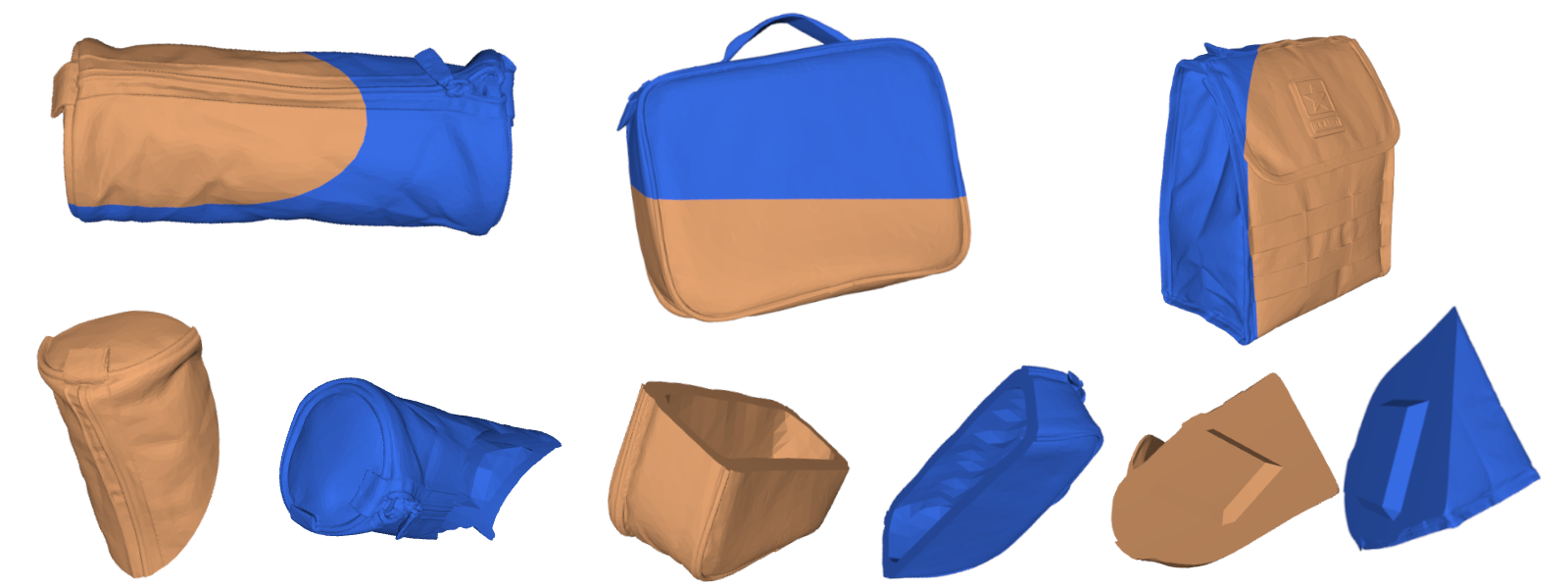} \\
  \vspace{10.0mm}
  \includegraphics[width=\linewidth]{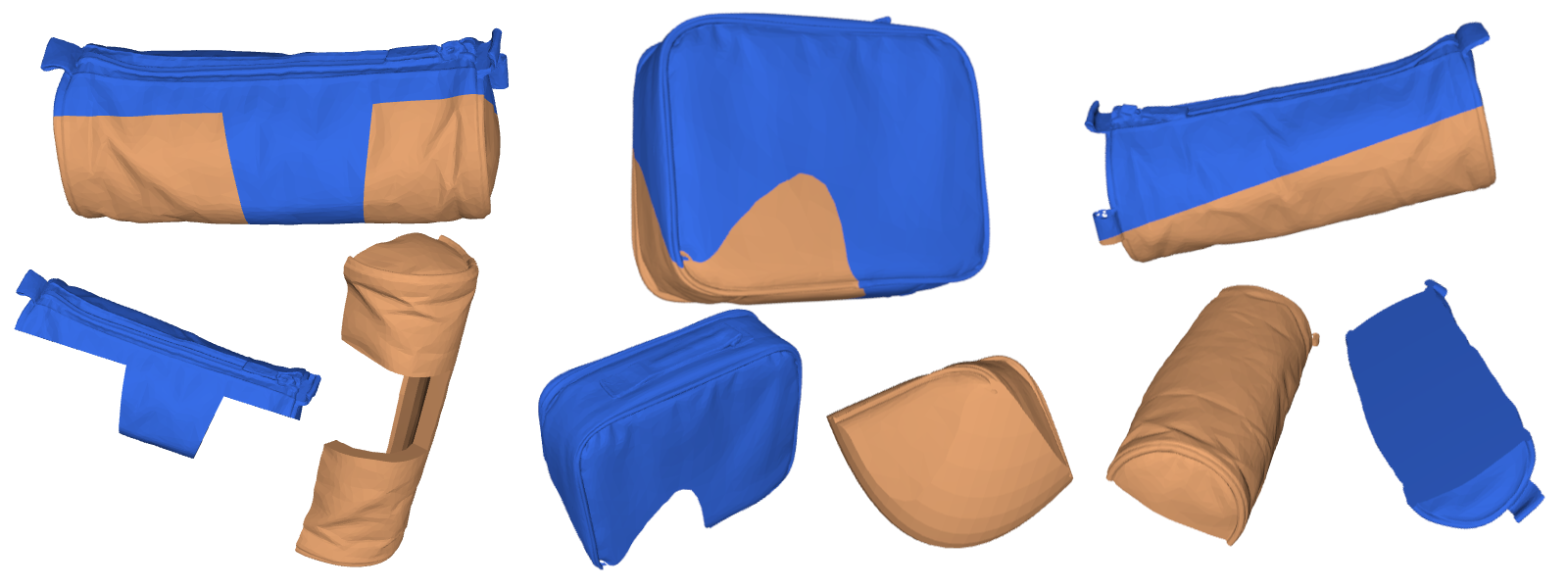} \\
  \vspace{10.0mm}
  \includegraphics[width=\linewidth]{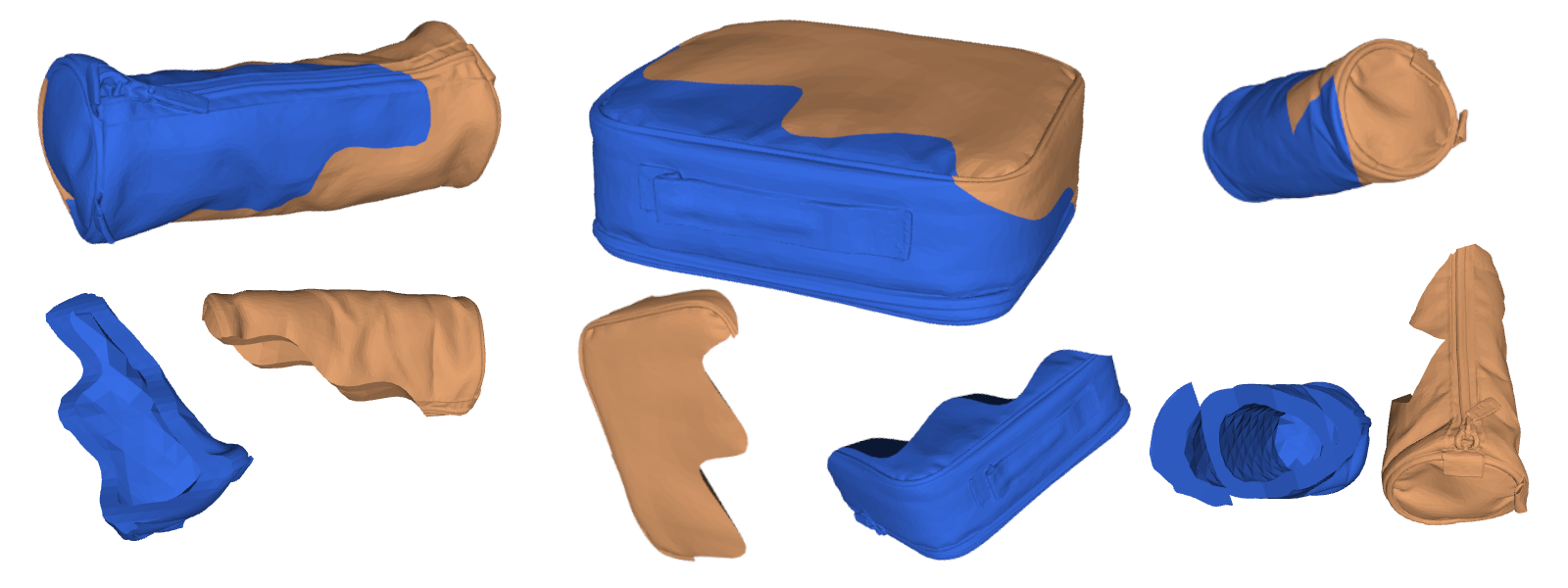}
  \caption{
  \textbf{Dataset visual examples.} 
  We present visual examples of the shape pairs in the bag category.
  }
  \label{fig:dataset-bag}
\end{figure*}

\begin{figure*}[t]
  \centering
  \includegraphics[width=\linewidth]{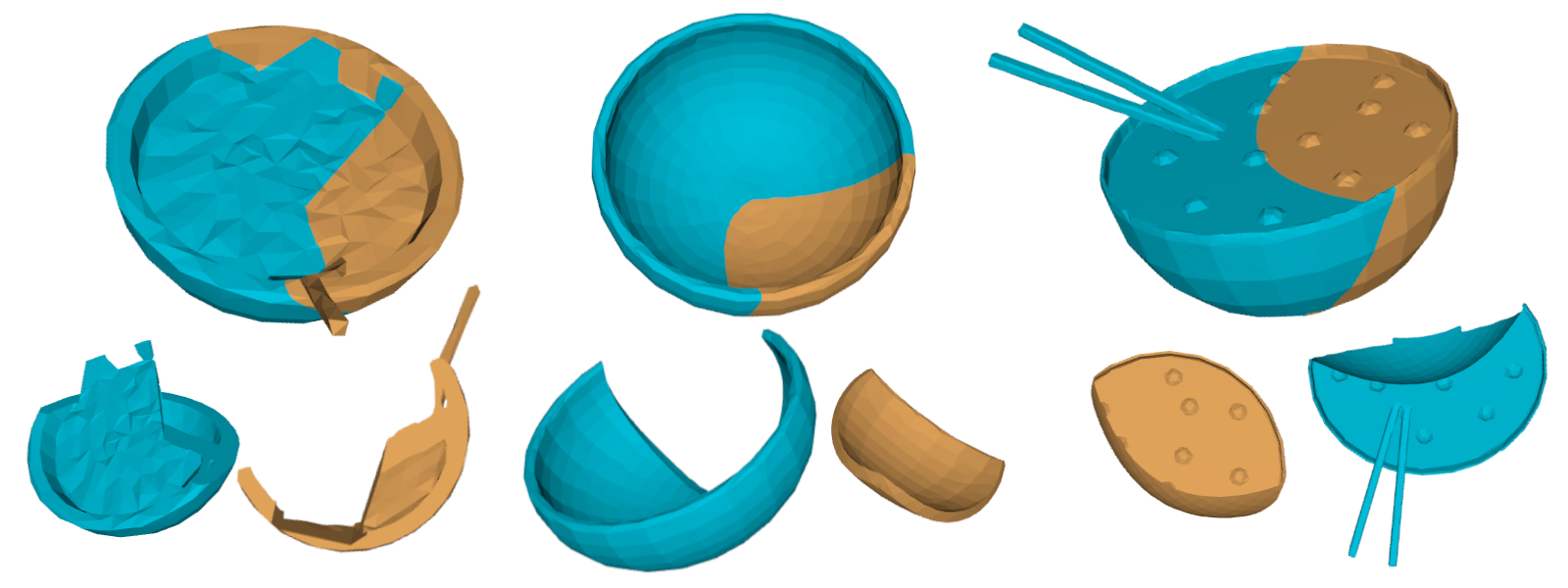} \\
  \vspace{10.0mm}
  \includegraphics[width=\linewidth]{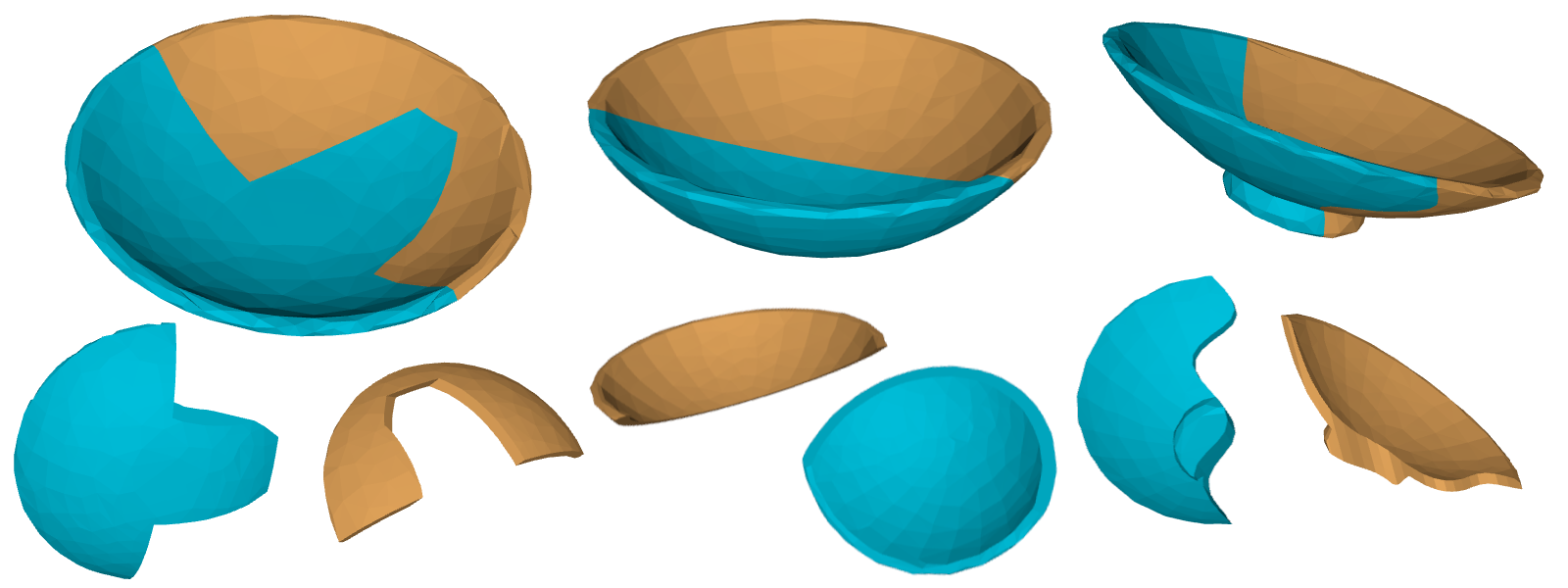} \\
  \vspace{10.0mm}
  \includegraphics[width=\linewidth]{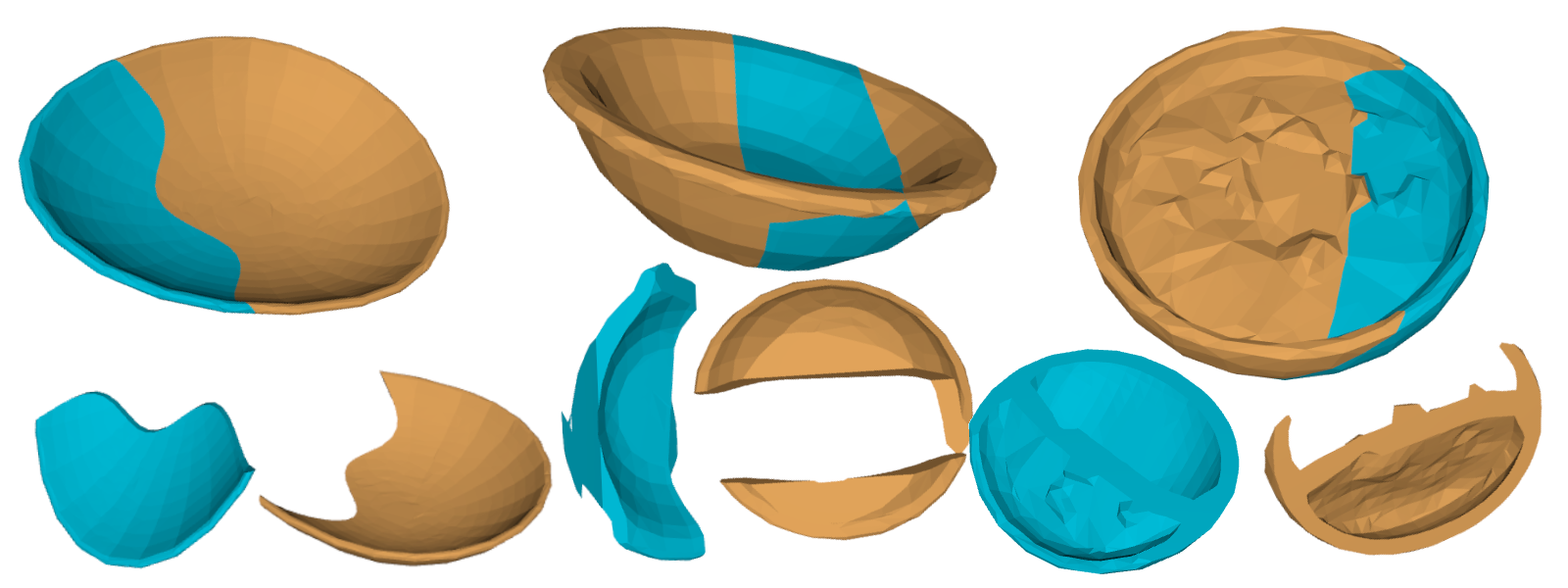}
  \caption{
  \textbf{Dataset visual examples.} 
  We present visual examples of the shape pairs in the bowl category.
  }
  \label{fig:dataset-bowl}
\end{figure*}

\begin{figure*}[t]
  \centering
  \includegraphics[width=\linewidth]{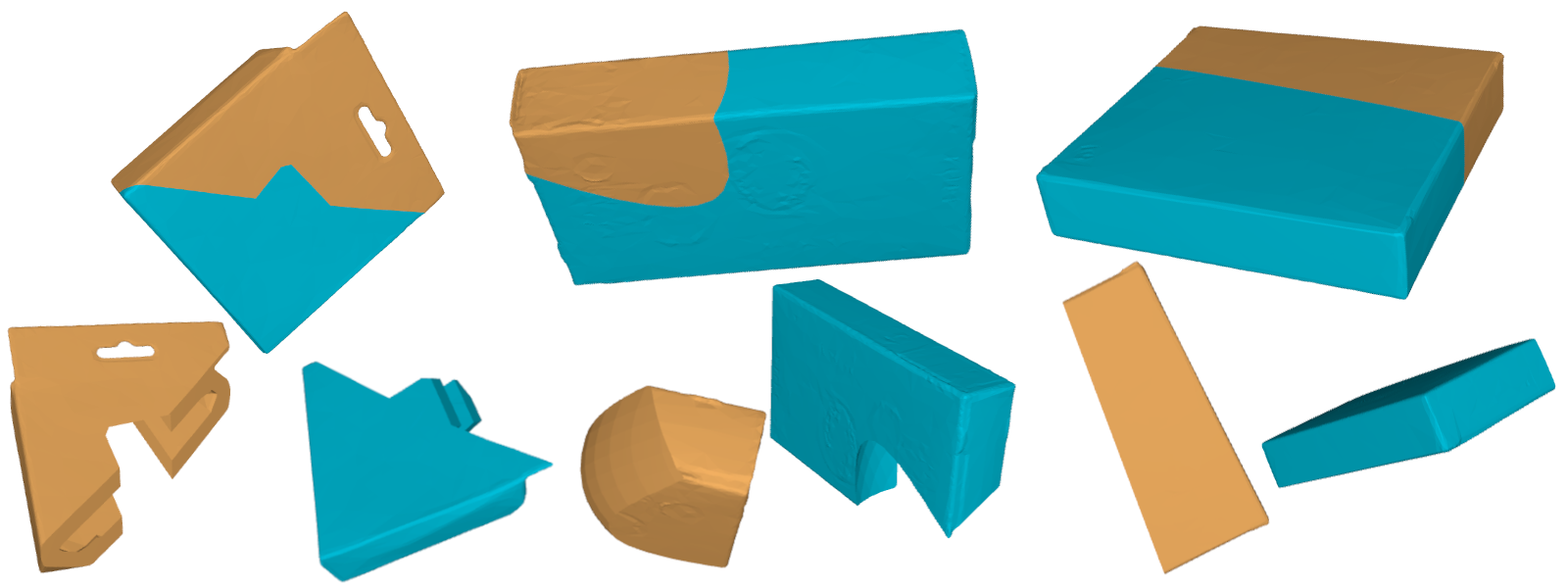} \\
  \vspace{10.0mm}
  \includegraphics[width=\linewidth]{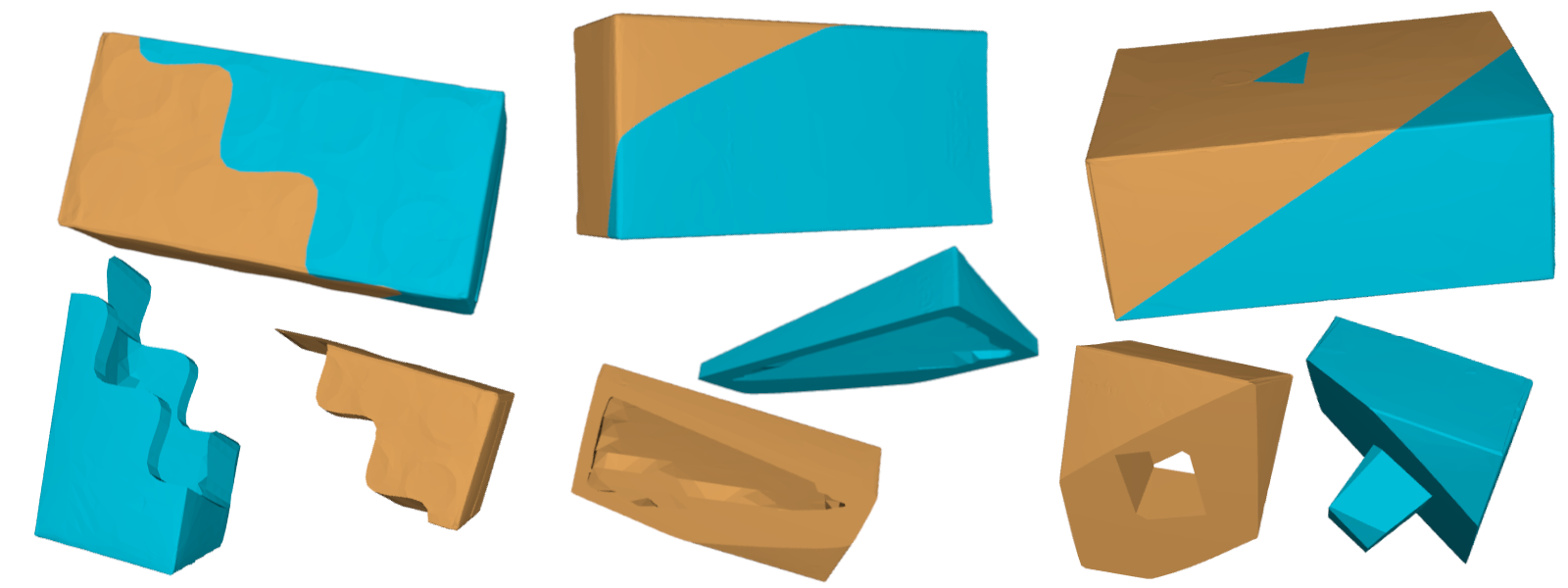} \\
  \vspace{10.0mm}
  \includegraphics[width=\linewidth]{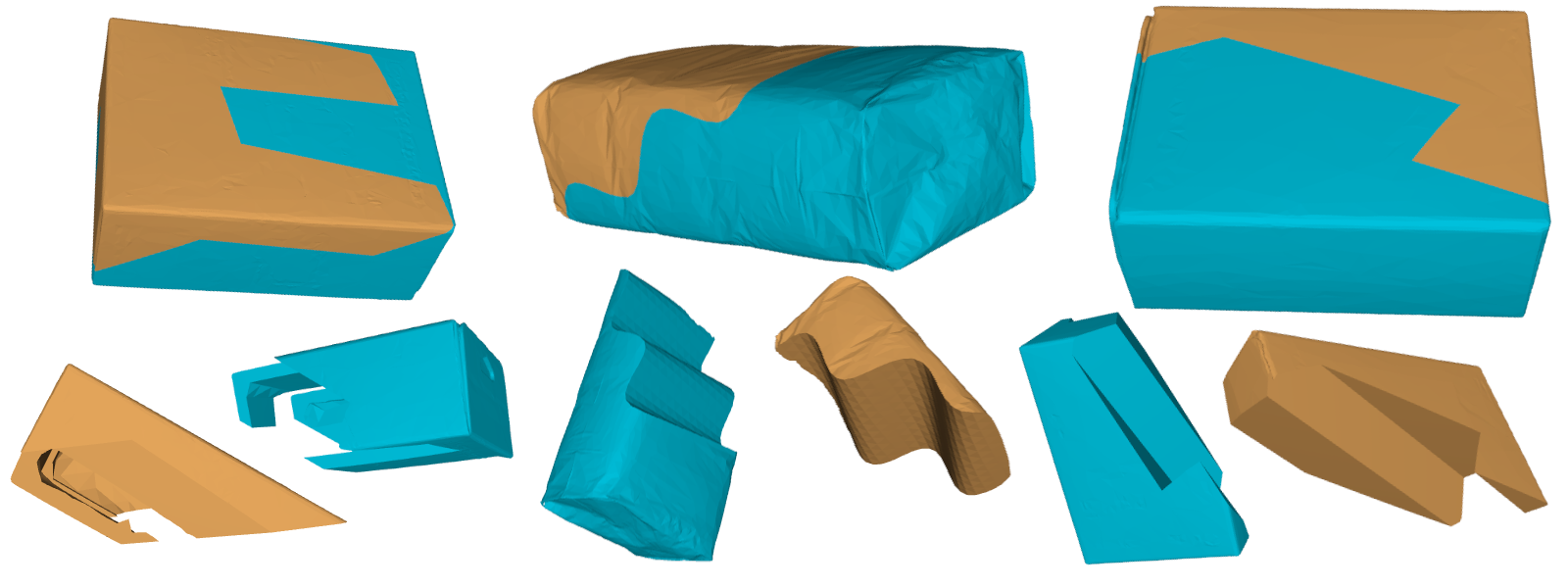}
  \caption{
  \textbf{Dataset visual examples.} 
  We present visual examples of the shape pairs in the box category.
  }
  \label{fig:dataset-box}
\end{figure*}

\begin{figure*}[t]
  \centering
  \includegraphics[width=\linewidth]{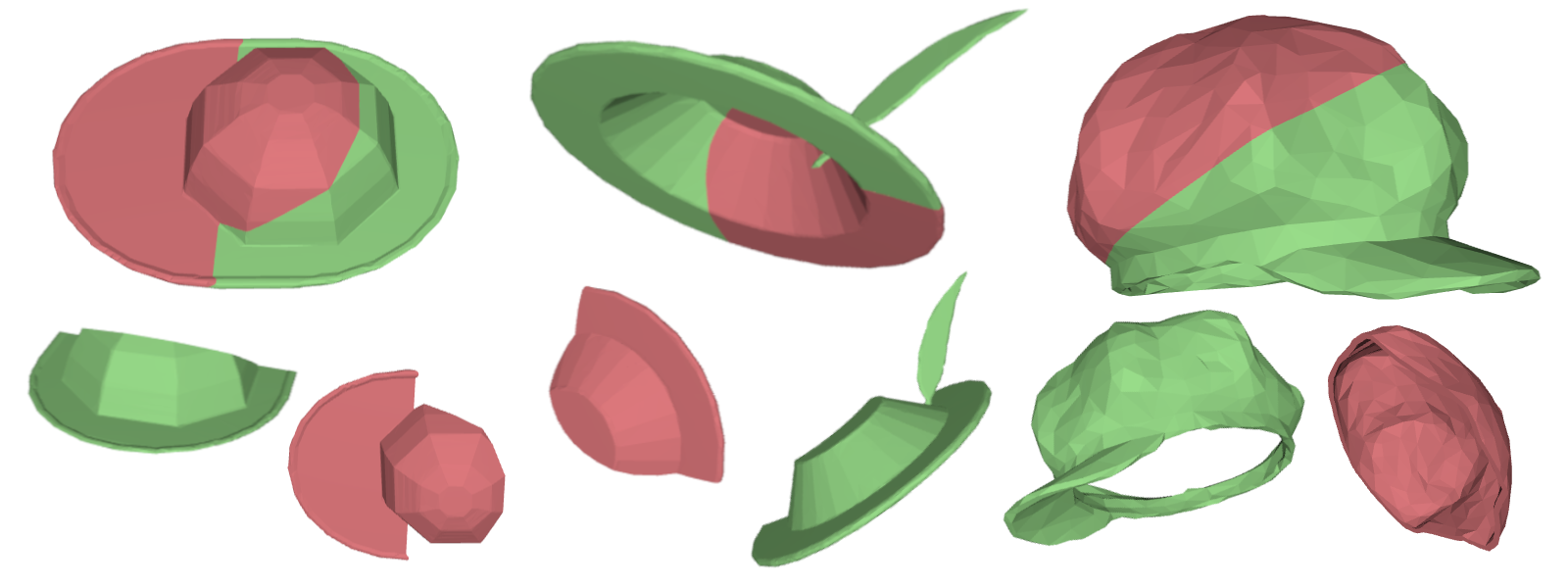} \\
  \vspace{10.0mm}
  \includegraphics[width=\linewidth]{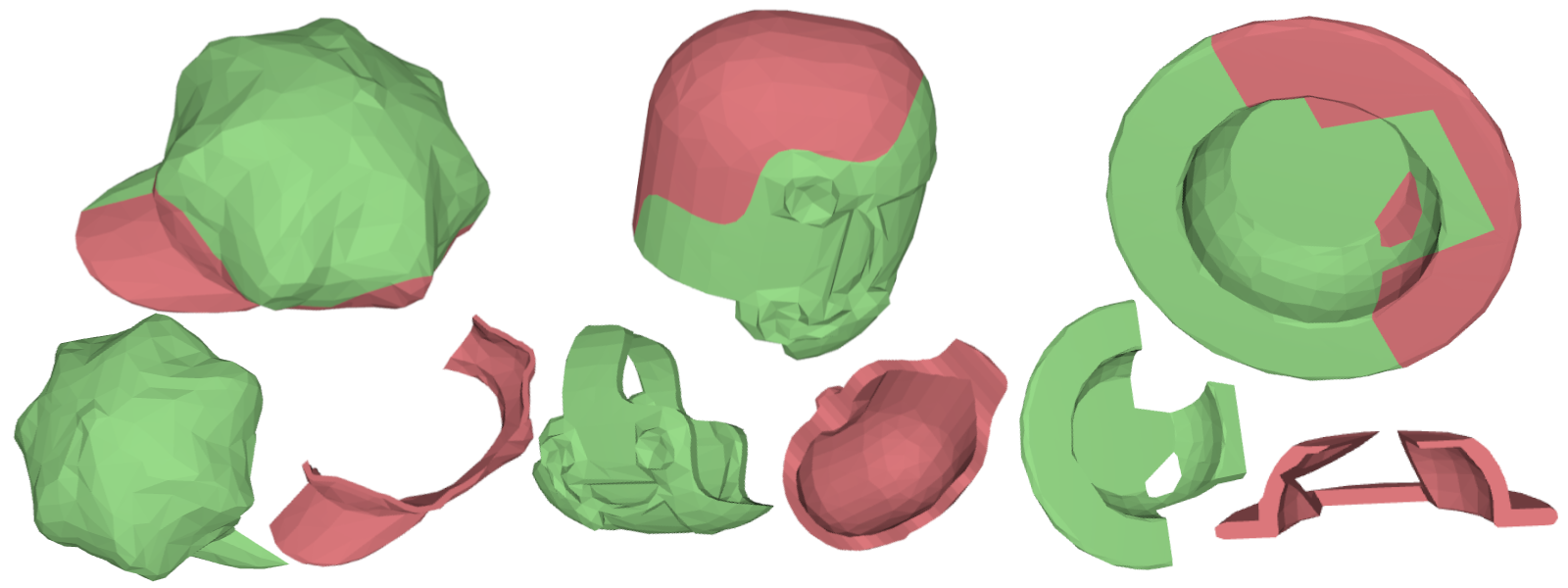} \\
  \vspace{10.0mm}
  \includegraphics[width=\linewidth]{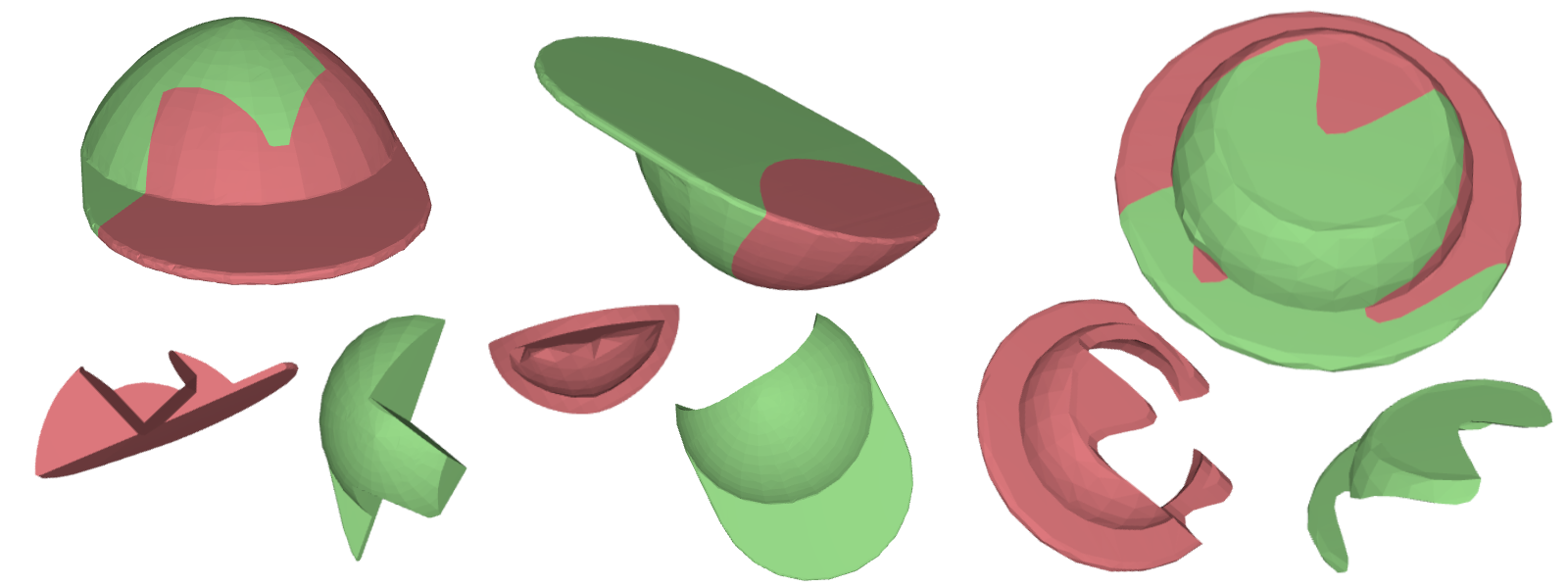}
  \caption{
  \textbf{Dataset visual examples.} 
  We present visual examples of the shape pairs in the hat category.
  }
  \label{fig:dataset-hat}
\end{figure*}

\begin{figure*}[t]
  \centering
  \includegraphics[width=\linewidth]{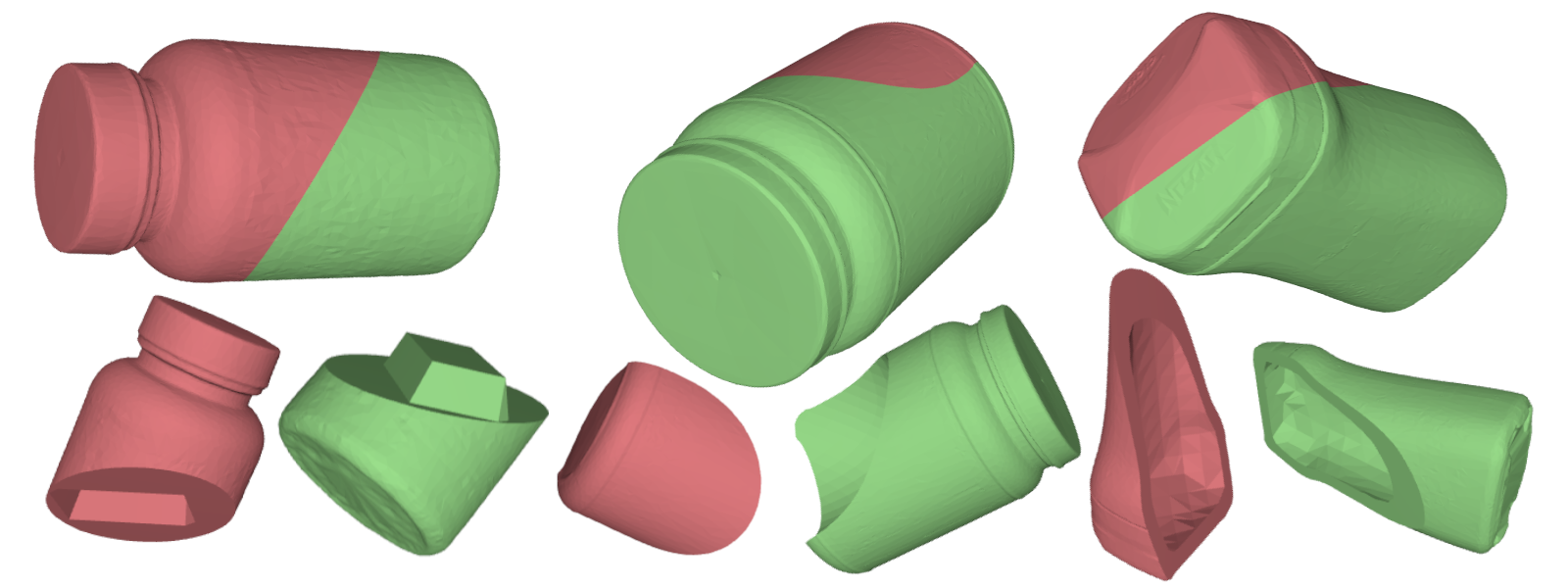} \\
  \vspace{10.0mm}
  \includegraphics[width=\linewidth]{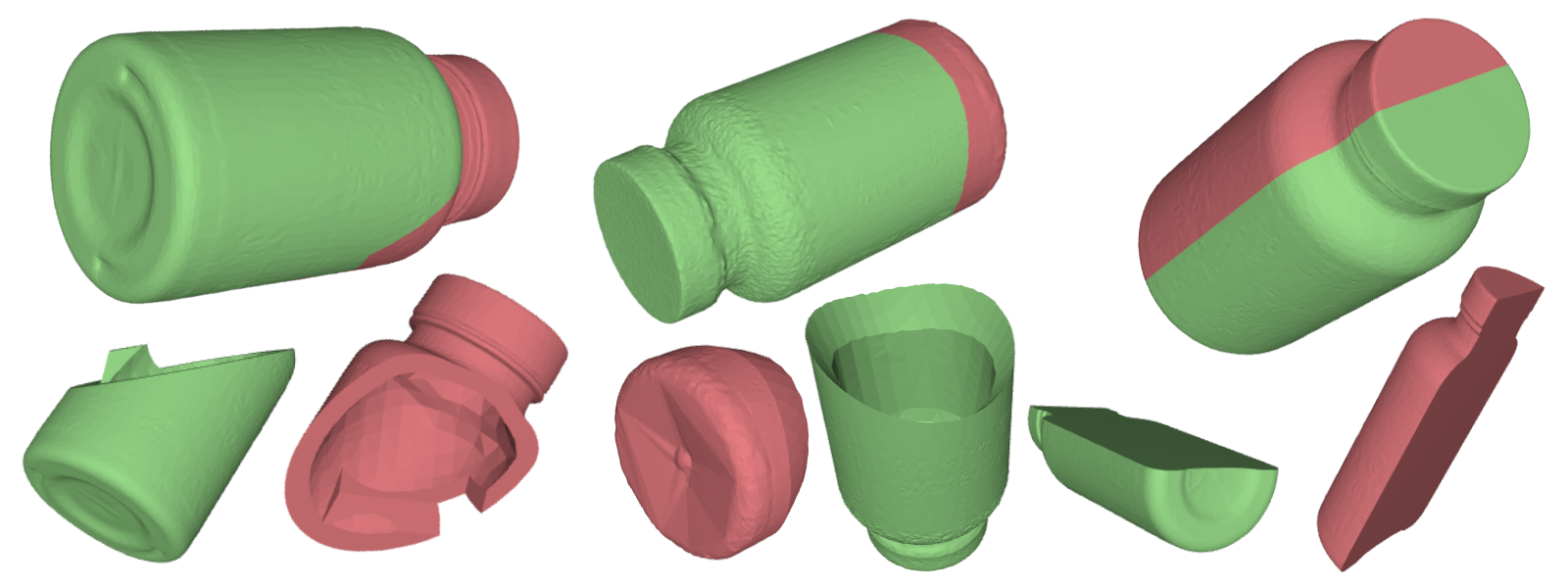} \\
  \vspace{10.0mm}
  \includegraphics[width=\linewidth]{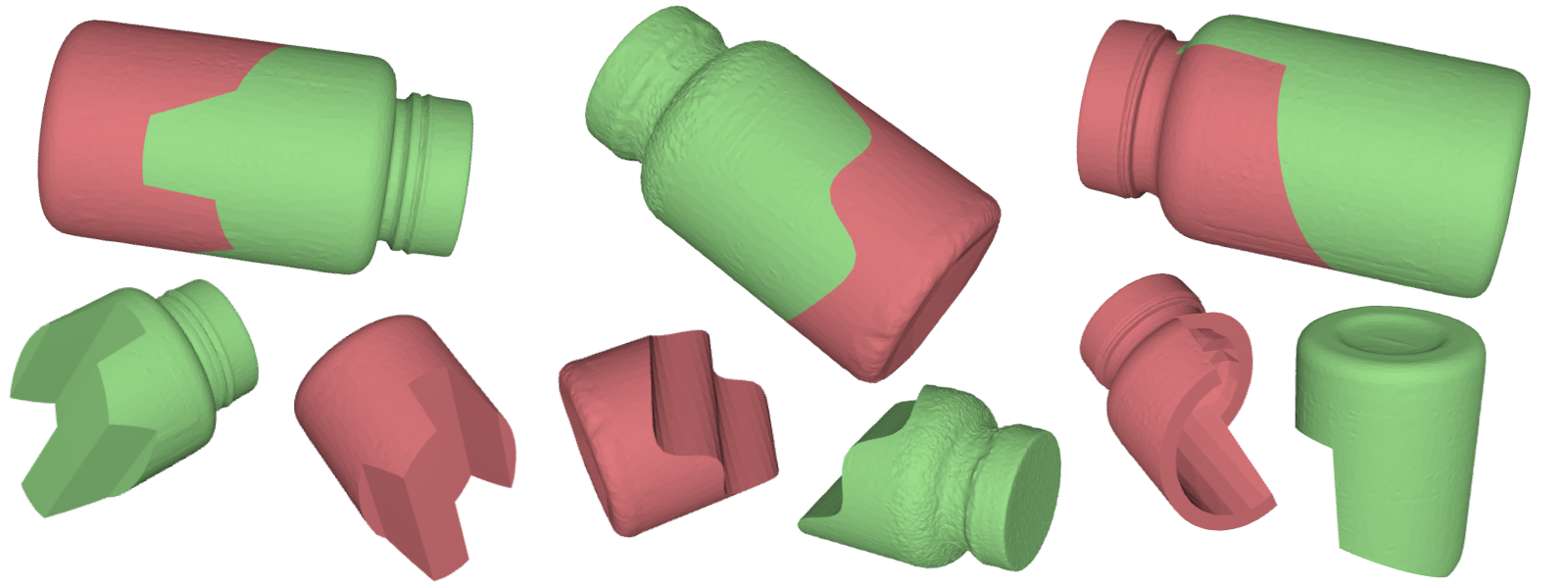}
  \caption{
  \textbf{Dataset visual examples.} 
  We present visual examples of the shape pairs in the jar category.
  }
  \label{fig:dataset-jar}
\end{figure*}

\begin{figure*}[t]
  \centering
  \includegraphics[width=\linewidth]{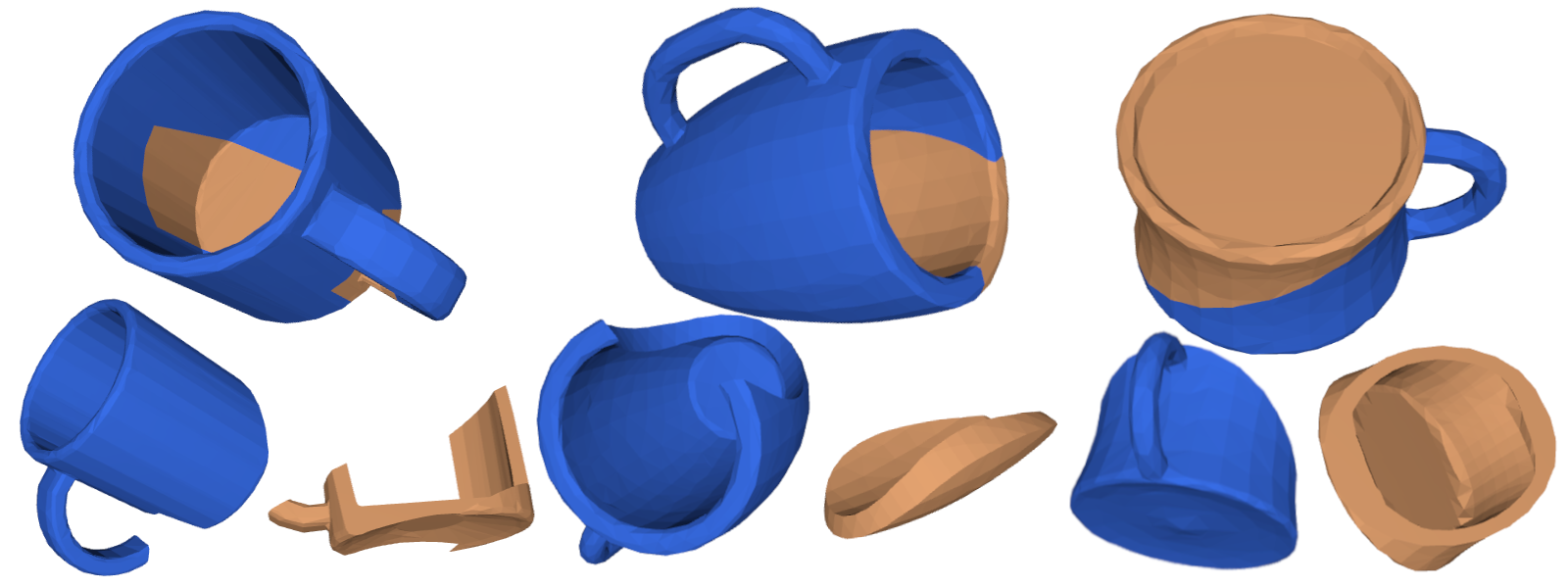} \\
  \vspace{10.0mm}
  \includegraphics[width=\linewidth]{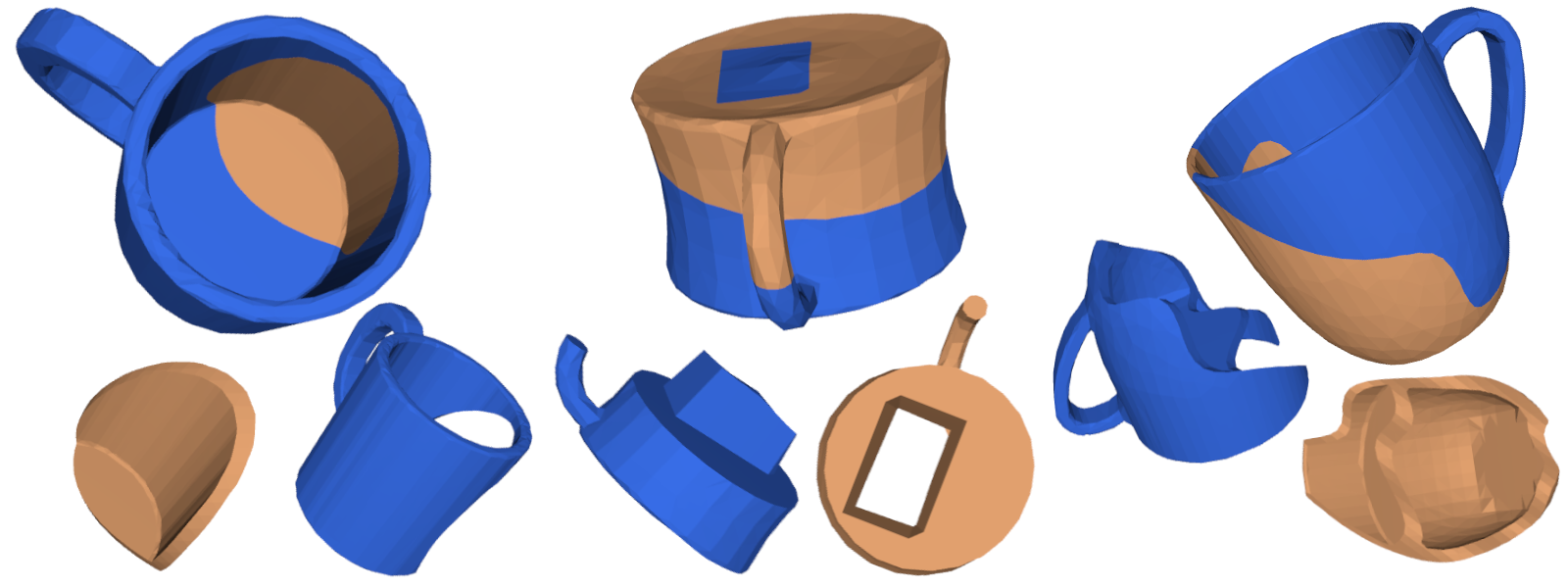} \\
  \vspace{10.0mm}
  \includegraphics[width=\linewidth]{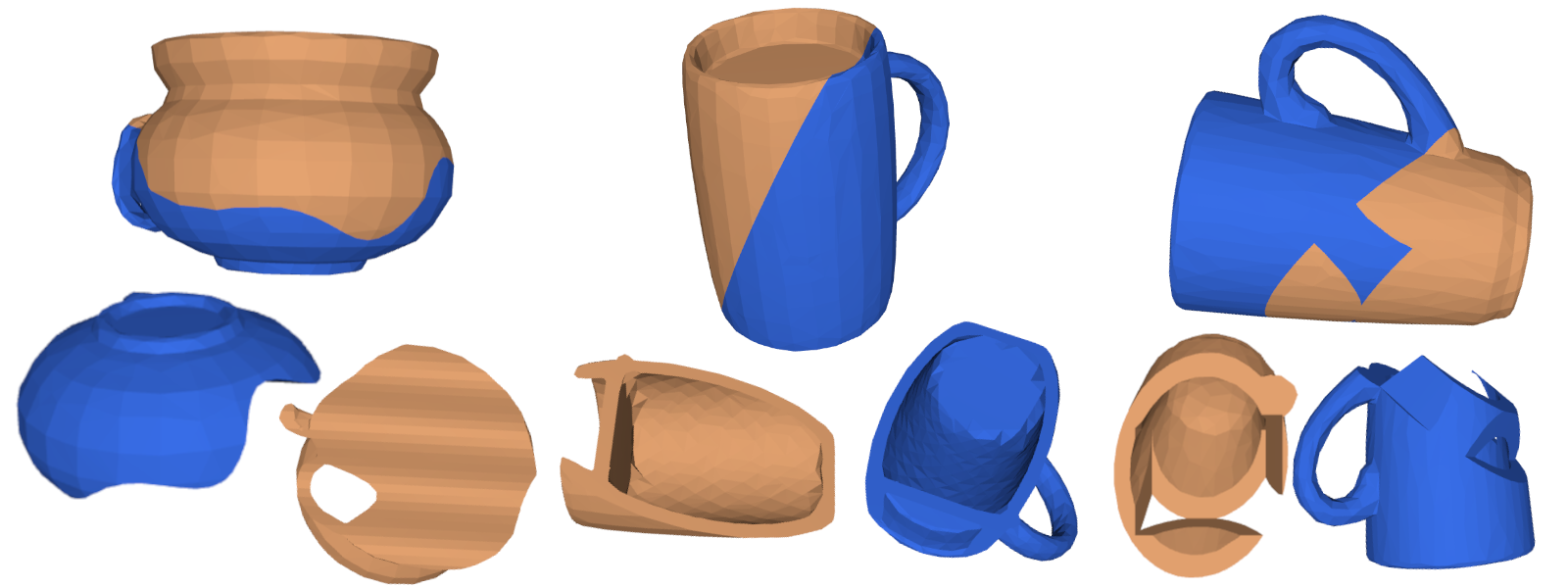}
  \caption{
  \textbf{Dataset visual examples.} 
  We present visual examples of the shape pairs in the mug category.
  }
  \label{fig:dataset-mug}
\end{figure*}

\begin{figure*}[t]
  \centering
  \includegraphics[width=\linewidth]{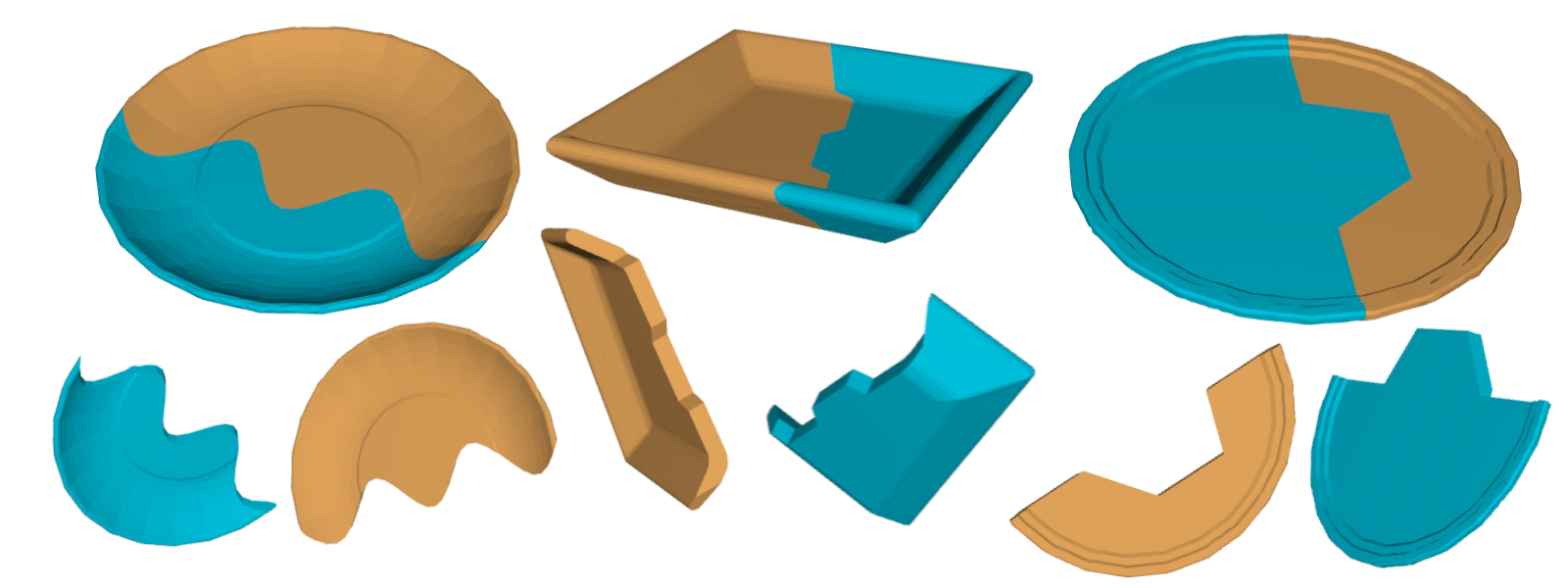} \\
  \vspace{10.0mm}
  \includegraphics[width=\linewidth]{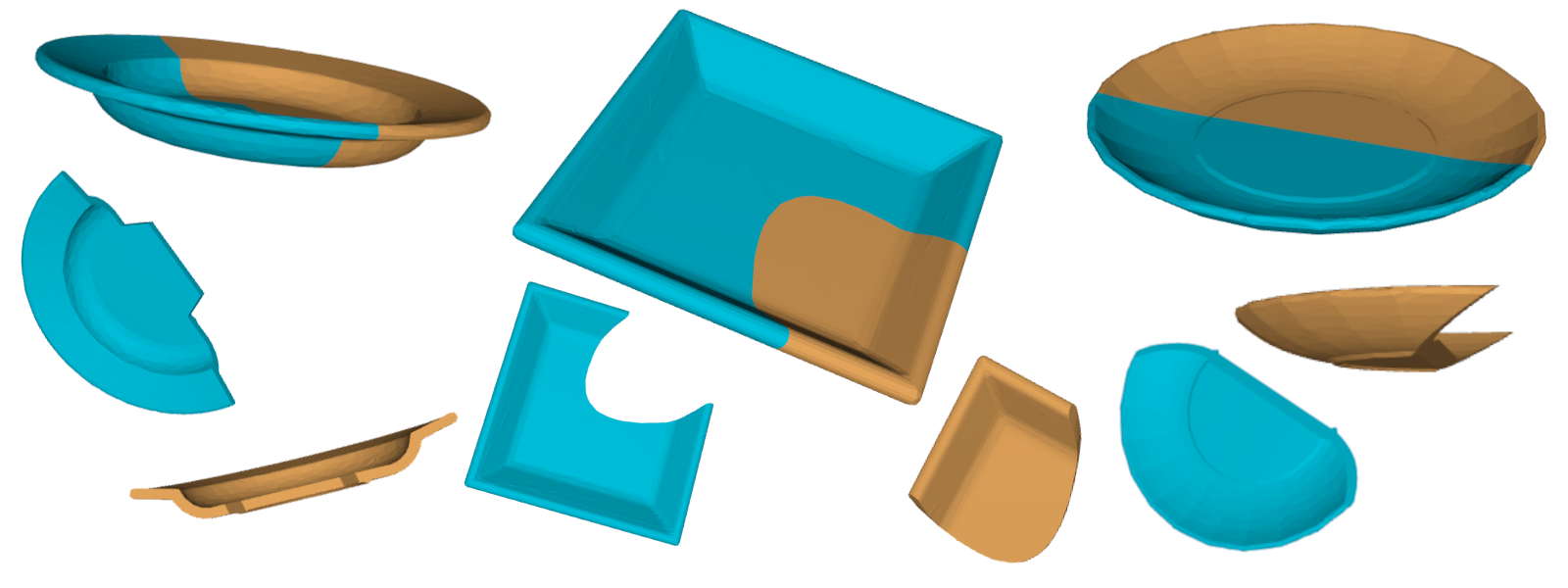} \\
  \vspace{10.0mm}
  \includegraphics[width=\linewidth]{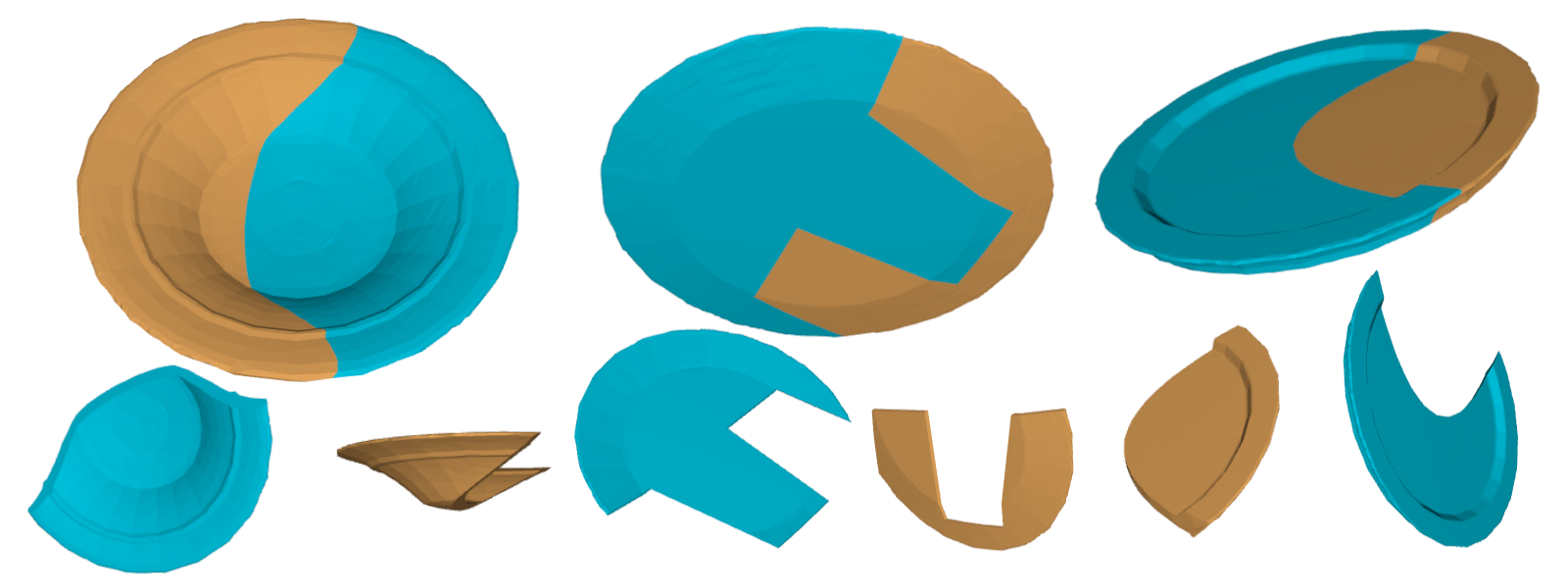}
  \caption{
  \textbf{Dataset visual examples.} 
  We present visual examples of the shape pairs in the plate category.
  }
  \label{fig:dataset-plate}
\end{figure*}

\begin{figure*}[t]
  \centering
  \includegraphics[width=\linewidth]{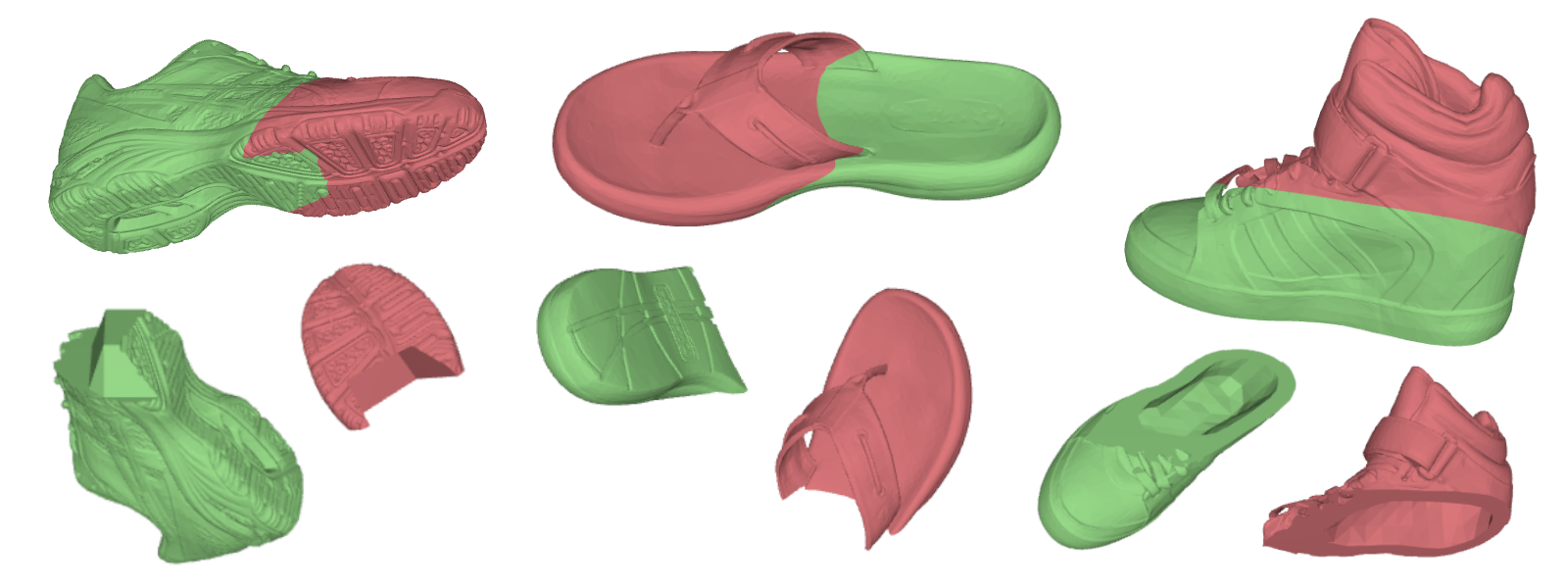} \\
  \vspace{10.0mm}
  \includegraphics[width=\linewidth]{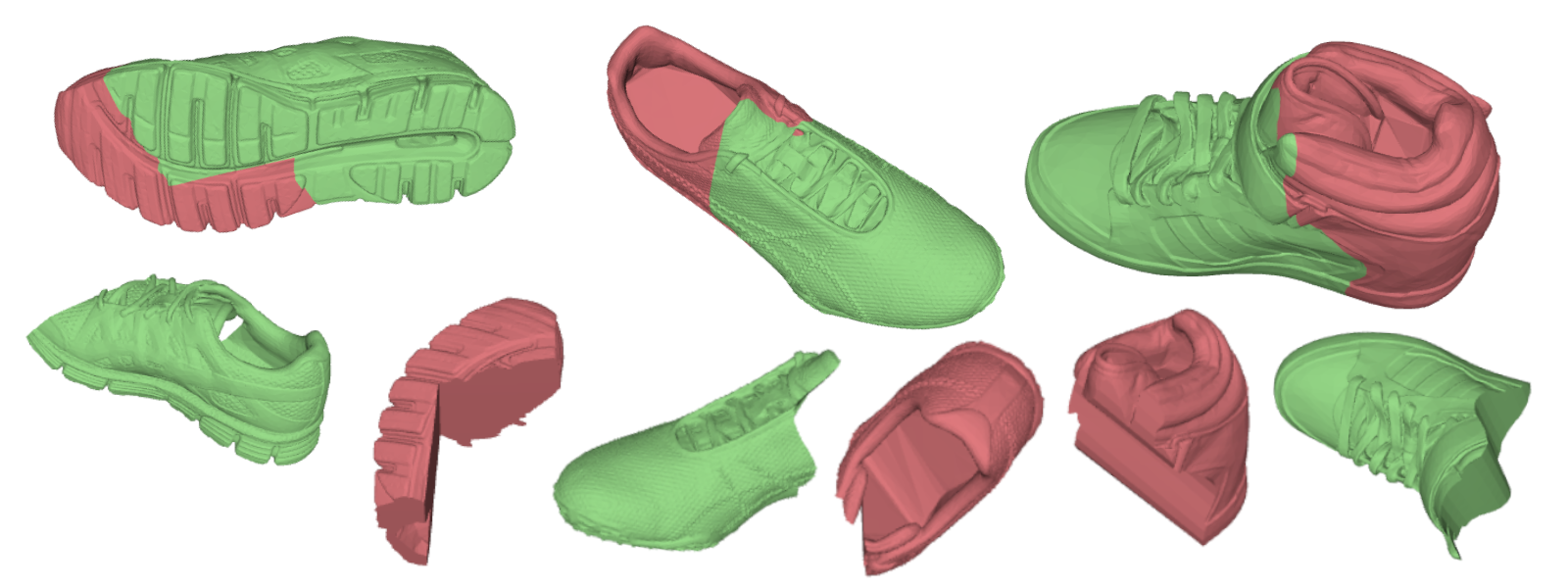} \\
  \vspace{10.0mm}
  \includegraphics[width=\linewidth]{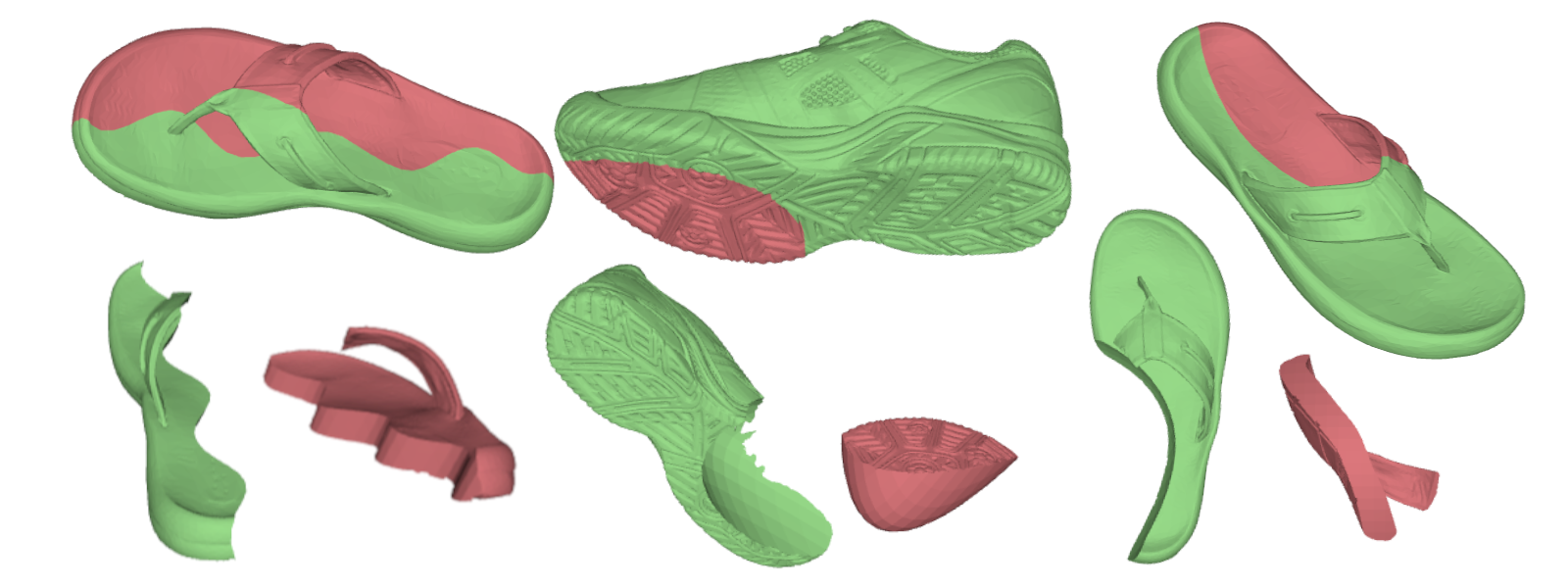}
  \caption{
  \textbf{Dataset visual examples.} 
  We present visual examples of the shape pairs in the shoe category.
  }
  \label{fig:dataset-shoe}
\end{figure*}

\begin{figure*}[t]
  \centering
  \includegraphics[width=\linewidth]{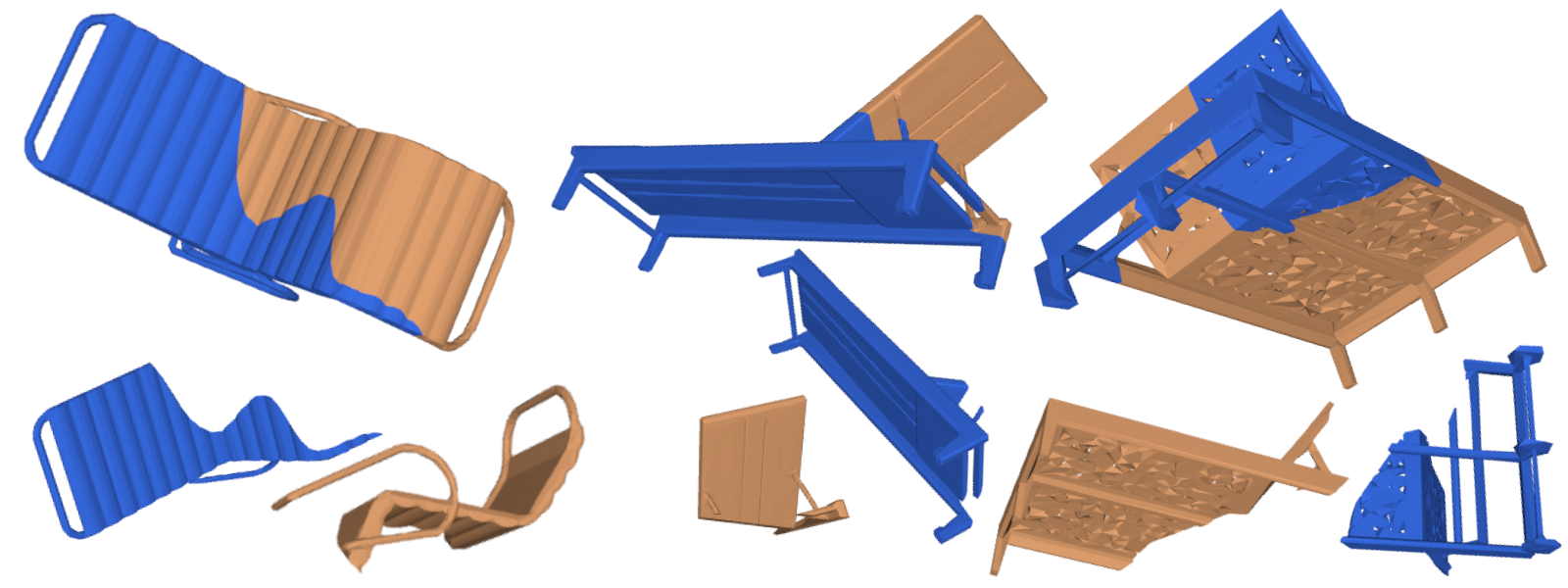} \\
  \vspace{10.0mm}
  \includegraphics[width=\linewidth]{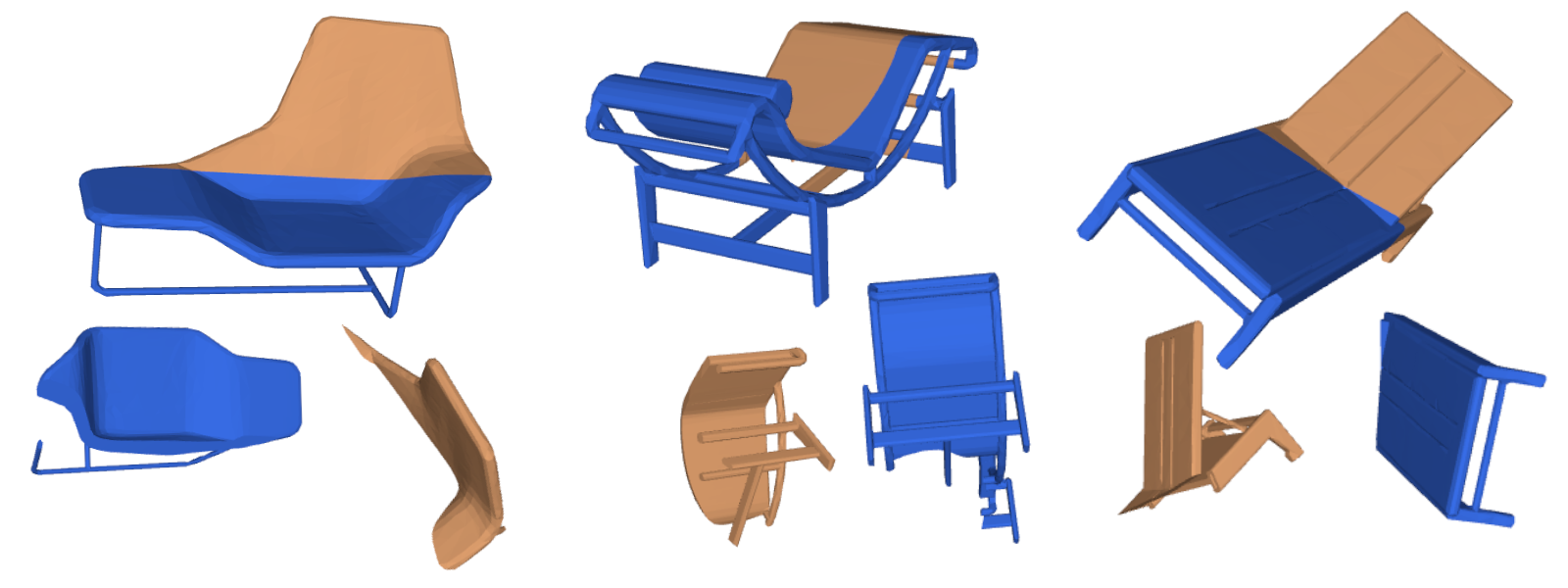} \\
  \vspace{10.0mm}
  \includegraphics[width=\linewidth]{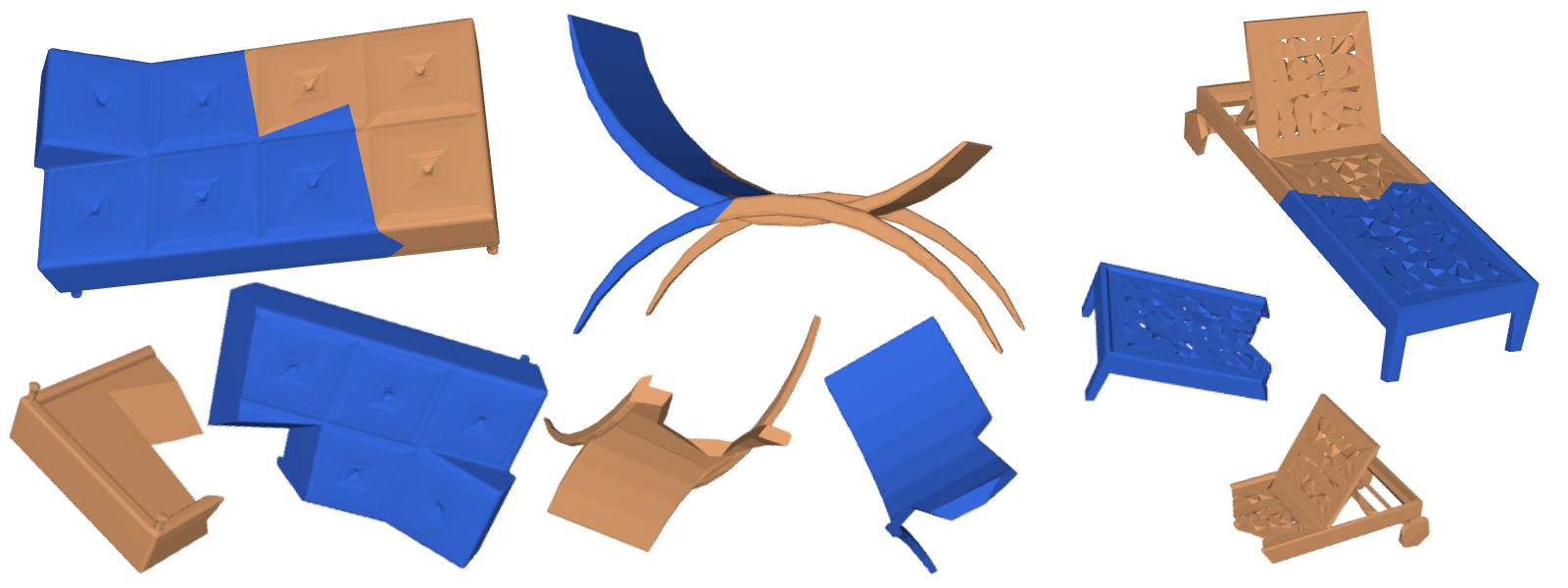}
  \caption{
  \textbf{Dataset visual examples.} 
  We present visual examples of the shape pairs in the sofa category.
  }
  \label{fig:dataset-sofa}
\end{figure*}

\begin{figure*}[t]
  \centering
  \includegraphics[width=\linewidth]{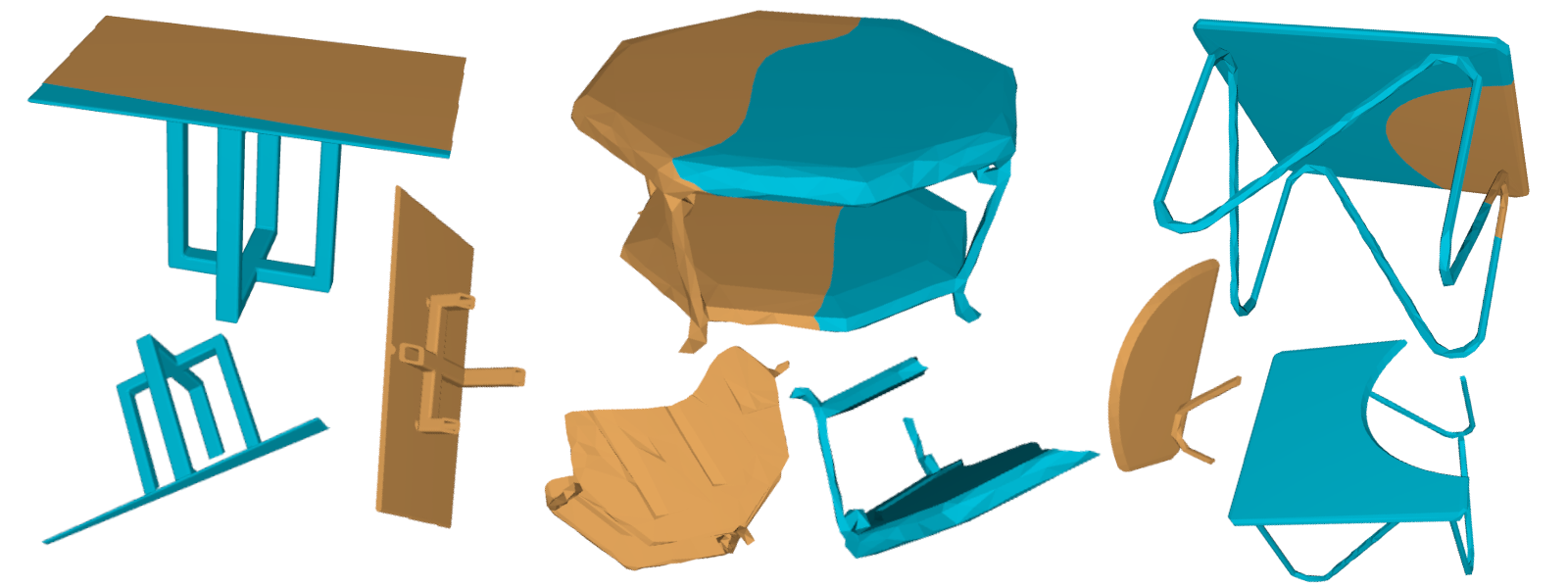} \\
  \vspace{10.0mm}
  \includegraphics[width=\linewidth]{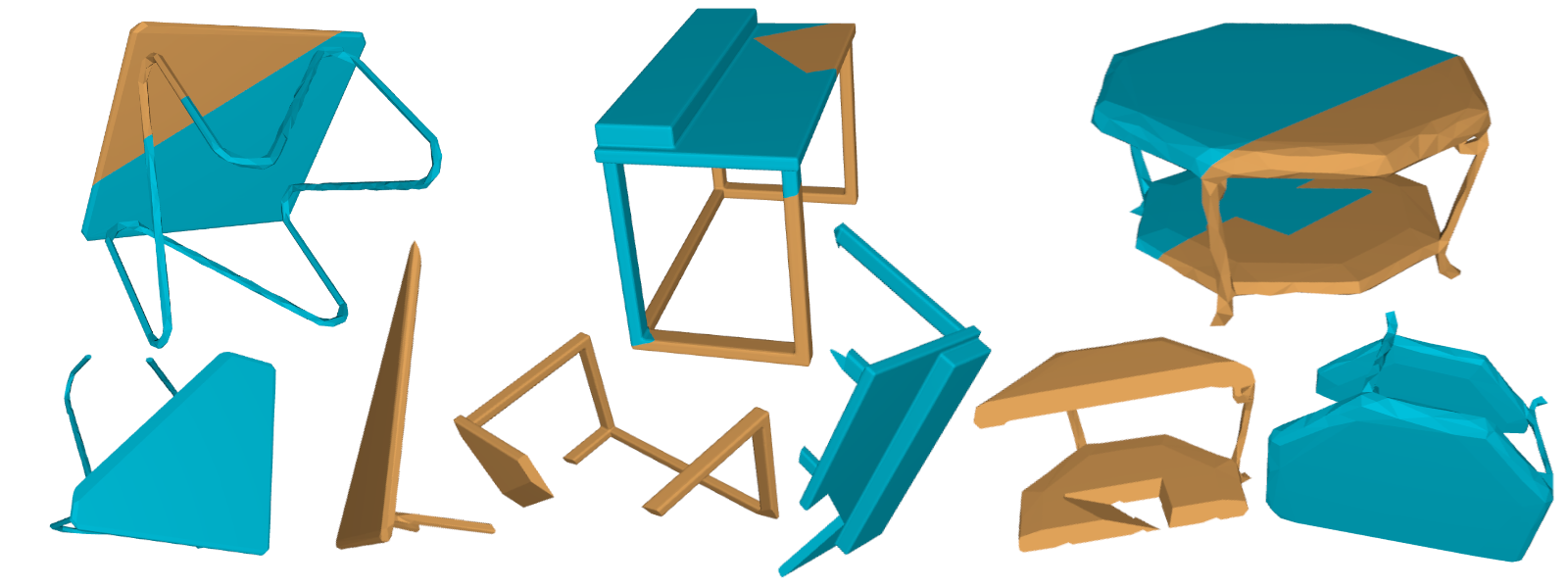} \\
  \vspace{10.0mm}
  \includegraphics[width=\linewidth]{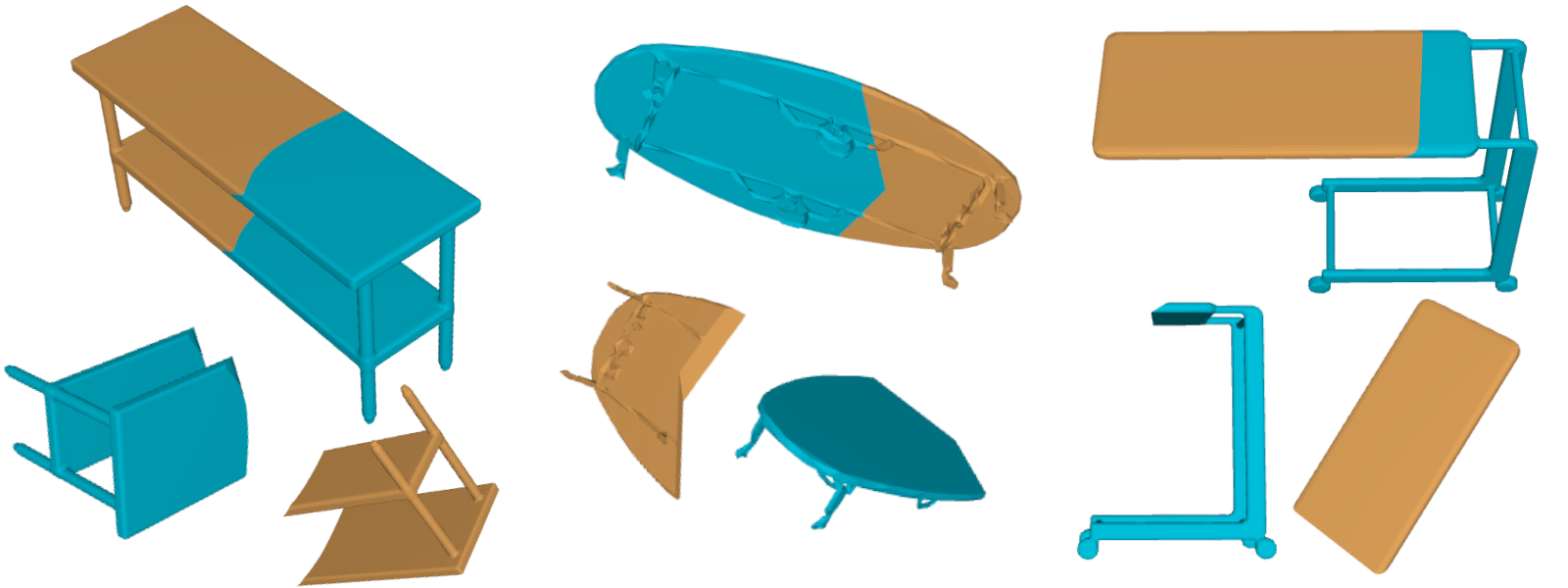}
  \caption{
  \textbf{Dataset visual examples.} 
  We present visual examples of the shape pairs in the table category.
  }
  \label{fig:dataset-table}
\end{figure*}

\begin{figure*}[t]
  \centering
  \includegraphics[width=\linewidth]{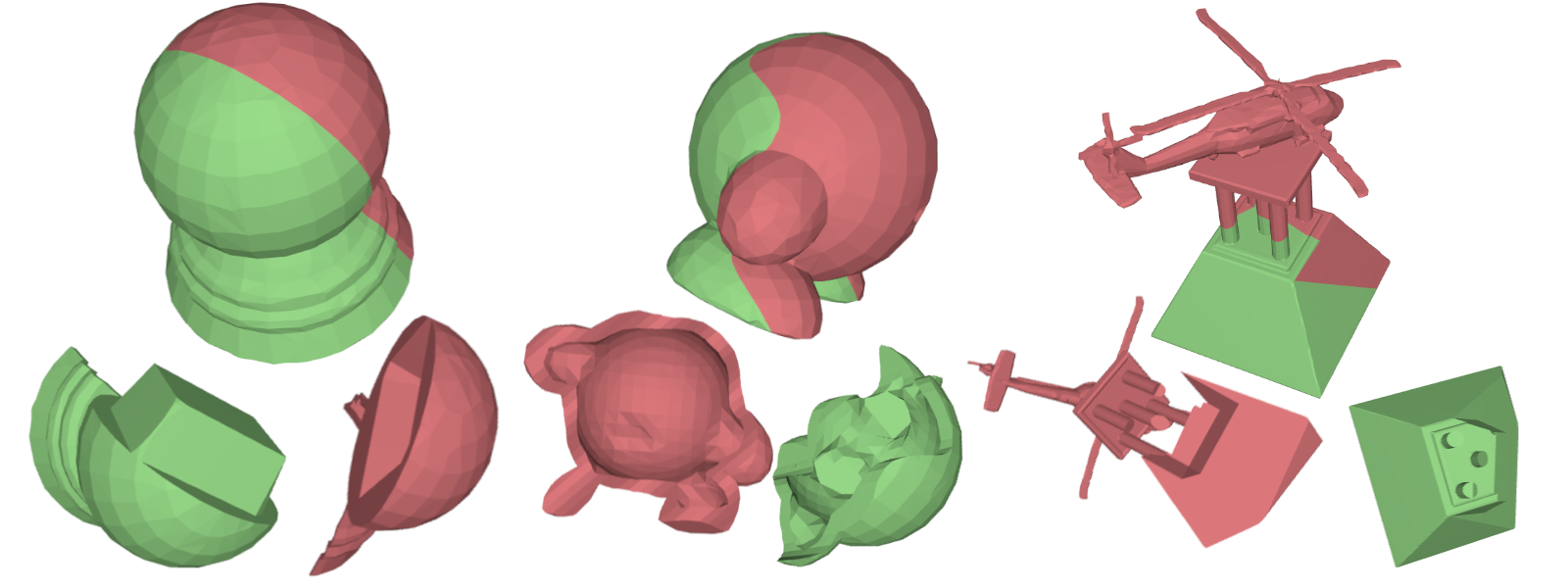} \\
  \vspace{10.0mm}
  \includegraphics[width=\linewidth]{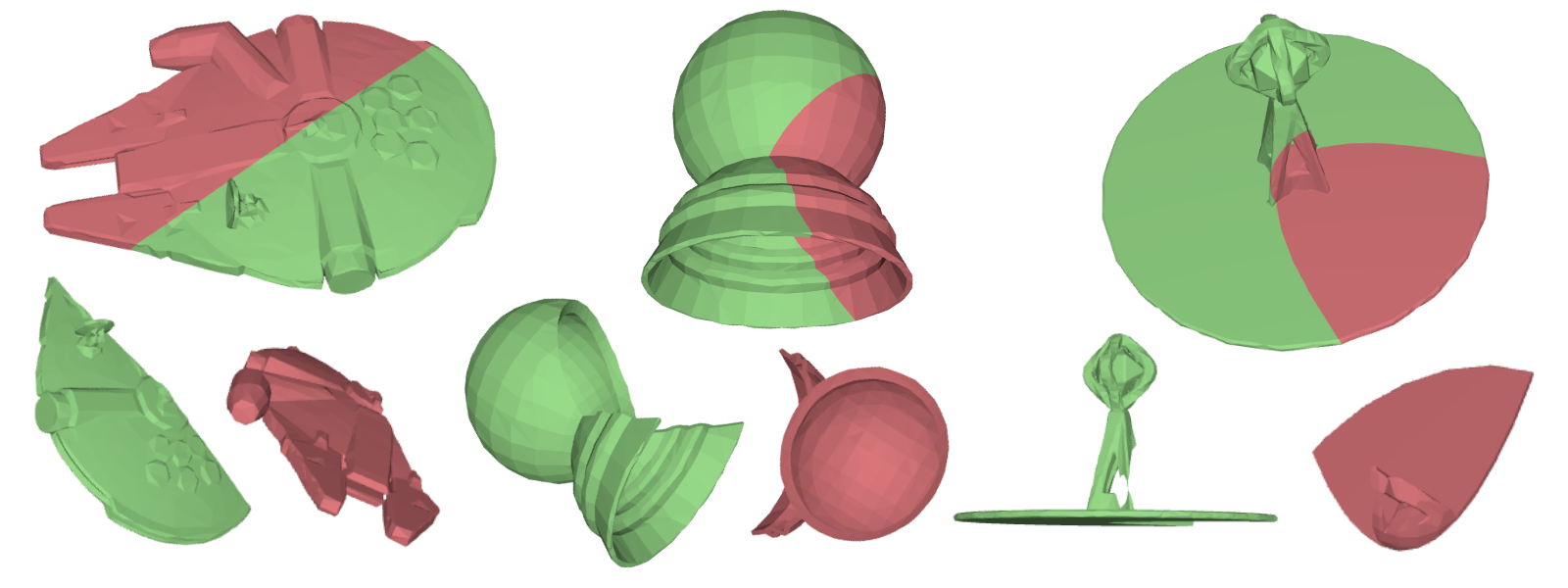} \\
  \vspace{10.0mm}
  \includegraphics[width=\linewidth]{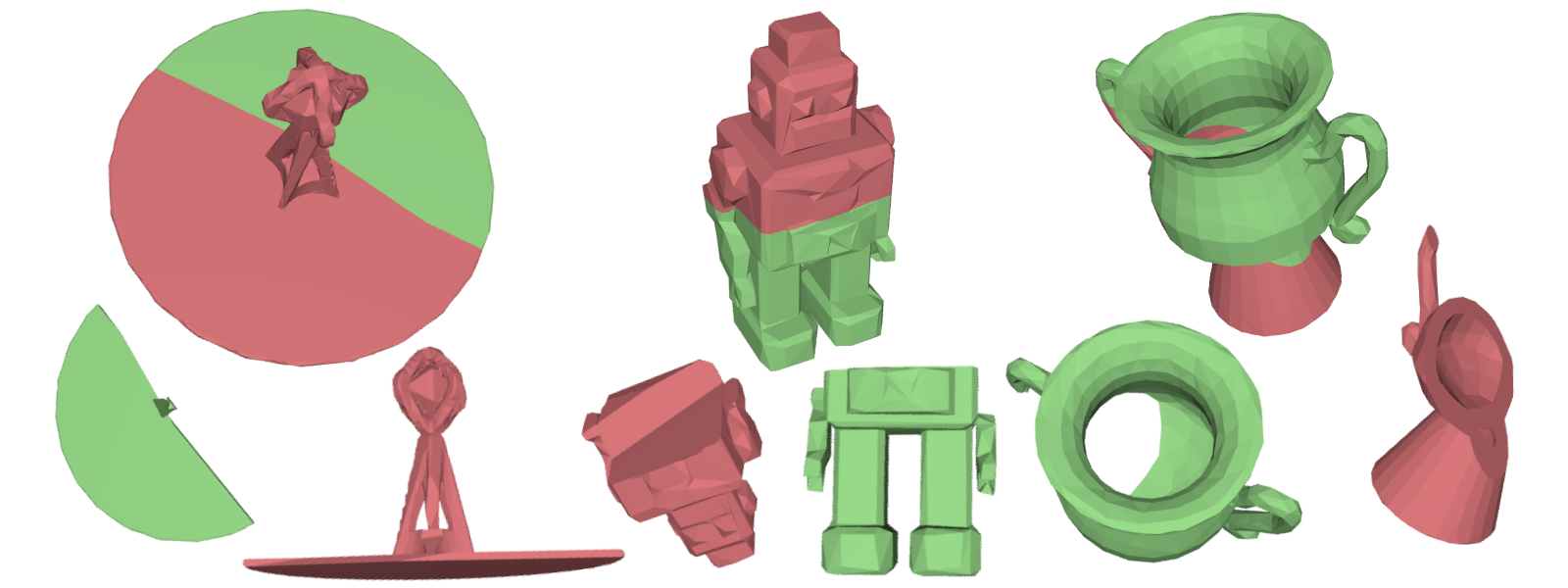}
  \caption{
  \textbf{Dataset visual examples.} 
  We present visual examples of the shape pairs in the toy category.
  }
  \label{fig:dataset-toy}
\end{figure*}

We present visual examples of the Geometric Shape Mating dataset in Figure~\ref{fig:dataset-bag} (bag), Figure~\ref{fig:dataset-bowl} (bowl), Figure~\ref{fig:dataset-box} (box), Figure~\ref{fig:dataset-hat} (hat), Figure~\ref{fig:dataset-jar} (jar), Figure~\ref{fig:dataset-mug} (mug), Figure~\ref{fig:dataset-plate} (plate), Figure~\ref{fig:dataset-shoe} (shoe), Figure~\ref{fig:dataset-sofa} (sofa), Figure~\ref{fig:dataset-table} (table), and Figure~\ref{fig:dataset-toy} (toy).

\section{Limitations and Future Work} \label{app:limitations}

While pairwise geometric shape mating is a core task to solve multi-part assembly, our current method is designed for mating two object parts.
Future work can be extending pairwise shape mating to multi-part shape mating.
On the other hand, our method currently assumes access to full point clouds.
A limitation of our method is mating two shapes given only partial point clouds.
Furthermore, if there are multiple possible mating configurations for a given pair of shapes, our model will only predict one solution.
Future work can be developing methods that learn the distribution of all possible solutions and predict shape mating configurations conditioned on an input signal (i.e., conditional shape mating).

{\small
\bibliographystyle{ieee_fullname}
\bibliography{reference}
}

\begin{figure*}[t]
  \begin{center}
  \mpage{0.16}{\includegraphics[width=\linewidth]{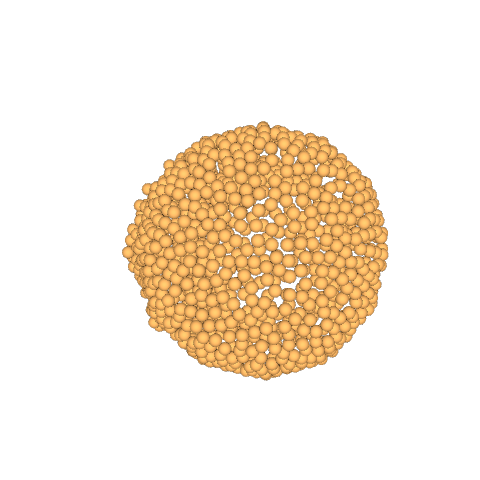}} \hfill
  \mpage{0.16}{\includegraphics[width=\linewidth]{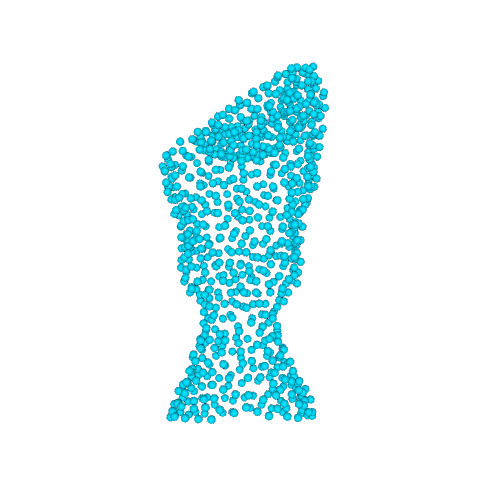}} \hfill
  \mpage{0.20}{\includegraphics[width=\linewidth]{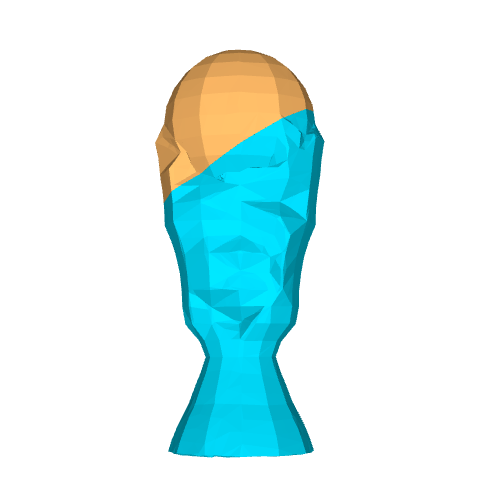}} \hfill
  \mpage{0.20}{\includegraphics[width=\linewidth]{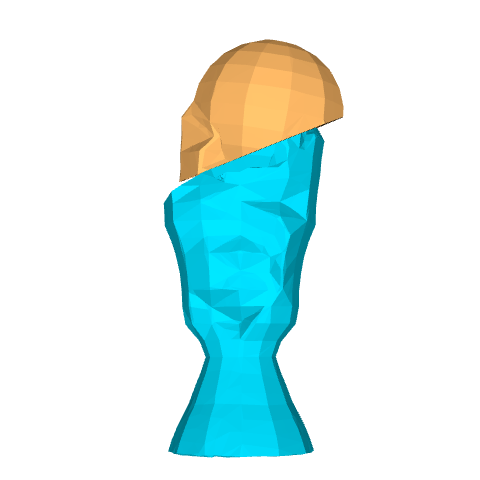}} \hfill
  \mpage{0.20}{\includegraphics[width=\linewidth]{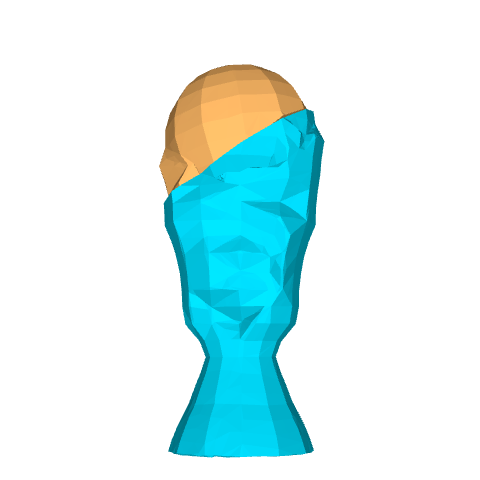}} \\
  \vspace{-2.0mm}
  \mpage{0.16}{\includegraphics[width=\linewidth]{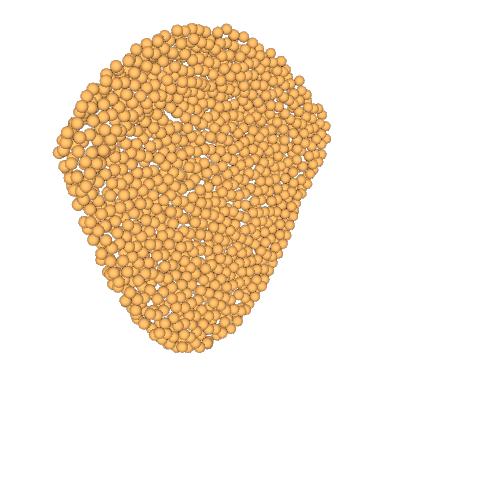}} \hfill
  \mpage{0.16}{\includegraphics[width=\linewidth]{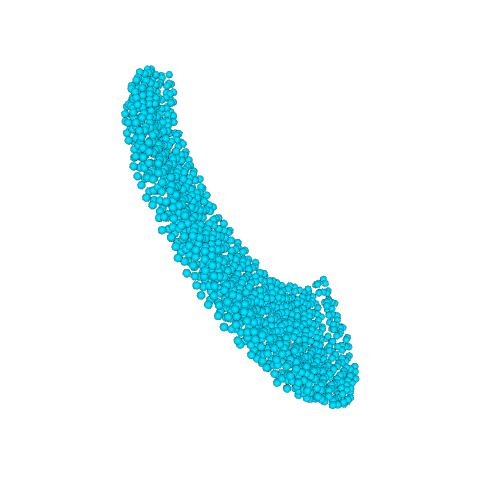}} \hfill
  \mpage{0.20}{\includegraphics[width=\linewidth]{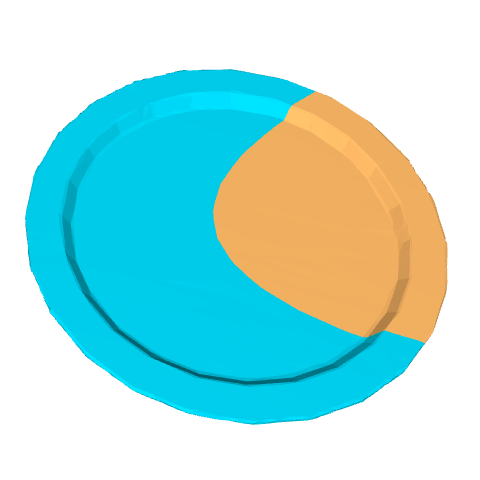}} \hfill
  \mpage{0.20}{\includegraphics[width=\linewidth]{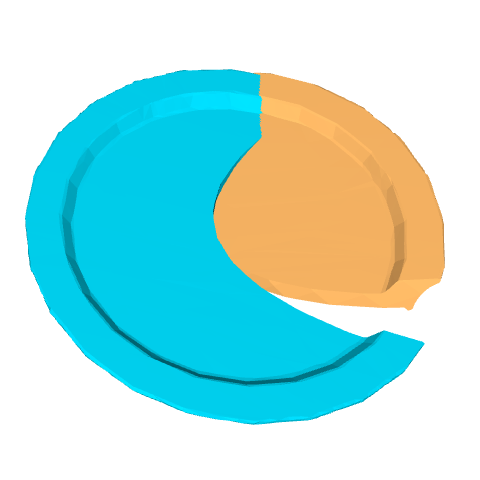}} \hfill
  \mpage{0.20}{\includegraphics[width=\linewidth]{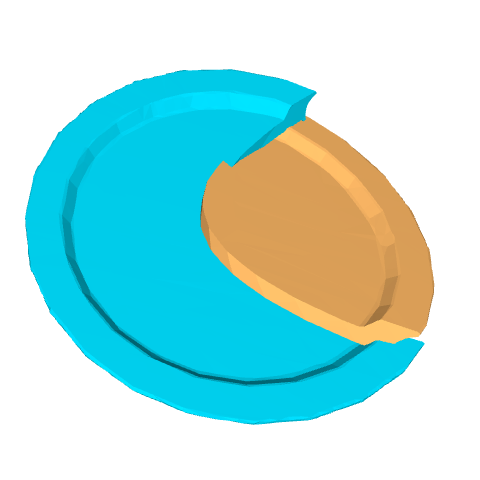}} \\
  \vspace{-4.0mm}
  \mpage{0.16}{\includegraphics[width=\linewidth]{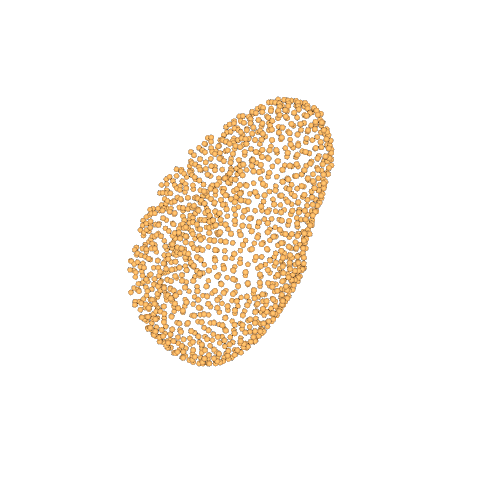}} \hfill
  \mpage{0.16}{\includegraphics[width=\linewidth]{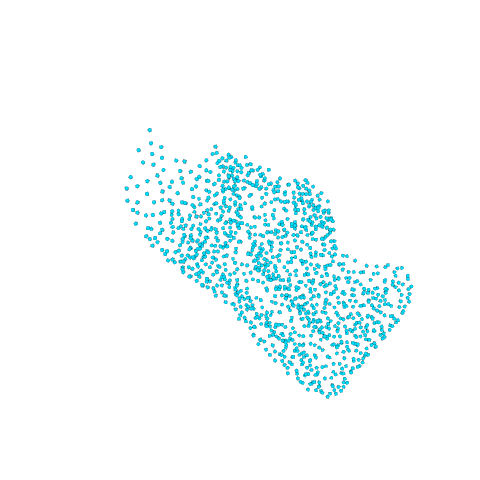}} \hfill
  \mpage{0.20}{\includegraphics[width=\linewidth]{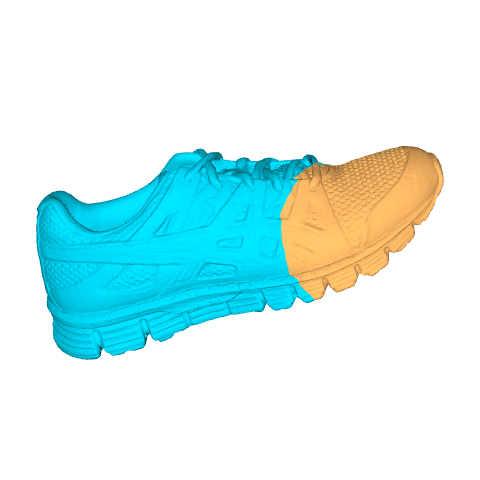}} \hfill
  \mpage{0.20}{\includegraphics[width=\linewidth]{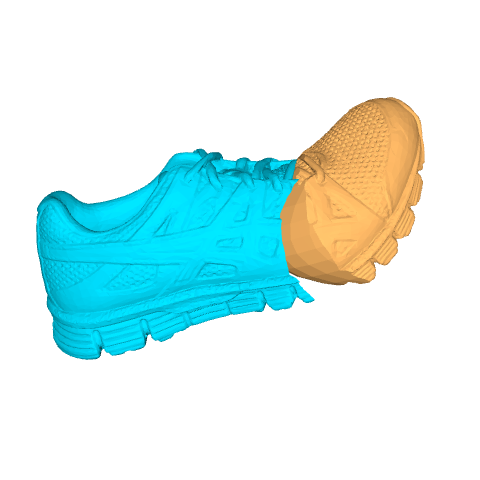}} \hfill
  \mpage{0.20}{\includegraphics[width=\linewidth]{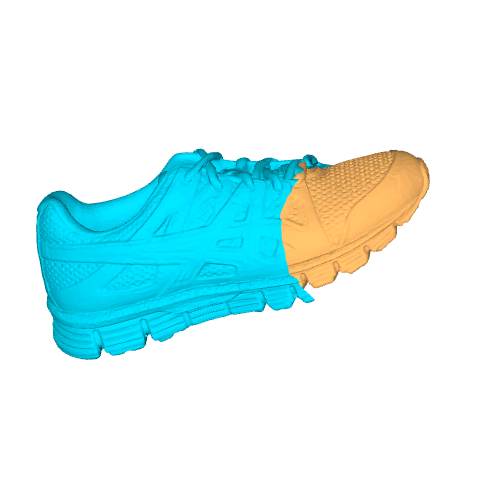}} \\
  \vspace{-4.0mm}
  \mpage{0.16}{\includegraphics[width=\linewidth]{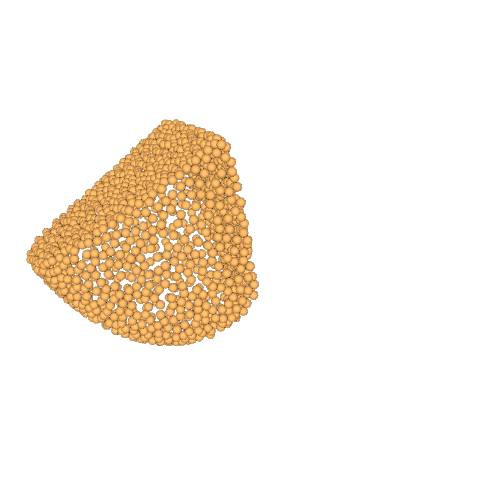}} \hfill
  \mpage{0.16}{\includegraphics[width=\linewidth]{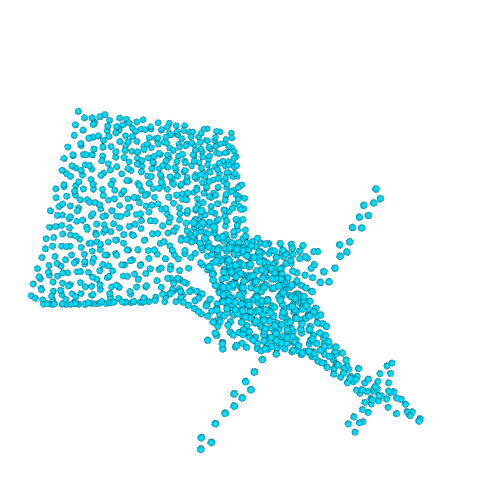}} \hfill
  \mpage{0.20}{\includegraphics[width=\linewidth]{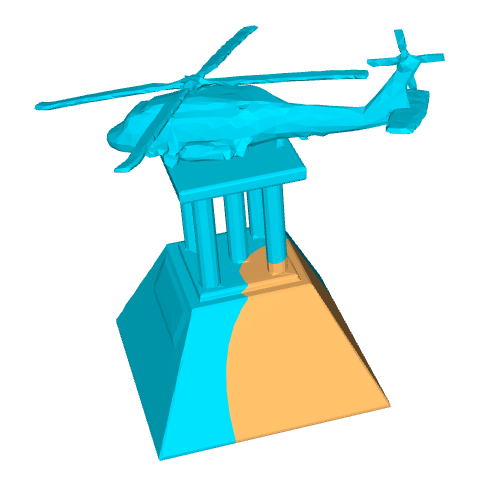}} \hfill
  \mpage{0.20}{\includegraphics[width=\linewidth]{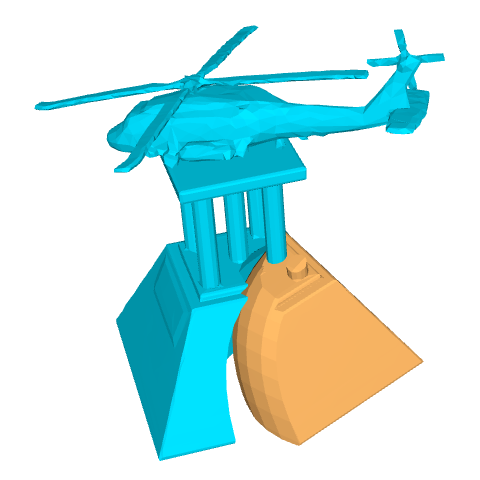}} \hfill
  \mpage{0.20}{\includegraphics[width=\linewidth]{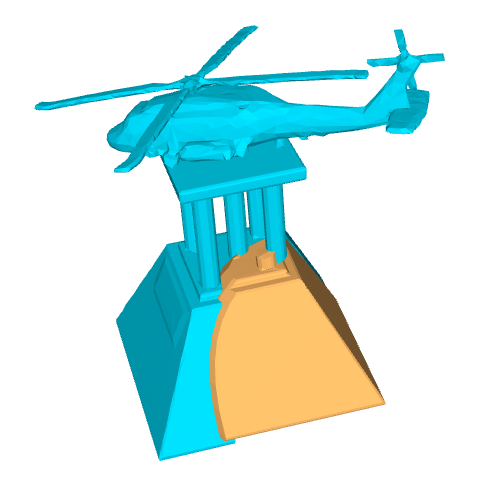}} \\
  \vspace{-2.0mm}
  \mpage{0.16}{\includegraphics[width=\linewidth]{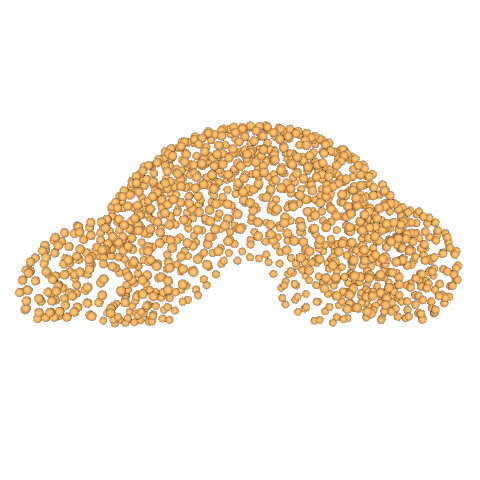}} \hfill
  \mpage{0.16}{\includegraphics[width=\linewidth]{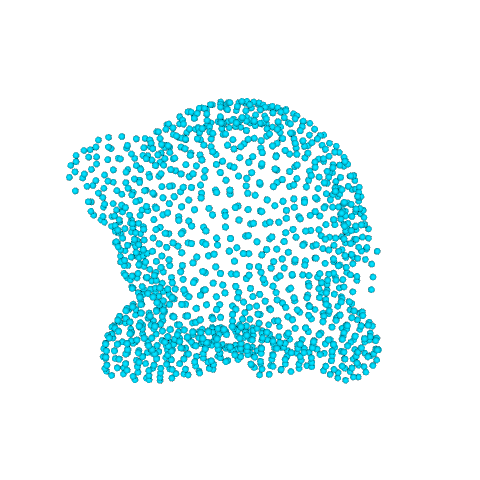}} \hfill
  \mpage{0.20}{\includegraphics[width=\linewidth]{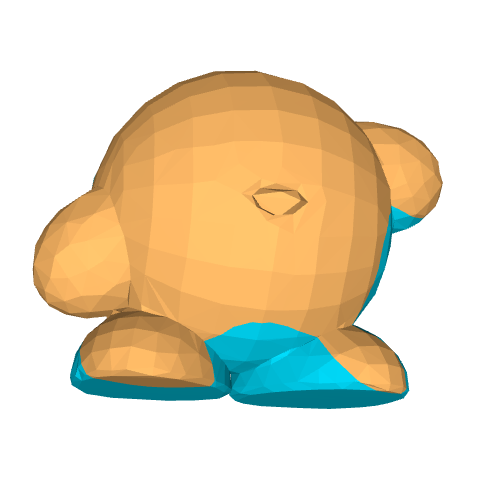}} \hfill
  \mpage{0.20}{\includegraphics[width=\linewidth]{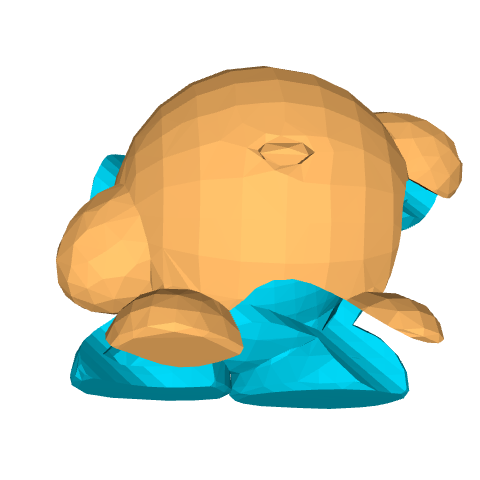}} \hfill
  \mpage{0.20}{\includegraphics[width=\linewidth]{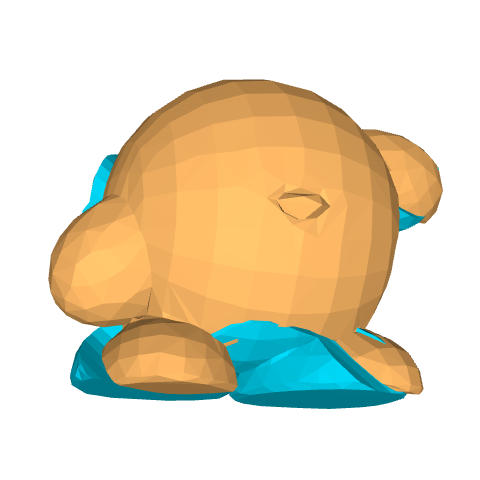}} \\
  \vspace{-4.0mm}
  \mpage{0.16}{\includegraphics[width=\linewidth]{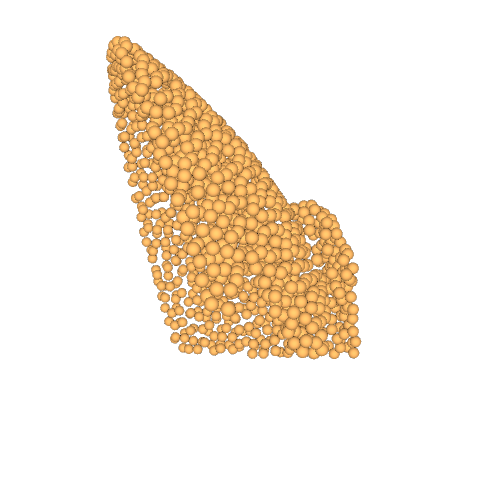}} \hfill
  \mpage{0.16}{\includegraphics[width=\linewidth]{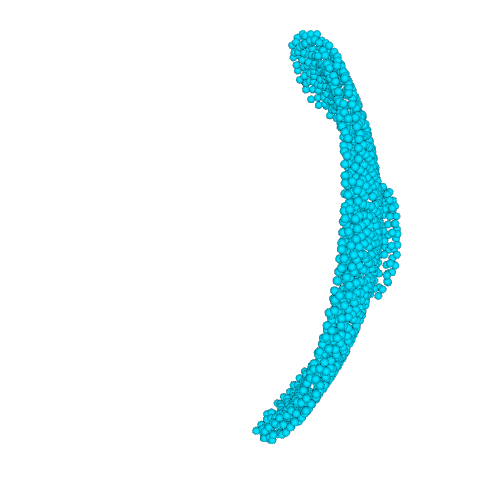}} \hfill
  \mpage{0.20}{\includegraphics[width=\linewidth]{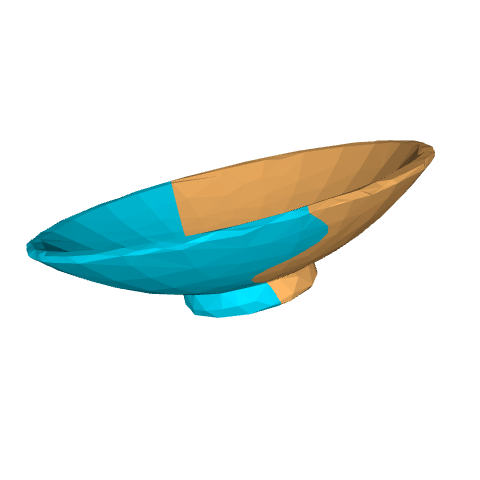}} \hfill
  \mpage{0.20}{\includegraphics[width=\linewidth]{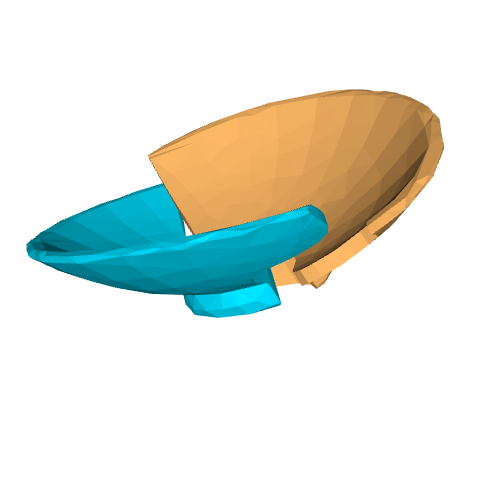}} \hfill
  \mpage{0.20}{\includegraphics[width=\linewidth]{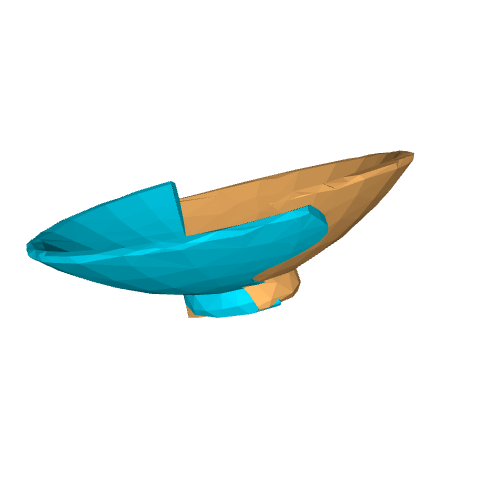}} \\
  \vspace{-6.0mm}
  \mpage{0.16}{\includegraphics[width=\linewidth]{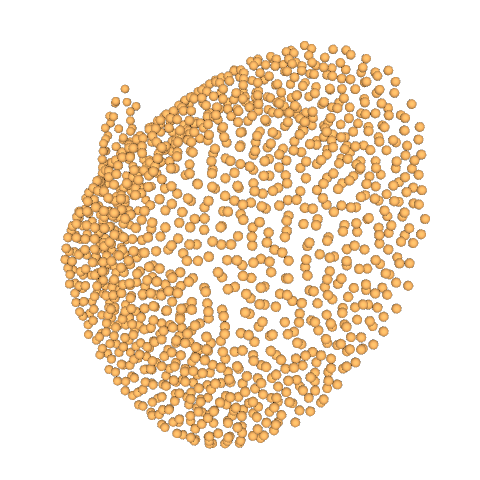}} \hfill
  \mpage{0.16}{\includegraphics[width=\linewidth]{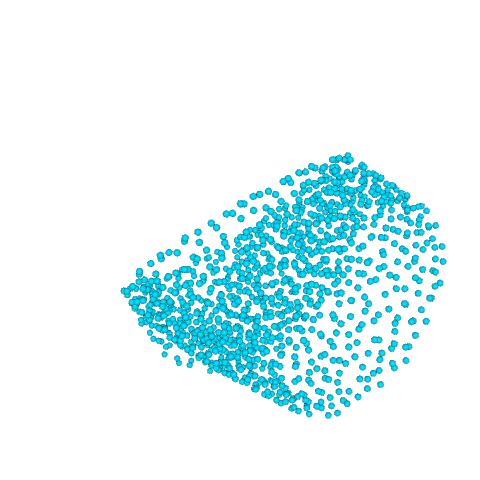}} \hfill
  \mpage{0.20}{\includegraphics[width=\linewidth]{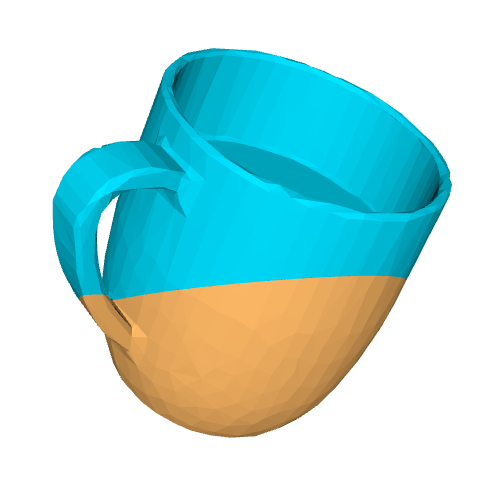}} \hfill
  \mpage{0.20}{\includegraphics[width=\linewidth]{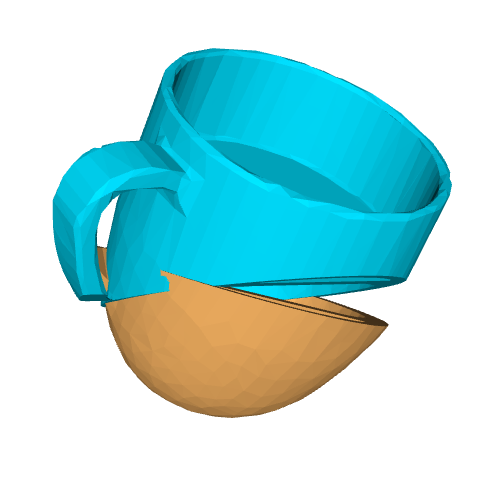}} \hfill
  \mpage{0.20}{\includegraphics[width=\linewidth]{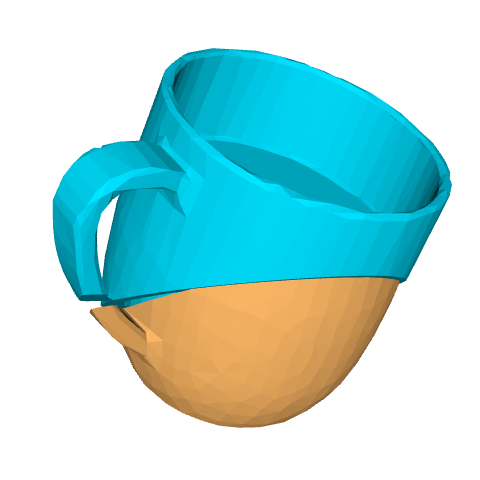}} \\
  \mpage{0.32}{Input Point Clouds} \hfill
  \mpage{0.20}{Ground Truth} \hfill
  \mpage{0.20}{GNN Assembly} \hfill
  \mpage{0.20}{\algoName} \\
  \caption{
  \textbf{Qualitative results of pairwise shape mating.}
  NSM predicts poses that accurately mate the two shapes together.
  }
  \label{fig:app-main}
  \end{center}
\end{figure*}

\end{document}